\documentclass[12pt]{iopart}

\usepackage{geometry}
\usepackage[T1]{fontenc}
\usepackage{lmodern}
\usepackage{iopams} 
\usepackage[symbol]{footmisc}
\usepackage{import}
\usepackage{iopams} 
\usepackage[symbol]{footmisc}
\usepackage[utf8]{inputenc} 
\usepackage[T1]{fontenc}    
\usepackage{hyperref}       
\usepackage{booktabs}       
\usepackage{nicefrac}       
\usepackage{microtype}      
\usepackage{enumitem}
\usepackage{wrapfig}
\usepackage{cancel}
\usepackage{pgfplots}
\usepackage{pdfpages}
\usepackage{listliketab}
\usepackage{graphicx}
\usepackage{amsmath,amssymb,amsthm,amscd,amsfonts,amsbsy}
\usepackage{mathtools}
\usepackage{url}
\usepackage{verbatim}
\usepackage{slashed}
\usepackage{bbm}
\usepackage{algorithm}
\usepackage{algorithmicx}
\usepackage{algpseudocode}
\usepackage{array}
\usepackage{csquotes}
\usepackage{braket}
\usepackage{float}
\usepackage[bf]{caption} 
\usepackage{adjustbox}
\usepackage{soul}
\usepackage{subcaption}

\usepackage[
backend=bibtex,
style=phys,
]{biblatex}
\usepackage{tabularray}

\makeatletter
\long\def\@makefntext#1{\parindent 1em\noindent 
 \makebox[1em][l]{\footnotesize\rm$\m@th{\arabic{footnote}}$}%
 \footnotesize\rm #1}
\def\@makefnmark{\textsuperscript{\hbox{${\arabic{footnote}}\m@th$}}}
\def\@thefnmark{\arabic{footnote}}
\makeatother

\algdef{SxNE}[DOBLOCK]{Do}{EndDoBlock}[1]{\algorithmicdo\ #1}{}%

\newcommand{\bitem}{\begin{list}{$\bullet$}%
{\setlength{\itemsep}{0pt}\setlength{\topsep}{0pt}%
\setlength{\rightmargin}{0pt}}} \newcommand{\eitem}{\end{list}}

\DeclareMathOperator*{\argmin}{arg\,min}

\theoremstyle{theorem}
\newtheorem{thm}{Theorem}[section]
\newtheorem{lem}[thm]{Lemma}
\newtheorem{prop}[thm]{Proposition}
\newtheorem{cor}[thm]{Corollary}

\theoremstyle{definition}
\newtheorem{dfn}[thm]{Definition}

\theoremstyle{remark}
\newtheorem*{rem}{Remark}

\newcommand{\bdfn}[1][]{\ifthenelse{\equal{#1}{}}{\begin{dfn}}{\begin{dfn}[#1]}}
\newcommand{\edfn}{\end{dfn}}
\newcommand{\bthm}[1][]{\ifthenelse{\equal{#1}{}}{\begin{thm}}{\begin{thm}[#1]}}
\newcommand{\ethm}{\end{thm}}
\newcommand{\bcor}[1][]{\ifthenelse{\equal{#1}{}}{\begin{cor}}{\begin{cor}[#1]}}
\newcommand{\ecor}{\end{cor}}
\newcommand{\blem}[1][]{\ifthenelse{\equal{#1}{}}{\begin{lem}}{\begin{lem}[#1]}}
\newcommand{\elem}{\end{lem}}
\newcommand{\bprop}[1][]{\ifthenelse{\equal{#1}{}}{\begin{prop}}{\begin{prop}[#1]}}
\newcommand{\eprop}{\end{prop}}
\newcommand{\brem}{\begin{rem}}
\newcommand{\erem}{\end{rem}}
\newcommand\bpf[1][]{
\ifthenelse{\equal{#1}{}}{
\begin{proof}}
{\begin{proof}(#1)}
}
\newcommand\epf{\end{proof}}

\newcommand{\eat}[1]{}

\newcommand{\zstroke}{%
  \text{\ooalign{\hidewidth -\kern-.3em-\hidewidth\cr$z$\cr}}%
}

\makeatletter
\newcommand{\iid}[0]{\mathbin{\overset{iid}{\kern\z@\sim}}}

\makeatletter
\DeclareRobustCommand*{\bfseries}{%
  \not@math@alphabet\bfseries\mathbf
  \fontseries\bfdefault\selectfont
  \boldmath
}
\makeatother

\begin{document}

\title[]{Neural Quasiprobabilistic Likelihood Ratio Estimation with Negatively Weighted Data}
\author{Matthew Drnevich*$^a$, Stephen Jiggins*$^b$, Judith Katzy$^b$, Kyle Cranmer$^c$}
\address{$^a$ Physics Department, New York University, \\ \; EMail: mdd424@nyu.edu}
\address{$^b$ Deutsches Elektronen-Synchrotron, DESY, \\ \; EMail: \{stephen.jiggins, judith.katzy\}@desy.de}
\address{$^c$ Physics Department, University of Wisconsin--Madison, \\ \; Data Science Institute, University of Wisconsin-Madison, \\ \; EMail: kyle.cranmer@wisc.edu}
\def\thefootnote{*}\footnotetext{Denotes primary contributors}\def\thefootnote{\arabic{footnote}}

\vspace{10pt}
\begin{indented}
\item[]\today
\end{indented}

\begin{abstract}
Motivated by real-world situations found in high energy particle physics, we consider a generalisation of the likelihood-ratio estimation task to a quasiprobabilistic setting where probability densities can be negative. By extension, this framing also applies to importance sampling in a setting where the importance weights can be negative. The presence of negative densities and negative weights, pose an array of challenges to traditional neural likelihood ratio estimation methods. We address these challenges by introducing a novel loss function. In addition, we introduce a new model architecture based on the decomposition of a likelihood ratio using signed mixture models, providing a second strategy for overcoming these challenges. Finally, we demonstrate our approach on a pedagogical example and a real-world example from particle physics.
\end{abstract}

\section{Introduction}
The likelihood ratio plays an important role in statistics and many domains of science. The Neyman-Pearson lemma states that it is the most powerful test statistic for simple statistical hypothesis testing problems~\cite{NeymanPearsonLemma} or binary classification problems. Likelihood ratios are also key to Monte Carlo importance sampling techniques \cite{Lemi09}. Unfortunately, in many areas of study the probability densities comprising the likelihood ratio are defined by implicit models, and so are intractable to compute explicitly \cite{MC_Implicit_Models}. 

Recently, neural density estimation and neural likelihood ratio estimation have emerged as powerful inference tools in the presence of intractable likelihoods~\cite{LearningLaw_286931,cranmer2016approximating,cranmer2020frontier,Brehmer_2020,hermans2020likelihoodfree}. The ability of these techniques to accurately approximate high dimensional densities and density ratios, solely from synthetically generated datasets is powering a revolution in simulation-based inference~\cite{cranmer2016approximating,cranmer2020frontier} and enables new approaches to importance sampling~\cite{Cranmer2024}, weakly supervised approaches to classification~\cite{metodiev2017classification}, anomaly detection algorithms~\cite{collins2019extending}, and inverse problems such as 'unfolding'~\cite{omnifold}. 

These approaches typically adhere to the standard Kolmogorov axioms of probability theory \cite{ANKolmogorov1933}. In particular, they assume the first axiom: the probability of an event is a non-negative real number. However, there are settings in which synthetically generated data (e.g. Monte Carlo sampling) $\{(\mathbf{x}_i, w_i)\}_{i=1}^{N}$ contains weights that are negative $w_{i}<0$. Setting aside for a moment the exact properties of quasiprobabilities, and why negative weights might be encountered, the goal of this work is to extend neural likelihood ratio estimation into a quasiprobabilistic setting where both negatively weighted samples and negative densities might occur. 

The manuscript is organized as follows. In Section~\ref{sec:QPnNegativeWeights} we introduce notation around importance sampling, which we extend to negative weights and quasiprobabilities. In Section~\ref{sec:NegativeWeightOrigin} we review some reasons that negative weighted events might occur within a scientific setting, in particular in the domain of high-energy particle physics, which motivated this work. Section~\ref{sec:Losses} summarises the challenges posed by negative weights and quasiprobabilities, when estimating quasiprobabilistic likelihood ratios with machine learning methods, and so introduce a new loss function to address these challenges. Section~\ref{sec:SMM} introduces an alternate strategy based on a signed mixture model. We demonstrate our approach on a pedagogical example in Section~\ref{sec:AppToyModel}, and a real world example found in particle physics in Section~\ref{sec:QP-SMEFT}. Finally, we conclude with closing remarks in Section~\ref{sec:Conclusions}.

\subsection{Importance Sampling and Negative Weights}\label{sec:QPnNegativeWeights}

Monte Carlo methods are used ubiquitously to sample from some target distribution $X \sim p_{\rm target}$, which may be a complex scientific simulator. In practice, importance sampling techniques are often used, which involves sampling from some other proposal distribution $\{\mathbf{x}_i\}_{i=1}^{N} \sim p_{\rm proposal}$, and then calculating the importance weights $w_i = p_{\rm target}(\mathbf{x}_i)/p_{\rm proposal}(\mathbf{x}_i)$~\cite{KloekImportanceSampling}. 
From this weighted dataset $\{\mathbf{x}_i, w_i\}_{i=1}^{N}$ one can estimate the expectation of some function $f(\mathbf{x})$ with respect to the target distribution as 
\begin{equation}\label{eq:weighted_average}
\mathbb{E}_{p_{\rm target}}[f(\mathbf{x})] = \mathbb{E}_{p_{\rm proposal}}[w(\mathbf{x})f(\mathbf{x})] \approx \frac{\sum_i w_i f(\mathbf{x}_i)}{ \sum_i w_i }\;. 
\end{equation}
In the context of machine learning, $f(\mathbf{x})$ would be a scalar loss function and the expected risk (with respect to the target distribution) would be calculated in terms of this empirical weighted average. 

More generally, the proposal and target distributions may be defined over a larger set of random variables $X$ and $Z$, where $Z$ denotes the latent variables. In that case, the weight $w_i = p_{\rm target}(\mathbf{x}_i,\mathbf{z}_i)/p_{\rm proposal}(\mathbf{x}_i, \mathbf{z}_i)$ is a deterministic function of the observed and latent variables. However, if we don't observe $Z$, then the deterministic relationship will not be exposed and  $W$ will behave as a random variable drawn from the conditional distribution 
\begin{equation}\label{eq:p_w_given_x}
    p_{\rm proposal}(w|\mathbf{x}) = \int  p_{\rm proposal}(\mathbf{z}|\mathbf{x}) \, \delta\left(w - \frac{p_{\rm target}(\mathbf{x},\mathbf{z})}{p_{\rm proposal}(\mathbf{x}, \mathbf{z})}\right)\, d\mathbf{z} \;.
\end{equation}
With this conditional probability one can show the following key relationship (see Appendix \ref{app:ConditionalWeightProposal})
\begin{equation}\label{eq:p_q_Expw}
    \int w \,p_{\rm proposal}(w|\mathbf{x}) \, dw =  \frac{p_{\rm target}(\mathbf{x})}{p_{\rm proposal}(\mathbf{x})} \; .
\end{equation}
In what follows, we will use $X$ to denote the random variable distributed according to the target distribution, $\tilde{X}$ to denote the random variable distributed according to the proposal distribution, and $W$ to be the importance weight, which is promoted to a random variable when the latent variable $Z$ is not observed. Importance sampling in this generalized setting can be expressed through the following notation 
\begin{equation}\label{eq:weight_definition}
    \mathbb{E}_{{X}}[f(X)] = \mathbb{E}_{\tilde{X},W}[W\,f(\tilde{X})] \; .
\end{equation}
It immediately follows\footnote{See Appendix \ref{app:Notation} for the adopted nomenclature in reference to expectation notation. } that $\mathbb{E}_{\tilde{X},W}[W]=1$.

In a traditional setting, the weights are non-negative, but as we expand upon below, there are real-world settings where one encounters negative weights. At a practical level, the presence of negative weights does not preclude the empirical weighted average in Eq.~\ref{eq:weighted_average} from being an effective estimate of the targeted expectation. However, one can go further. Consider the situation where the conditional expectation of the weights is negative, \textit{i.e.} $\mathbb{E}_{W|\tilde{X}=\mathbf{x}}[W]<0$. This would correspond to a region of negative density $p_{\rm target}(\mathbf{x})<0$, which is not allowed under the standard Kolmogorov axioms \cite{ANKolmogorov1933}, but we can use Eq.~\ref{eq:p_q_Expw} to effectively define a \textit{quasiprobability distribution} that we denote $q(\mathbf{x})$. Note, this quasiprobability distribution still satisfies $\int q(\mathbf{x}) d\mathbf{x} = \mathbb{E}_{\tilde{X},W}[W] = 1$. Alternatively, one might define quasiprobability distributions in a more axiomatic way, which we discuss briefly in Appendix \ref{app:HahnJordan}. 

The introduction of quasiprobabilities also opens the door to a new type of importance sampling where an original set of weighted samples corresponding to $q_{\rm proposal}(\mathbf{x})$ are reweighted to a target quasiprobabilty distributions $q_{\rm target}(\mathbf{x})$ through the quasiprobabilistic likelihood ratio $r(\mathbf{x}) = q_{\rm target}(\mathbf{x})/q_{\rm proposal}(\mathbf{x})$. This is the use-case that originally motivated this work.

\subsection{Negative Weights: Where do we find them?}
\label{sec:NegativeWeightOrigin}
There are multiple settings where negative weights, or negative densities, can emerge in physics. In particle physics they often emerge as artifacts of a calculation scheme used for high-precision predictions. They also emerge as artifacts of non-physical mathematical approximations that are knowingly made in service of some other goal. There are also cases where there are useful mathematical quantities that are not directly observable, yet invite an interpretation of negative density~\cite{Feynman:154856}. Many of these settings involve quantum mechanics, or quantum field theory, but there may be other settings where negative weights or densities also appear (e.g. financial modelling \cite{BurginFinance}).

One example comes from particle physics where two scattering processes quantum mechanically interfere. Let $\mathcal{M}_1(x)$ and $\mathcal{M}_2(x)$ be the complex-valued matrix elements associated to these two processes, where $x$ characterises the final state of the scattering as observed by an experiment. The Born rule\footnote{More formally Gleason's Theorem.} states that the probability density $p_{12}(x) \propto |\mathcal{M}_1(x)+\mathcal{M}_2(x)|^2$ \cite{MR0096113}. This physical probability density is guaranteed to be non-negative. At the same time, it is common to expand the density as follows: $p_{12}(x) \propto |\mathcal{M}_1(x)|^2+2\mathcal{R}\{\mathcal{M}_1(x)\mathcal{M}_2^*(x)\}+|\mathcal{M}_2(x)|^2$, and interpret the three individual terms as components of a mixture model. Note that the middle `interference term' can be negative for some values of $x$; however, isolating the individual terms is not physically meaningful. Nevertheless, this decomposition is sometimes convenient, or deceptively attractive. For instance, a physicist may have a simulation for a physical scenario where only the first process is practical to sample from, leading to $\{x_i\} \sim p_1(x) \propto |\mathcal{M}_1(x)|^2$. One might then want to use importance sampling to re-weight the distribution, so that it follows $p_{12}(x)$, by assigning importance weights $w_i=p_{12}(x_i)/p_1(x_i)$ to each event, rather than sampling from $p_{12}$ directly. It is tempting to use the same approach to model the unphysical interference term, which can lead to negative weights $w_i \propto 2\mathcal{R}\{\mathcal{M}_1(x_i)\mathcal{M}_2^*(x_i)\}/|\mathcal{M}_1(x_i)|^2$.

The example above is analogous to a very common situation that appears in precision measurements in high-energy particle physics when formulated in terms of an effective field theory (EFT) \cite{Grzadkowski_2010}. In this approach, the Standard Model (SM) of particle physics is seen as a low energy remnant of a much more fundamental theory, where the heavy degrees of freedom (fields) representing some new physics are \textit{effectively} modelled as contact interactions with the known SM fields \cite{RevModPhys.96.015006}. These interactions reside above some mass scale $\Lambda$, therefore depending on the types of field interactions, each contact interaction is suppressed according to an appropriate power of $\Lambda$. Schematically, one might describe these interactions as follows, $p_{\textrm{BSM}}(x) \propto |\mathcal{M}_{SM}(x) + \frac{c_6}{\Lambda^2}\mathcal{M}_{6}(x)+ \frac{c_8}{\Lambda^4}\mathcal{M}_{8}(x) + \dots |^2$, where $\mathcal{M}_6$ and $\mathcal{M}_8$ correspond to so-called dimension-six and dimension-eight operators \cite{Brivio_2019}, respectively. Strategically, it makes sense to target the effects of $\mathcal{M}_6$ first since $\mathcal{M}_8$ is suppressed by additional powers of $\Lambda$. However, upon expansion one sees that the term $|\mathcal{M}_{6}|^2$ and $2\mathcal{R}\{\mathcal{M}_{SM}\mathcal{M}^*_8\}$ are both equally suppressed by $\Lambda^4$. Therefore a truncation strategy of $p_{\rm BSM}$ based on $\Lambda$ can not cleanly isolate $\mathcal{M}_6$ without being influenced by the interference effects from $\mathcal{M}_8$. It is possible that a truncation strategy that does not consistently include all of the effects of a given matrix element (operator) may result in a localized region of phase space being assigned (an unphysical) negative-density. Concretely, If one were to sample from $|\mathcal{M}_{SM}|^{2}$ and re-weight the distribution to target $|\mathcal{M}_{SM}|^{2} + 2\frac{c_6}{\Lambda^2}\mathcal{R}\{\mathcal{M}_{SM}\mathcal{M}^*_6\}$, then one may encounter negative weights. This situation is the topic of discussion in Section~\ref{sec:QP-SMEFT}.

A third situation that occurs frequently in the context of particle physics has to do with the treatment of Quantum Chromodynamic (QCD) radiation. The most accurate calculations of QCD mediated processes involve matching matrix element (ME) calculations at beyond leading order accuracy in perturbation theory, with parton showers (PS) that re-sum effects from soft radiation to leading-logarithmic accuracy. The matching process broadly speaking, factorises the phase space into distinct regions based on the number of extra partons in the final state, and the structure of the emission that took place. The ME and PS parts are then combined via merging/matching algorithms, for example infrared subtraction (IR) techniques~\cite{Frixione_1996,Catani_1997,Catani_2002}. This process can yield negatively weighted events~\cite{danziger2021reducing,Frederix_2020}, due to the fact that in some regions of phase space the ME over estimates the scattering probability, therefore the PS has no choice but to subtract from this process by applying a negative weight. This is a very real problem faced by the large experiments at the Large Hadron Collider, which carries substantial penalties in terms of computational resource utilisation \cite{MC-HL-LHC_2021}.


Finally, we  point out that in the Wigner-Weyl transform paradigm the density matrix of a quantum mechanical system is mapped to a position-momentum phase space function known as the Wigner function \cite{PhysRev.40.749}. As a quasiprobabilistic distribution it can be negative in localised regions of the phase space. The Wigner function is a useful object in several areas of quantum physics, including quantum optics and quantum information~\cite{HILLERY1984121,Ferrie_2011}, therefore it is possible that modelling this quasiprobabilistic object with the methods discussed in this paper may be fruitful.

\subsection{Related Work}

The issue of negative weights has been studied in HEP, with particular focus on proton-proton collisions at the Large Hadron Collider (LHC) \cite{LyndonEvans_2008,ATLAS-PMGR-2023-001,MonteCarloChallenges-2021,ATL-PHYS-PUB-2022-026}, where synthetic data generation via Monte Carlo (MC) generation codes are heavily utilised. 
The community has primarily been concerned with removing negative weights from the MC generated samples, due to the computational footprint of large data experiments (e.g. the ATLAS \cite{TheATLASCollaboration_2008}). 

Efforts have either been directed towards reducing the fraction of negative weights produced directly via the MC generation code during sample creation, such as in the \textsc{aMC@NLO} method used in \textsc{MadGraph5} and \textsc{S-Mc@Nlo}/\textsc{Meps@Nlo} methods in \textsc{Sherpa} \cite{Frederix_2020,danziger2021reducing}. Or, as additional post-processing techniques that re-assign the existing weights with an average value for a given localised region of input space \cite{Andersen_2020,Andersen:2021mvw,Nachman_2020,Borisyak_2019}. Regardless, the general approach has been to remove the negative weights from the dataset, thereby bypassing the problem.

\label{sec:Introduction}

\section{Loss Functions for Likelihood Ratio Estimation}
\label{sec:Losses}
We consider the task of learning the quasiprobabalistic likelihood ratio between two true target distributions:
\begin{align}
    r^*_q(\mathbf{x}) \equiv \frac{q(\mathbf{x}|Y=1)}{q(\mathbf{x}|Y=0)}
\end{align}


There are two common approaches for estimating this ratio through minimizing particular loss functions. One approach is to directly estimate the likelihood ratio, using an appropriate loss function, such that the optimal function is the likelihood ratio. For example, these include the Maximum Likelihood Classifier (MLC) and Square Root (SQR) loss functions \cite{PhysRevD.103.116013}. However, these loss functions restrict the output of the function to positive values due to terms involving logarithms and square roots, respectively. Consequently, any function optimized with one of these losses will be incapable of learning negative values for the ratio.

The other approach involves estimating the ratio via the ``ratio trick''\footnote{See Appendix \ref{app:LRE} for a brief introduction to this trick.} \cite{cranmer2016approximating}. This method relies on training a binary classifier with the Binary Cross Entropy (BCE) or Mean-Squared Error (MSE) loss function, since the optimal classifier $s^*(\mathbf{x})$ for these losses is related to the likelihood ratio $r^*(\mathbf{x})$ through the transformations:

\begin{equation}\label{eq:inv_ratio_trick}
    s^*(\mathbf{x}) = \frac{r^*(\mathbf{x})}{1 + r^*(\mathbf{x})} \quad \Longleftrightarrow \quad r^*(\mathbf{x}) = \frac{s^*(\mathbf{x})}{1-s^*(\mathbf{x})}
\end{equation}
Due to the logarithmic terms in the BCE loss function, any classifier trained with this loss is constrained to output values in the range $(0,1)$ which prevents it from expressing values of the ratio that are negative. 

The MSE loss function has no restriction on the output of the model, but other problems arise. According to Equation \ref{eq:inv_ratio_trick}, the transformation between the optimal classifier and the ratio has a pole at $r(\mathbf{x}) = -1$. If the problem of interest takes values $r_q \in [r_{\textrm{min}}, r_{\textrm{max}}]$ with $-1 \in [r_{\textrm{min}}, r_{\textrm{max}}]$ then the classifier must learn to only output values in the range $(-\infty, r_{\textrm{max}}/(1+r_{\textrm{max}})] \cup [r_{\textrm{min}}/(1+r_{\textrm{min}}), \infty)$ which is discontinuous and unbounded.

Even if a classifier is capable of learning discontinuities, there will still be numerical instabilities associated with using MSE. For tasks in which the ratio takes on values in the neighborhood of $-1$, the corresponding classifier output approaches infinity according to Equation \ref{eq:inv_ratio_trick}.  Due to the finite bit representation of floating-point values, optimization algorithms will be susceptible to numerical instabilities, such as overflow. 

We solve these problems by constructing a new loss function with an adjustable transformation between the optimal function and the likelihood ratio. The \textit{pole-adjustable ratio estimation} (PARE) loss function for binary classification is defined as\footnote{Equivalently, $\mathcal{L}_{\textrm{PARE}}(s,y;t_0,t_1) = (1-y)(1-s\cdot t_0)^2 + y(1 - s\cdot t_1)^2$}: 
\begin{equation}
    \mathcal{L}_{\textrm{PARE}}(s,y;\mathbf{t}) \equiv (1-s\cdot t_y)^2
\end{equation}
where $\mathbf{t} = (t_0, t_1)$ has two free parameters $t_0,t_1 \in \mathbb{R}$ and $y \in \{0,1\}$ is the class label. The corresponding optimal function for this loss is:
\begin{equation}\label{eq:pare_loss}
\begin{split}
    s_{\textrm{PARE}}^*(\mathbf{x};\mathbf{t}) &= \argmin_{s}\mathbb{E}_{\tilde{X},Y,W}\left[W\cdot\mathcal{L}_{\textrm{PARE}}(s(\tilde{X}),Y;\mathbf{t})\right] \\
           &= \frac{t_0 q(\mathbf{x}|Y=0) + t_1 q(\mathbf{x}|Y=1)}{t_0^2 q(\mathbf{x}|Y=0) + t_1^2 q(\mathbf{x}|Y=1)} \\
           &= \frac{t_0 + t_1 r_{{q}}(\mathbf{x})}{t_0^2 + t_1^2 r_{{q}}(\mathbf{x})}
\end{split}
\end{equation}
with a derivation shown in Appendix \ref{app:Proofs}.

\begin{figure}
\centering
\includegraphics[scale=0.5]{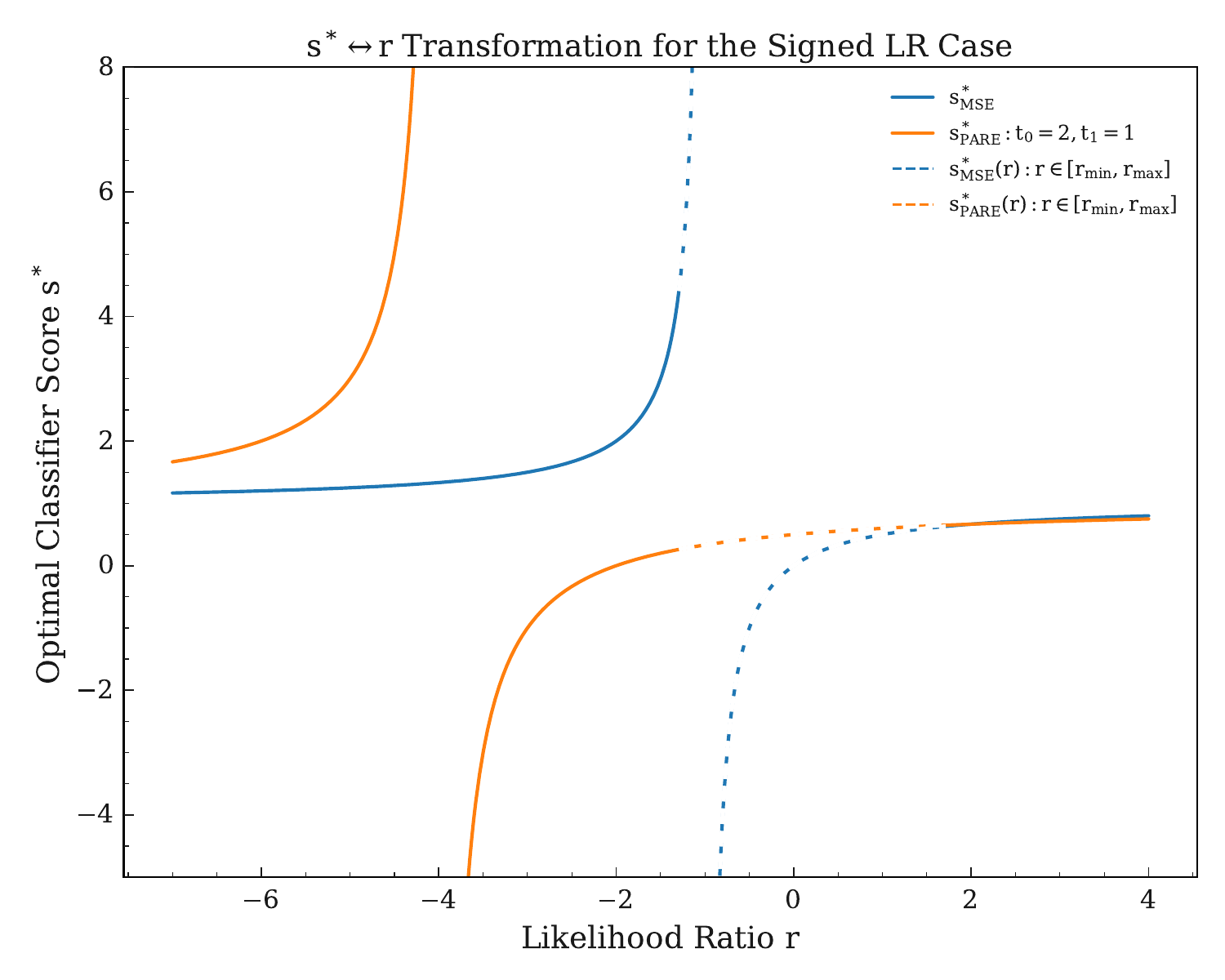}
\caption{Relationship between the likelihood ratio $r$ and the optimal classifier $s^{*}$, for the MSE and PARE loss functions, as given by equations \ref{eq:inv_ratio_trick} \& \ref{eq:pare_loss}, respectively. The dashed line represents the range of $s^{*}$ and $r$, encompassed by the Signed LR dataset described in Section \ref{sec:AppToyModel}.} \label{fig:ratio_transforms}
\end{figure}

This new loss function has no restriction on the range of the learned model for $r_q$ and does not have a pole at $r_{q}(\mathbf{x}) = -1$ in the transformation between the optimal function $s_{\textrm{PARE}}^*$ and $r_q$. The pole is not completely removed, but rather has been shifted to
\begin{equation}
    r_{\textrm{pole}} \equiv -(t_0/t_1)^2
\end{equation}
where there is the freedom to configure $t_0$ and $t_1$ such that the pole is sufficiently far from the expected range of the likelihood ratios for the dataset of interest. If chosen appropriately, then the likelihood ratios for the given dataset will be contained on one side of the pole where the transformation is continuous and stable between $s_{\textrm{PARE}}^*$ and $r_q$.
This transformation for the MSE and PARE losses is shown in Figure \ref{fig:ratio_transforms} for the Signed LR dataset from Section \ref{sec:AppToyModel} with $t_0=2$ and $t_1=1$. For the range of ratios in this dataset, the transformation for MSE covers a discontinuity and an unbounded range, while the transformation for PARE is continuous and the range is compact. In general, the best choice of the parameters may benefit from some hyperparameter search but Appendix \ref{app:loss_considerations} discusses some approaches for choosing $t_0$ and $t_1$. 

Now it is possible to learn the quasiprobability likelihood ratio indirectly by using a universal approximator, such as a neural network, to model $s_{\textrm{PARE}}$ and then applying the ratio trick according to Equation \ref{eq:pare_loss} as long as the model has no restrictions on its output.
Additionally, by shifting the pole far away from the range of ratios for a particular dataset, this loss will not be susceptible to the same numerical instabilities as MSE.

\section{Ratio of Signed Mixtures Model}
\label{sec:SMM}

 

Alternatively, the quasiprobablistic ratio can be estimated by re-framing the problem and decomposing this task into several standard probabilistic likelihood ratio estimation sub-tasks. This decomposition is defined by partitioning the input space according to the sign of the weights, and then each sub-task is to learn the likelihood ratio between two different partitions. We define the probability density functions restricted to each of these partitions as
\begin{equation}
\begin{split}
    p_{w_+}(\mathbf{x}|Y) &\equiv p(\mathbf{x}|Y,W\geq0) \\
    p_{w_-}(\mathbf{x}|Y) &\equiv p(\mathbf{x}|Y,W<0),
\end{split}
\label{eq:ProbabilityPartition}
\end{equation}
such that the overall density functions can be described as a mixture of the positively and negatively-weighted components:
\begin{equation}\label{eq:jordan_decomp}
\begin{split}
    q(\mathbf{x}|Y=y) &= c_y p_{w_+}(\mathbf{x}|Y=y) + (1-c_y) p_{w_-}(\mathbf{x}|Y=y)\\
\end{split}
\end{equation}
where $c_y \in [1,\infty)$ are scalar coefficients, and $q(\mathbf{x}|Y)$ are the quasiprobability density functions\footnote{The partition and this decomposition of $q(\mathbf{x}|Y)$, is motivated by the Hahn-Jordan decomposition of signed probabilistic measures, see Appendix \ref{app:HahnJordan} for details.}. This parametric form ensures that the overall quasiprobability density still integrates to $1$, while the limits on the coefficients $c_y$ ensures that the density of each component corresponds to weights of the correct sign. The optimal coefficients are given by
\begin{equation}\label{eq:coeff_exact}
    c^*_y = \mathbb{E}\left[\left. W\cdot H(W) \right| Y=y\right],
\end{equation}
where $H(w) := [w\geq 0]$ is the Heaviside function. 

Following Refs \cite{cranmer2016approximating,Brehmer:2018eca}, the overall likelihood ratio can then be decomposed as follows:
\begin{equation}\label{eq:rhat}
\begin{split}
    r_{q}(\mathbf{x};\mathbf{c}) &= \frac{q(\mathbf{x}|Y=1)}{q(\mathbf{x}|Y=0)} \\
    &= \frac{c_1 p_{w_+}(\mathbf{x}|Y=1) + (1-c_1)p_{w_-}(\mathbf{x}|Y=1)}{c_0 p_{w_+}(\mathbf{x}|Y=0) + (1-c_0)p_{w_-}(\mathbf{x}|Y=0)} \\
    &= \left[\left(\frac{c_0}{c_1}\right)r^{-1}_{++}(\mathbf{x}) + \left(\frac{1-c_0}{c_1}\right)r^{-1}_{-+}(\mathbf{x})\right]^{-1} \\
    &\quad+ \left[\left(\frac{c_0}{1-c_1}\right)r^{-1}_{+-}(\mathbf{x}) + \left(\frac{1-c_0}{1-c_1}\right)r^{-1}_{--}(\mathbf{x})\right]^{-1}
\end{split}
\end{equation}
where
\begin{equation}
    r_{+-}(\mathbf{x}) \equiv \frac{p_{w_-}(\mathbf{x}|Y=1)}{p_{w_+}(\mathbf{x}|Y=0)}
\end{equation}
and similarly for the other $r_{\pm\pm}(\mathbf{x})$. It is important for the validity of this form of the decomposition that there is sufficient support for each of the subdensity ratios $r_{\pm\pm}$ so that they are well-defined over the data domain.

Estimating this likelihood ratio is therefore reduced to the problem of learning the four separate ``subdensity ratios'' $r_{\pm\pm}(\mathbf{x})$ and the mixture coefficients $\mathbf{c} = (c_0,c_1)$. This is achieved by training separately four ratio estimation models (e.g. CARL \cite{cranmer2016approximating}) on each of the relevant subsets of the data according to Algorithm \ref{alg:subdensities}. This training is done using the absolute values of the weights for each pair of subsets which ensures the learnt sub-density ratio is probabilistic. This enables us to use any traditional ratio estimation techniques for the sub-tasks and can reduce the overall variance incurred during training as demonstrated in Section \ref{subsec:NegWeightStudy}.

After estimating the subdensity ratios, the coefficients $c_y$ need to be estimated as well. According to Equation \ref{eq:coeff_exact} a good estimate using the training dataset for each class is:
\begin{equation}\label{eq:coeff_estimate}
    \hat{c}_y = \frac{\sum_{i, w_i \geq 0} w_i}{\sum_{i} w_i}
\end{equation}
where each $w_i$ is from the dataset with class label $y$. A model $\hat{r}_q(\mathbf{x};\hat{\mathbf{c}})$ constructed according to Equation \ref{eq:rhat} with learnt sub-density ratios $\hat{r}_{\pm\pm}$ and coefficients $\hat{\mathbf{c}}$ estimated according to Equation \ref{eq:coeff_estimate} is referred to as a \textit{Ratio of Signed Mixtures Model} or RoSMM.

\begin{algorithm*}[t]
\caption{Training the sub-density ratio estimators}\label{alg:subdensities}
\begin{algorithmic}
\For{$y$ \textbf{in} $\{0,1\}$}
    \State Load $\mathcal{D}^y$ \Comment{The dataset from each class $y$}
    \State $\mathcal{D}^y_{+} \gets \{(\mathbf{x},w) : (\mathbf{x},w) \in \mathcal{D}^y,  w\geq 0\}$
    \State $\mathcal{D}^y_{-} \gets \{(\mathbf{x},|w|) : (\mathbf{x},w) \in \mathcal{D}^y,  w < 0\}$
\EndFor
\For{$k_0$ \textbf{in} $\{+, -\}$}
    \For{$k_1$ \textbf{in} $\{+, -\}$}
        \State $N \gets \text{Min}(\text{Size}(\mathcal{D}^0_{k_0}), \text{Size}(\mathcal{D}^1_{k_1}))$
        \For{$y$ \textbf{in} $\{0,1\}$}
            \State Draw $N$ samples without replacement $\{(\mathbf{x}_i,w_i)\}_{i=1}^N \sim \mathcal{D}^y_{k_y}$
            \State $S \gets \frac{1}{N}\sum_{i=1}^{N} w_i$
            \State $\mathcal{D}^y \gets \{((\mathbf{x}_i,w_i / S), y)\}_{i=1}^{N}$ 
            \Comment{Normalize weights and assign target}
        \EndFor
        \State $\mathcal{D} \gets \mathcal{D}^0 \cup \mathcal{D}^1$
        \State Train a standard binary classifier $\hat{s}_{(k_0,k_1)}$ using dataset $\mathcal{D}$ \Comment{Do this in parallel}
    \EndFor
\EndFor
\end{algorithmic}
\end{algorithm*}

\subsection{Further Optimizing the Ratio of Signed Mixtures Model}\label{subsec:optimizing_mixture}

If the subdensity ratio estimates  $\hat{r}_{\pm\pm}(\mathbf{x})$ are perfect, then the construction according to Equation \ref{eq:rhat} is a good estimate of the likelihood ratio $r_q$. However, in practice these subdensity ratio estimates will have some error and additional optimization to improve the model is possible.


One approach is to return to the ratio trick, and define the optimal classifier ($s^{*}$) 
in terms of the learnt RoSMM ($\hat{r}_{q}$). This transformation between the classifier and likelihood ratio is characteristic to a given loss function. Then, given the classifier and loss function pairing, one can optimize the coefficients according to:
\begin{equation}
    \hat{\mathbf{c}}_{\mathcal{L}} = \argmin_{{\mathbf{c}} \in \mathbb{R}^2}\mathbb{E}_{\tilde{X},Y,W}\left[W\cdot\mathcal{L}\left(\hat{s}(\tilde{X};\mathbf{c}),Y\right)\right],
\end{equation}
where $\hat{s}(\mathbf{x};\mathbf{c})$ is the estimate for the classifier by applying the ratio trick transformation to $\hat{r}_q(\mathbf{x};\mathbf{c})$ for the given loss function.
However, as mentioned in Section \ref{sec:Losses} there are various problems with traditional loss choices that can make this procedure ill-defined; e.g. the ratio trick for MSE contains a pole at $\hat{r}_{q}(\mathbf{x};\mathbf{c}) = -1$. In order to optimize the model in this way, we must use the $\mathcal{L}_{\textrm{PARE}}$ loss described in Section \ref{sec:Losses}. This enables two different strategies for improving the RoSMM model:

The first approach is to use the RoSMM model for the likelihood ratio to estimate the optimal function $s_{\textrm{PARE}}^*$ according to Equation \ref{eq:pare_loss} and then tune the coefficients $(c_0,c_1)$ in the model by minimizing the loss. More precisely, we plug in our initial estimate $\hat{r}_{q}(\mathbf{x};\hat{\mathbf{c}})$ composed of the four pre-trained subdensity ratio models with coefficients initialized according to Equation \ref{eq:coeff_estimate} and minimize $\mathcal{L}_{\textrm{PARE}}$ to optimize the coefficients with every other component in the model fixed: 
\begin{equation} \label{eq:mix_loss1}
\begin{split}
    \hat{\mathbf{c}}_{\textrm{PARE}} &= \argmin_{\mathbf{c} \in \mathbb{R}^2}\mathbb{E}_{\tilde{X},Y,W}\left[W\cdot\mathcal{L}_{\textrm{PARE}}\left(\frac{t_0 + t_1\hat{r}_{q}(\tilde{X};\mathbf{c})}{t_0^2 + t_1^2\hat{r}_{q}(\tilde{X};\mathbf{c})},Y;\mathbf{t}\right)\right]
\end{split}
\end{equation}
A model trained using this approach is referred to as a Ratio of Signed Mixtures Model with coefficient tuning, or $\textrm{RoSMM}_c$.

As an additional option to achieve better performance of the model, it is also possible to use the same approach as the $\textrm{RoSMM}_c$ model but let the subratio models be updated as well. This removes some theoretical understanding of the composition of the overall model with the potential benefit of achieving better overall performance. The pre-training of the subratio models is important to ensure the overall model is initialized in a favorable region of the loss landscape before jointly optimizing $\mathbf{c}$ and $r_{\pm\pm}$. 
A model trained using this approach is referred to as a Ratio of Signed Mixtures Model with full tuning, or $\textrm{RoSMM}_r$. In either setup, one can use Algorithm \ref{alg:mixture} to find $\hat{r}_q(\mathbf{x};\hat{\mathbf{c}})$.

\begin{algorithm*}[t]
\caption{Optimizing the Ratio of Signed Mixtures Model}\label{alg:mixture}
\begin{algorithmic}
\State $\mathbf{c} \gets \{0,0\}$
\State $\mathbf{w}_{\textrm{sum}} \gets \{0,0\}$
\For{$y$ \textbf{in} $\{0,1\}$} \Comment{Initialize the mixture coefficients}
    \State Load $\mathcal{D}^y$ \Comment{The datasets from each class $y$}
    \For{$(\mathbf{x}_i,w_i)$ \textbf{in} $\mathcal{D}^y$} \Comment{Initialize the coefficients}
        \State $\mathbf{w}_{\textrm{sum}}[y] \gets \mathbf{w}_{\textrm{sum}}[y] + w_i$
        \If{$w_i \geq 0$}
            \State $\mathbf{c}[y] \gets \mathbf{c}[y] + w_i$
        \EndIf
    \EndFor
    \State $\mathbf{c}[y] \gets \mathbf{c}[y] / \mathbf{w}_{\textrm{sum}}[y]$
\EndFor
\State Load sub-density models $\hat{s}_{(+,+)}, \hat{s}_{(+,-)}, \hat{s}_{(-,+)}, \hat{s}_{(-,-)}$
\For{$k$ \textbf{in} $\{(+,+), (+,-), (-,+), (-,-)\}$}
    \State $\hat{r}_{k}(\mathbf{x}) \gets (1 - \hat{s}_{k}(\mathbf{x})) /     \hat{s}_{k}(\mathbf{x})$ \Comment{$r_{k}(\mathbf{x}) = p_{k[0]}(\mathbf{x}|Y=0) / p_{k[1]}(\mathbf{x}|Y=1)$}
    \If{\textbf{not} $\text{full tuning}$}
        \State Turn off gradient updates for $\hat{r}_{k}$
    \EndIf
\EndFor
\State $r(\mathbf{x}) \gets \left[\frac{\mathbf{c}[0]}{\mathbf{c}[1]}\hat{r}_{(+,+)}(\mathbf{x}) + \frac{(1-\mathbf{c}[0])}{\mathbf{c}[1]}\hat{r}_{(-,+)}(\mathbf{x})\right]^{-1} + \left[\frac{\mathbf{c}[0]}{(1-\mathbf{c}[1])}\hat{r}_{(+,-)}(\mathbf{x}) + \frac{(1-\mathbf{c}[0])}{(1-\mathbf{c}[1])}\hat{r}_{(-,-)}(\mathbf{x})\right]^{-1}$
\State Choose $t_0, t_1$ \Comment{Determines the pole at $r(\mathbf{x}) = -(t_0/t_1)^2$}
\State $s(\mathbf{x}) \gets (t_0 + t_1 r(\mathbf{x})) / (t_0^2 + t_1^2 r(\mathbf{x}))$
\State $\mathcal{D} \gets \{((\mathbf{x},w),t_0) : (\mathbf{x},w) \in \mathcal{D}^0\} \cup \{((\mathbf{x},w),t_1) : (\mathbf{x},w) \in \mathcal{D}^1\}$
\State Minimize $\frac{1}{N}\sum_i^N w_i\left(1 - s(\mathbf{x}_i)t_i\right)^2$ over $((\mathbf{x}_i,w_i),t_i) \in \mathcal{D}$
\end{algorithmic}
\end{algorithm*}

\section{Pedagogical Example: Signed Gaussian Mixture Models}
\label{sec:AppToyModel}
In this section the proposed quasiprobabilistic Ratio of Signed Mixtures Model for likelihood ratio estimation is applied to an `analytic toy' problem. The goal is to use the likelihood ratio estimate as an importance sampling based re-weighting technique, that maps expectations (and histograms) from one distribution to another. In this toy example, the reference distribution is a two dimensional (2D) nonnegative Gaussian mixture model, and the target distribution is a 2D signed Gaussian mixture model. 


\subsection{Signed Gaussian Mixture}

The quasiprobability density functions used in this example are constructed as a mixture of two 2D Gaussians\footnote{A review of 2D Gaussian distributions is given in the Appendix section \ref{app:2D_gaussians}.} in order to naturally introduce the possibility of negative weights. Specifically, a mixture is constructed with the coefficient $c \in (1,\infty)$:

\begin{align}
\label{eq:GaussMixModel}
    q(x,y;c,\sigma_1,\sigma_2) = c\overbrace{p_1(x,y;\sigma_1)}^{\mathclap{\mathrm{G_1}}} + (1-c)\underbrace{p_2(x,y;\sigma_2)}_{\mathclap{\mathrm{G_2}}}
\end{align}
where the two density functions $p_1,p_2$ can be thought of as corresponding to the random variables
\begin{align*}
    G_1 &\sim \mathcal{N}(0, \sigma_{1}^2 I_{2}), \\
    G_2 &\sim \mathcal{N}(0, \sigma_{2}^2 I_{2}),
\end{align*}
respectively. In this construction, one can also change variables to polar coordinates and get independent distributions of $r$ and $\phi$, such that

\begin{align*}
    q(x,y;c,\sigma_1,\sigma_2) &= \left[cp_1(r;\sigma_1) + (1-c)p_2(r;\sigma_2)\right]p(\phi) \\
    &\equiv q(r;c,\sigma_1,\sigma_2)p(\phi),
\end{align*}
where:
\begin{align}
\label{eq:PolarGaussMixModel}
    q(r;c,\sigma_1,\sigma_2) \equiv c\overbrace{p_1(r;\sigma_1)}^{\mathclap{\mathrm{G_1}}} + (1-c)\underbrace{p_2(r;\sigma_2)}_{\mathclap{\mathrm{G_2}}}.
\end{align}
In the polar coordinate representation, essentially all characteristic information about the distribution is encoded in the radial distribution. 

\subsection{Data Generation}
\label{sec:ToyGaussianDataGen}



The sampling procedure is best described with the realisation of random variables, in which the input space $\tilde{X}$ and weights $W$ are sampled according to:
\begin{align}
    B  &\sim \mathrm{Bernoulli}\left( \frac{c}{2c-1} \right), \\
    (\tilde{X}, W) &\sim (B\cdot G_1+(1-B)\cdot G_2,\ 2B-1)
\end{align}

such that the weights take on values $\pm 1$. Two test cases are considered: one where the true likelihood ratio is purely nonnegative (Nonnegative LR) and one where the true likelihood ratio is signed (Signed LR). In both test cases, a single reference dataset is used which has a nonnegative density function, referred to as the reference sample, it has supervisory labels Y=0:
\begin{equation*}
    q^{\mathrm{ref.}}(x,y; 4/3, 2.5, 2.3).
\end{equation*}

For the target distributions two datasets are generated, one for each test case:
\begin{align*}
    \mathrm{Nonnegative \, LR:}& \, q^{\mathrm{target}}(x,y; 2, 2.5, 1.42) \\
    \mathrm{Signed \, LR:}& \, q^{\mathrm{target}}(x,y; 2, 2.5, 1.2),
\end{align*}
where the only difference between the two samples is $\sigma_{2} = 1.42 \rightarrow 1.2$. Table \ref{tab:toy_settings} summarises the parameter settings for both the reference and target distributions, with the above key difference highlighted in bold.

\begin{table}[h]
\caption{Signed Gaussian Mixture Dataset Settings}\label{tab:toy_settings}
\centering
\begin{tblr}{colspec={m{4cm}||>{\centering\arraybackslash}m{1.25cm}|>{\centering\arraybackslash}m{1.25cm}|>{\centering\arraybackslash}m{1.25cm}}}
 \hline
 \SetCell[c=4]{c}{\textbf{Nonnegative Likelihood Ratio}} \\
 \hline
 Dataset Class & $c_y$ & $\sigma_1$ & $\sigma_2$\\
 \hline
 Reference (Y=0) & 4/3 & 2.5 & 2.3 \\
 Target (Y=1) & 2 & 2 & \textbf{1.42} \\
 \hline
 \SetCell[c=4]{c} \textbf{Signed Likelihood Ratio} \\
 \hline
 Dataset Class & $c_y$ & $\sigma_1$ & $\sigma_2$\\
 \hline
 Reference (Y=0) & 4/3 & 2.5 & 2.3 \\
 Target (Y=1) & 2 & 2 & \textbf{1.2} \\
 \hline
\end{tblr}
\end{table}

\begin{figure}[t]
\centering
\includegraphics[scale=0.45]{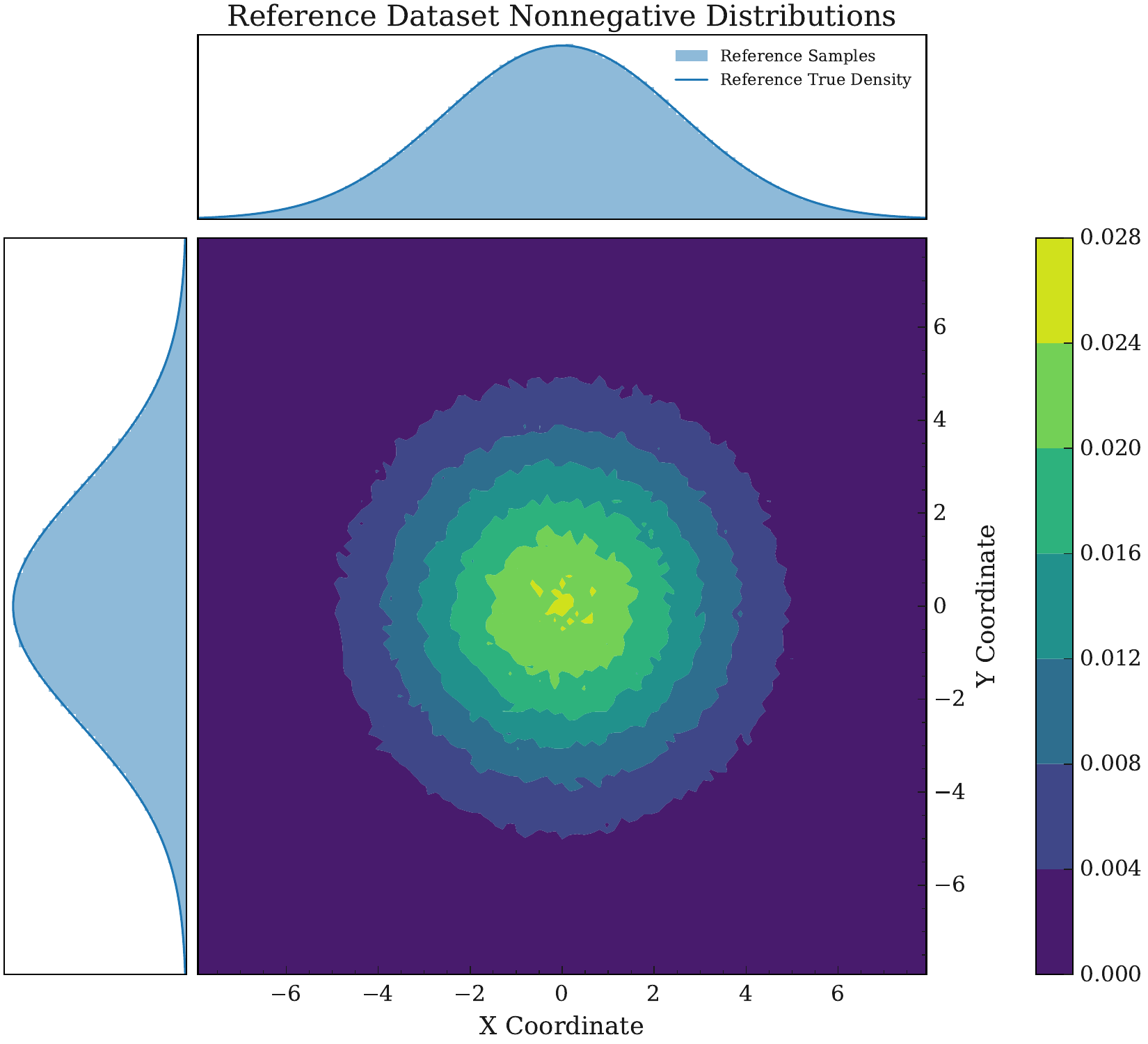}
\caption{Signed Gaussian mixture reference distribution for the Nonnegative \& Signed LR test case, using the settings in Table \ref{tab:toy_settings}. The one-dimensional distributions on the sides represent the marginal $X$ and $Y$ coordinate distributions.}\label{fig:toy_ref_distributions}
\end{figure}

\begin{figure}[h]
\centering
\includegraphics[scale=0.45]{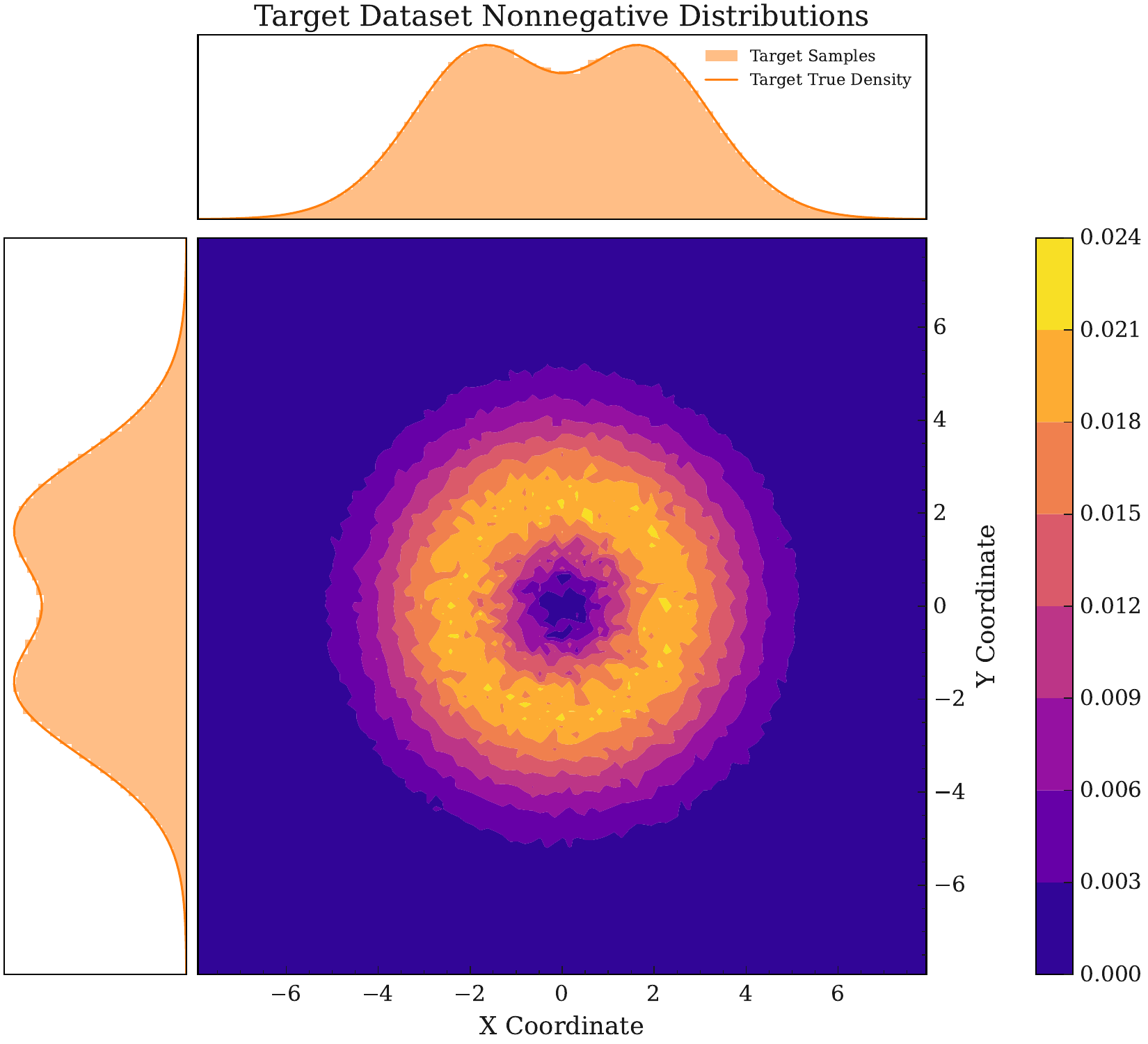}
\caption{Signed Gaussian mixture target distribution for the Nonnegative LR test case, using the settings in Table \ref{tab:toy_settings}. The one-dimensional distributions on the sides represent the marginal $X$ and $Y$ coordinate distributions.}\label{fig:toy_prob_target_distributions}
\end{figure}

\begin{figure}[h]
\centering
\includegraphics[scale=0.45]{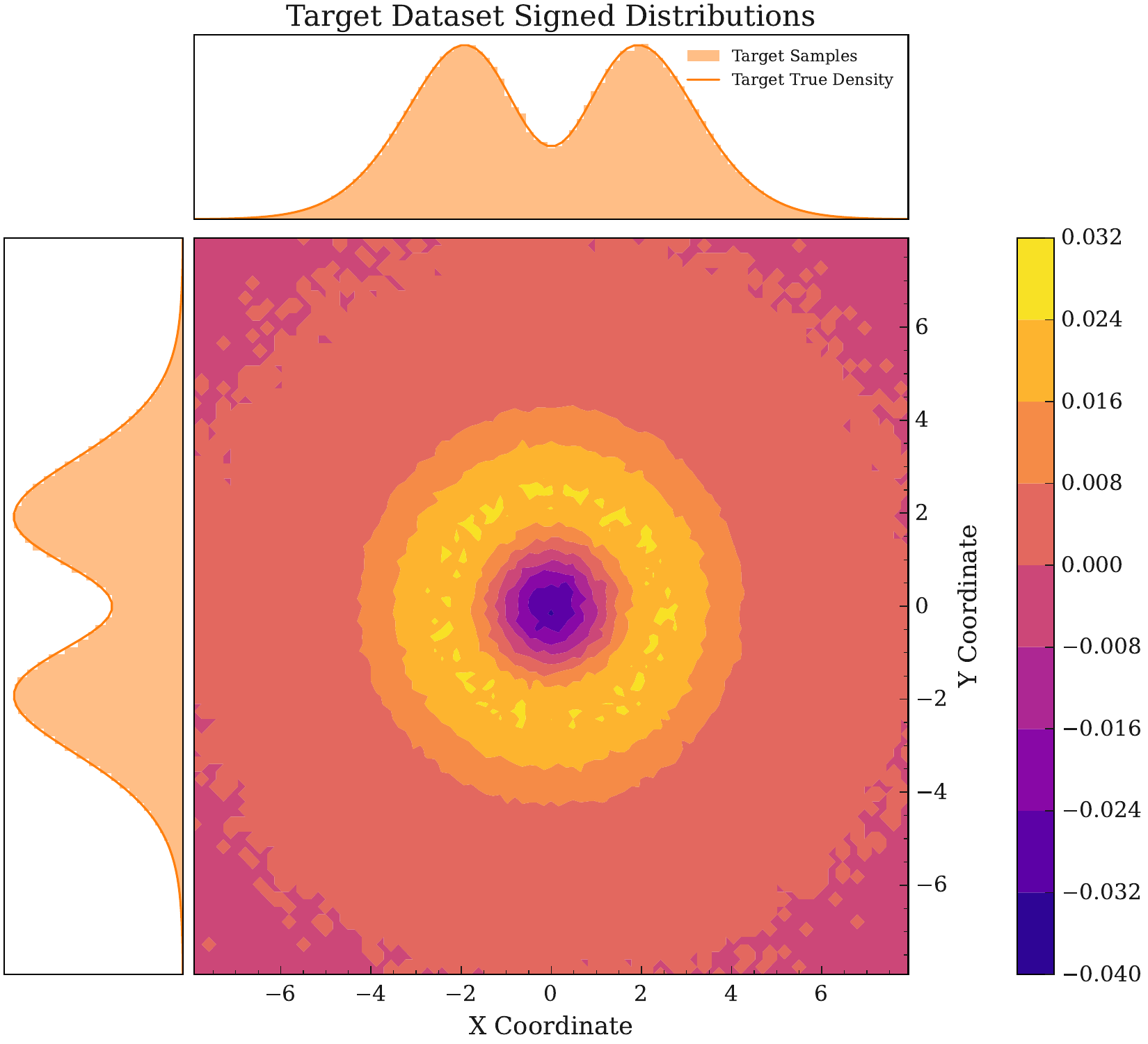}
\caption{Signed Gaussian mixture target distribution for the Signed LR test case, using the settings in Table \ref{tab:toy_settings}. The one-dimensional distributions on the sides represent the marginal $X$ and $Y$ coordinate distributions.}\label{fig:toy_quasi_target_distributions}
\end{figure}

Figures \ref{fig:toy_ref_distributions}-\ref{fig:toy_quasi_target_distributions} show the reference and two target dataset distributions in the $x-y$ domain, with the corresponding 1-dimensional marginal projections. In comparing Figure \ref{fig:toy_prob_target_distributions} against Figure \ref{fig:toy_quasi_target_distributions}, a high negative density region can be observed around the origin $(x,y) = (0,0)$, resulting in a target dsitribution that violates Kolmogorov's first axiom. This is the consequence of the smaller standard deviation parameter $\sigma_{2}$ of the negatively-signed mixture component $G_{2}$ in the Signed LR target distribution relative to the Nonnegative LR target distribution. 

\begin{figure*}[t]
\centering
\includegraphics[scale=0.45]{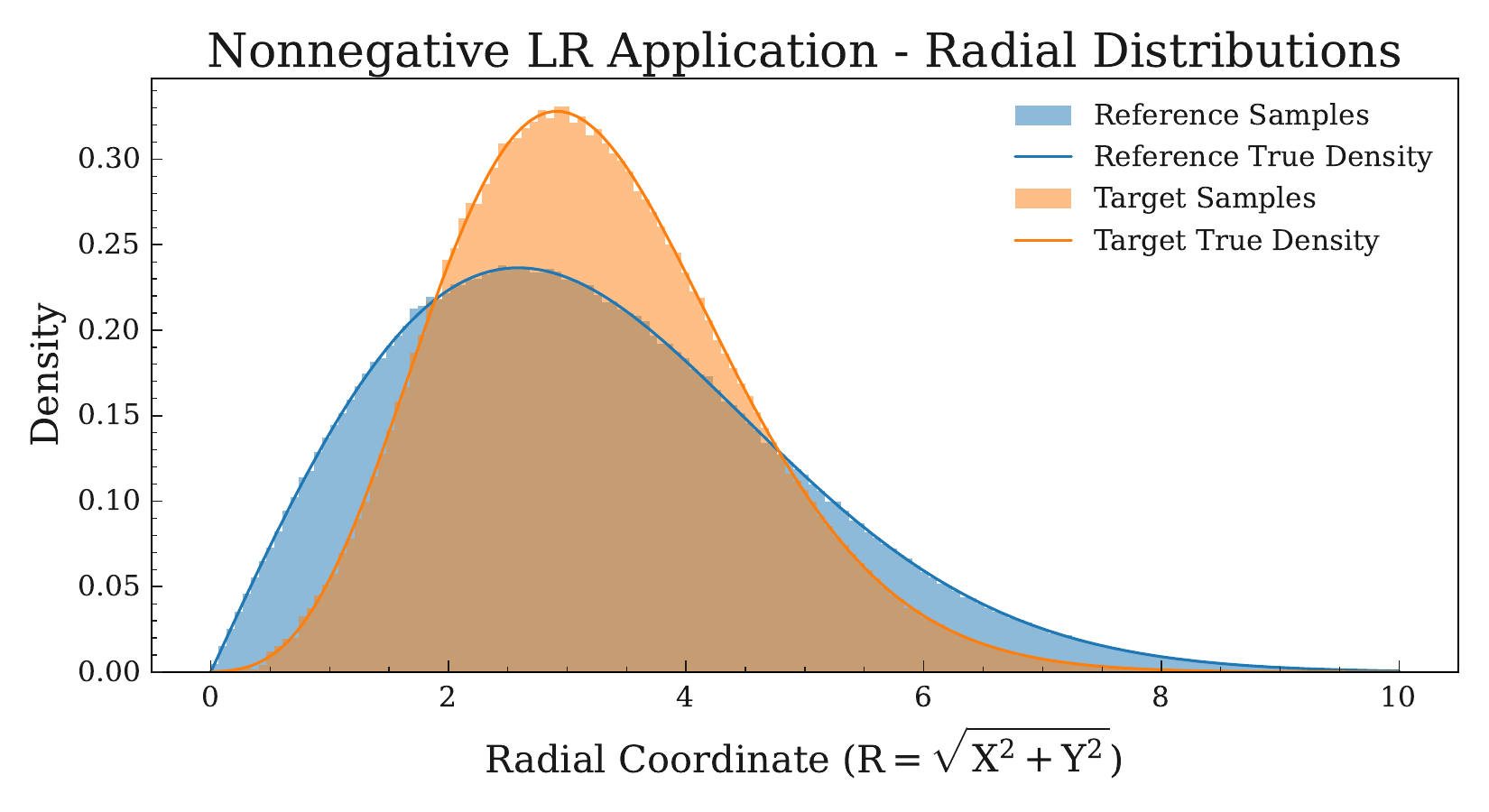}
\includegraphics[scale=0.45]{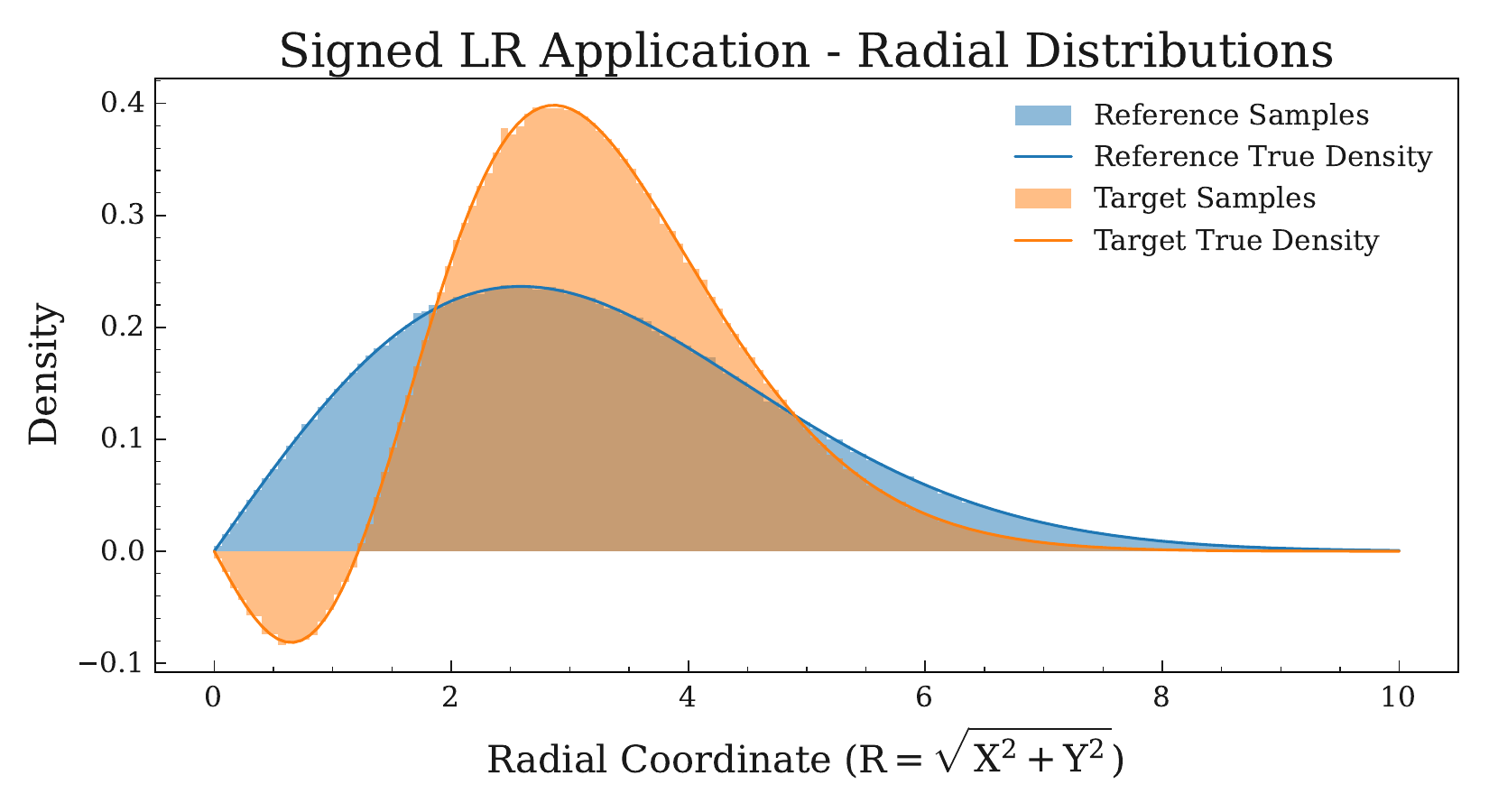}
\caption{Reference and target radial distributions for the Nonnegative and Signed LR test cases, using the settings in Table \ref{tab:toy_settings}.}\label{fig:toy_prob_radial_distributions}
\end{figure*}

For the applications studied here, all information is essentially encoded in the radial coordinate, as outlined by Equation \ref{eq:PolarGaussMixModel} (see Appendix \ref{app:2D_gaussians} for more details). The distributions for both test cases can be seen in Figure \ref{fig:toy_prob_radial_distributions}. The negative density region is clearly visible in the Signed LR test case for small radial values, unlike the $x/y-\textrm{projections}$ in Figure \ref{fig:toy_quasi_target_distributions}. This phenomenon is representative of a common practical problem in data science fields involving high dimensional data spaces, in which the summary/observational basis of the data conceals potential quasiprobabislitic properties. Therefore, for this example it is sufficient to focus on the radial distribution to understand model performance since it captures all of the data information.

For each application, $2\cdot 10^6$ samples were generated for the training data, $6\cdot 10^5$ samples for validation, and $1.4 \cdot 10^6$ for testing. The training and validation data is generated according to Algorithm \ref{alg:camel} in Appendix \ref{sec:appendix_algorithms}. For the Nonnegative LR distributions, a test dataset is generated by numerically solving the quantile function (i.e. inverse transform sampling) such that all data points have a weight of $+1$. For the Signed LR distributions, a test dataset is generated in the same manner as the training and validation data since it is not possible to sample directly from a quasiprobability distribution with regions of negative density.

\subsection{Machine Learning Setup}

All models were constructed and trained using PyTorch \cite{NEURIPS2019_9015} with a batch size of 256 and the Adam optimizer with a learning rate of $10^{-4}$. Training is stopped after twenty sequential epochs when the validation loss is greater than the lowest validation loss across all epochs. An epoch is defined as either the entire training dataset or $10^5$ training samples, whichever is smaller. The same settings were used for both test cases - the Nonnegative LR and the Signed LR tests. The number of parameters in each neural model have roughly the same number of total parameters, thereby limiting each neural likelihood ratio estimator model to the same level of expressibility. The architectures and model-specific settings are:
\begin{itemize}
    \item MLP Classifier
    \begin{itemize}
        \item Inputs: 2
        \item 2 Hidden Layers: 64 nodes each (Linear with ReLU)
        \item Output: 1 (Sigmoid)
        \item Loss: Binary Cross Entropy
    \end{itemize}
    \item Ratio of Signed Mixtures Model - composed of four sub-ratios and two mixture coefficients. Each sub-ratio has the following settings:
    \begin{itemize}
        \item Inputs: 2
        \item 2 Hidden Layers: 32 nodes each (Linear with ReLU)
        \item Output: 1 (Sigmoid)
        \item Loss: Binary Cross Entropy
    \end{itemize}
\end{itemize}

Three flavours of the ratio of signed mixtures model are trained:
\begin{itemize}
\item $\textrm{RoSMM}$ - Ratio of Signed Mixtures Model using equations \ref{eq:rhat} \& \ref{eq:coeff_estimate}
\item $\textrm{RoSMM}_{c}$ - Ratio of Signed Mixtures Model with coefficient tuning according to equation \ref{eq:mix_loss1}
\item $\textrm{RoSMM}_{r}$ - Ratio of Signed Mixtures Model with coefficient and sub-density ratio tuning
\end{itemize}
where in all three cases the pre-trained sub-density ratio models according to Algorithm \ref{alg:subdensities}, are the same. The parameters $t_0 = 25619$ and $t_1 = 58$ were chosen such that the pole for the loss $\mathcal{L}_{\textrm{PARE}}$ is at $r_{\textrm{pole}}(\mathbf{x}) \approx -195105$ which is sufficiently far from any true ratio expected for the data distributions. Additionally, these specific values of $t_0$ and $t_1$ were chosen to improve stability of the loss function $\mathcal{L}_{\textrm{PARE}}$ during optimization\footnote{See Appendix \ref{app:loss_considerations} for a discussion on choosing these parameters.}.


\subsection{Model Optimization and Negative Weights} \label{subsec:NegWeightStudy}

Before showing results for the different models and applications, we discuss the impact of negatively weighted data on training a model in the standard probabilistic setting.
Training a model on weighted data is known to have a detrimental impact on the optimization procedure, but this work explores the effect of negative weights in particular.


As demonstrated in Section \ref{sec:QPnNegativeWeights}, using a Monte Carlo importance sampling methodology yields weighted data $(\mathbf{x}_i,w_i) \sim (\tilde{X},W)$. In contrast, using a direct sampling strategy yields ``unweighted'' data $\mathbf{x}_{i} \sim X$.
When optimizing parameterized functions $s(\mathbf{x};\theta)$, with configurable parameters $\theta$, using stochastic gradient descent (SGD) \cite{alma9998754279107871,bottou-91c}, each parameter update $\theta^{t+1}$ at a time-step $t$ is sample dependent, and so has a variance associated with the sampling methodology used.
The difference in the variance of the parameter updates between these two sampling strategies is given by:

\begin{equation}\label{eq:variance_gradient}
\begin{split}
    &\text{Var}_{\tilde{X},Y,W}\big(\theta^{t+1}\big) - \text{Var}_{X,Y}\big(\theta^{t+1}\big) \\
    &\quad = \frac{\gamma^2}{N_{\textrm{batch}}} \mathbb{E}_{\tilde{X},Y,W}\left[(W^2-W) \cdot \left(\nabla_\theta\mathcal{L}(s(\tilde{X};\theta), Y)\bigr|_{\theta = \theta^t}\right)^2\right]
\end{split}
\end{equation}

where $\gamma$ is the SGD learning rate, $N_{\textrm{batch}}$ is the batch size, and $\mathcal{L}(s,y)$ is a loss function. Consequently, this difference is systematically increased by negative weights because wherever $W<0$ it follows that $W^2 - W > 0$. Additionally, this difference grows with the variance of the weights since $\mathbb{E}\left[W^2 - W\right] = \text{Var}(W)$ for class-balanced datasets. 

Although these weights do not pose an issue to learning the optimal function, the extra variance introduced by the negative weights hinders the convergence of algorithms like SGD and can impact the performance of the learned function.
To understand the effect of negatively weighted data on the optimisation task for probabilistic classifier-based likelihood ratio estimation problems, a decorrelated study of the negative weight fraction $\eta$, and weight variance $\sigma_{w}$ is performed.

This is achieved by generating samples with the same settings as the Nonnegative LR test case, but with the key difference that data realisations are sampled using the inverse transform sampling method \cite{Devroye1986}. As a result, all weights for this dataset are equal to $+1$ (unity). For each of the reference and target distributions in the dataset, 8 million samples are generated, with a split of 55\%, 15\%, and 30\% for training, validation, and testing, respectively. This ``unweighted'' dataset, of 16 million entries, is only used for the study in this section. 

The data instances in the training and validation sets are then assigned weights independent of $\tilde{X}$, thereby formulating a weighted dataset.
The weight random variable $W$ is parameterized by the variables $\eta$ and $\sigma_{w}$:
\begin{align}
    \eta &\equiv P(W < 0) \\
    \sigma_{w} &\equiv \sqrt{\text{Var}(W)}
\end{align}
with the limitations that $\sigma_{w} \in \mathbb{R}_{\geq 0}$ and $\eta \in \left(0, \frac{\sigma_{w}^2}{1+\sigma_{w}^2}\right)$. The weight distribution is then described by
\begin{align}
    W &\sim 1 + \frac{\sigma_{w}}{\sqrt{\eta(1-\eta)}}\left(\eta - \text{Bern}(\eta)\right)
\end{align}
which has the following properties
\begin{align*}
    \mathbb{E}\left[W\right] &= 1 \\
    \mathbb{E}\left[W^2 - W\right] &= \text{Var}(W) = \sigma_{w}^2 \\
    P(W<0) &= \eta
\end{align*}
By construction this setup will generate a sample with non-unit weights, but with a fixed $\eta$ and $\sigma_{w}$. Consequently, the impact of the parameters $\eta$ and $\sigma_{w}$ can then be studied in a decorrelated fashion.

A dense grid of 1000 individual datasets are generated in the $\eta-\sigma_{w}$ parameter space plane using the above strategy, with 50 uniformly distributed values for $\sigma_{w} \in [1,5]$, and 20 uniformly distributed values for $\eta \in \left[0.02, \frac{\sigma_w^2}{1+\sigma_w^2}\right]$. An MLP classifier and a $\textrm{RoSMM}_r$ model were then trained using the weighted dataset at each $(\eta,\sigma_{w})$ point, with the same model construction as outlined in the previous subsection. After training, the models were evaluated on the ``unweighted'' reference and target test datasets in order to determine performance in terms of importance-based reweighting.

\begin{figure}[t] 
\centering
\includegraphics[scale=0.47]{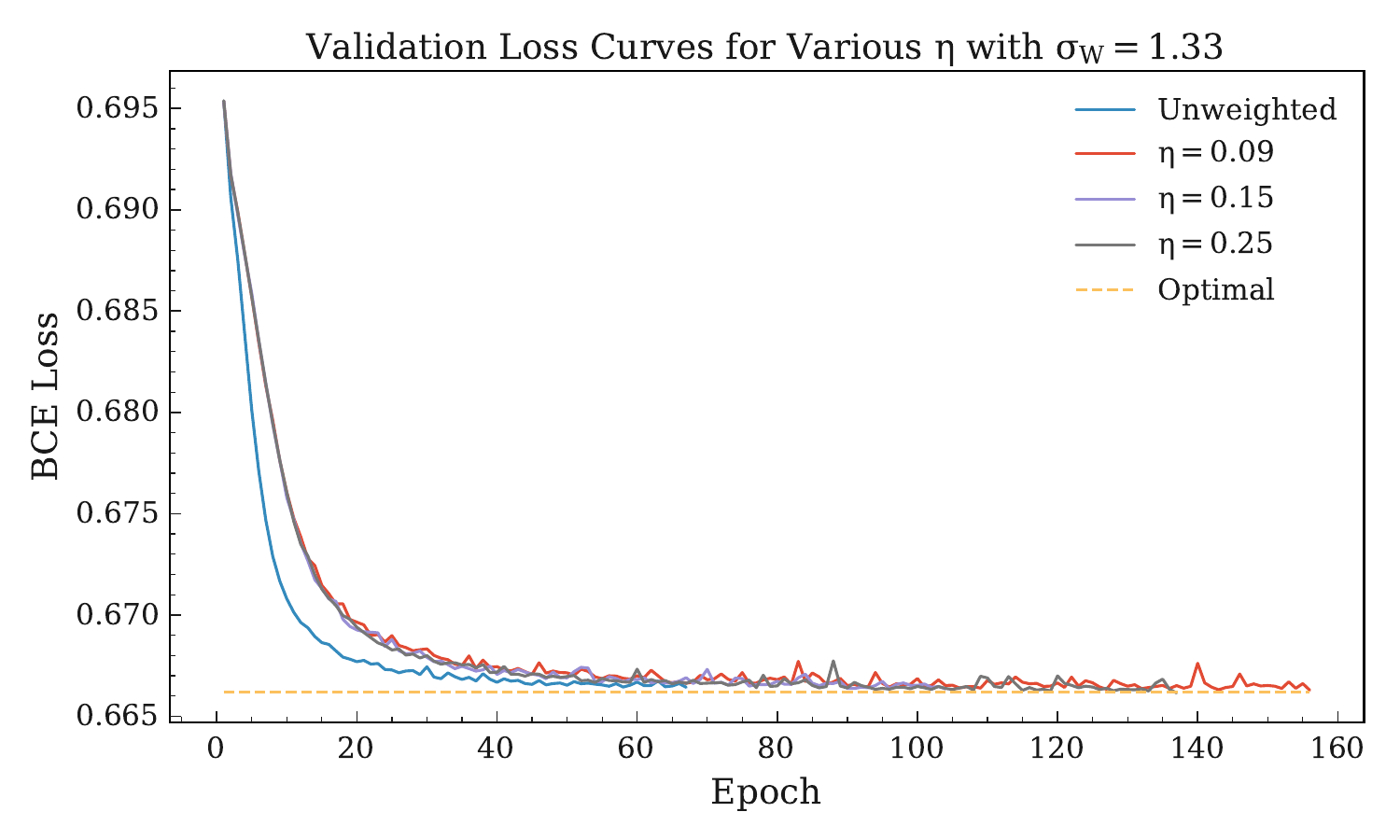}
\caption{Comparison of some validation loss curves for MLP classifers trained on the unweighted and weighted datasets for different parameter settings. The analytical optimal loss is also added for reference.}\label{fig:Validation_loss_curves}
\end{figure}

\begin{figure}[t]
\centering
\includegraphics[scale=0.45]{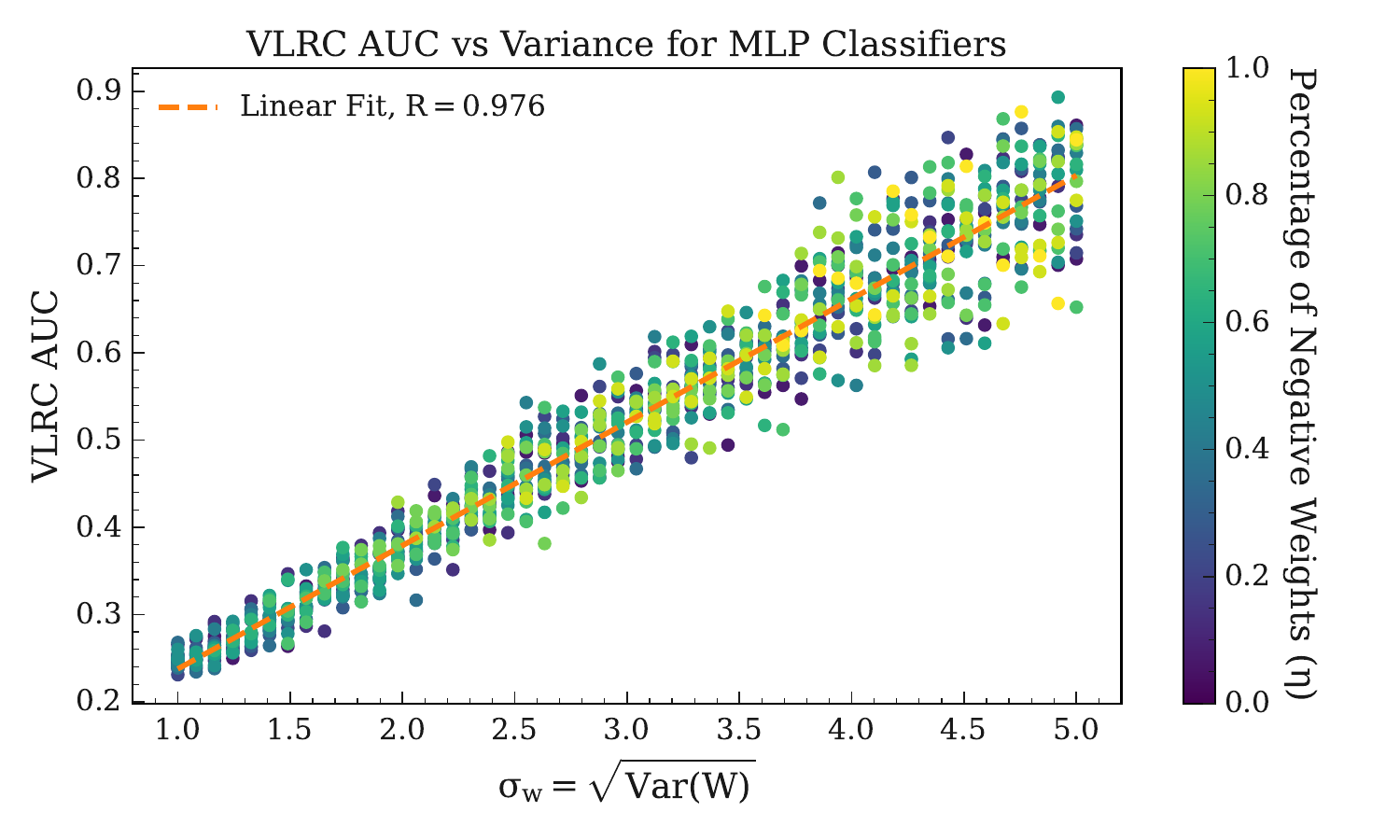}
\caption{Scatterplot of the Area Under the Curve for the Validation Loss Residual Curves (VLRC AUC) vs the standard deviation of the weights $\sigma_{w}$ with coloring based on the fraction of negative weights $\eta$. Each datapoint represents an MLP classifier trained for a particular choice of $(\sigma_{w}, \eta)$. The best fit line is also shown with the correlation coefficient $R=0.976$.}\label{fig:vlrc_fit_plot}
\end{figure}

First we look at the impact of these parameters on the convergence rate of the simple classifiers. At each parameter point the area under the validation loss curve is computed relative to the final validation loss at the stopping criteria point. To simplify notation this metric is referred to as the Validation Loss Residual Curve Area Under the Curve, or VLRC AUC. Some example validation loss curves are shown in Figure \ref{fig:Validation_loss_curves}. In Figure \ref{fig:vlrc_fit_plot} the plot demonstrates the relationship between the VLRC AUC, $\sigma_{w}$, and $\eta$. The experiments show no clear relationship between the VLRC AUC and $\eta$, but an approximately linear relationship between VLRC AUC and $\sigma_{w}$ with correlation coefficient $R = 0.976$.

Next we look at the effect of weight-induced variance on the models intended application. The learning objective in this paper is density ratio estimation, with its application as an importance sampling based re-weighting methodology. The performance is therefore framed by how well target distribution and the re-weighted reference distribution using the learned model, match. This can be quantified by a distance measure between the two distributions. The measure selected for these studies is Tsallis relative entropy ($D_{S_{2}}$; see Appendix \ref{app:DistMeasures_TS} for details). Tsallis relative entropy is a generalised form of the standard Boltzmann-Gibbs entropy, that is applicable to quasiprobabilities. It has a configurable hyperparameter $\alpha$ (entropic index), where in the limit $\alpha \rightarrow 1$, Tsallis relative entropy is equivalent to the Kullback-Leibler divergence; for these studies $\alpha=2$ is used.

\begin{figure}[t]
\centering
\includegraphics[scale=0.42]{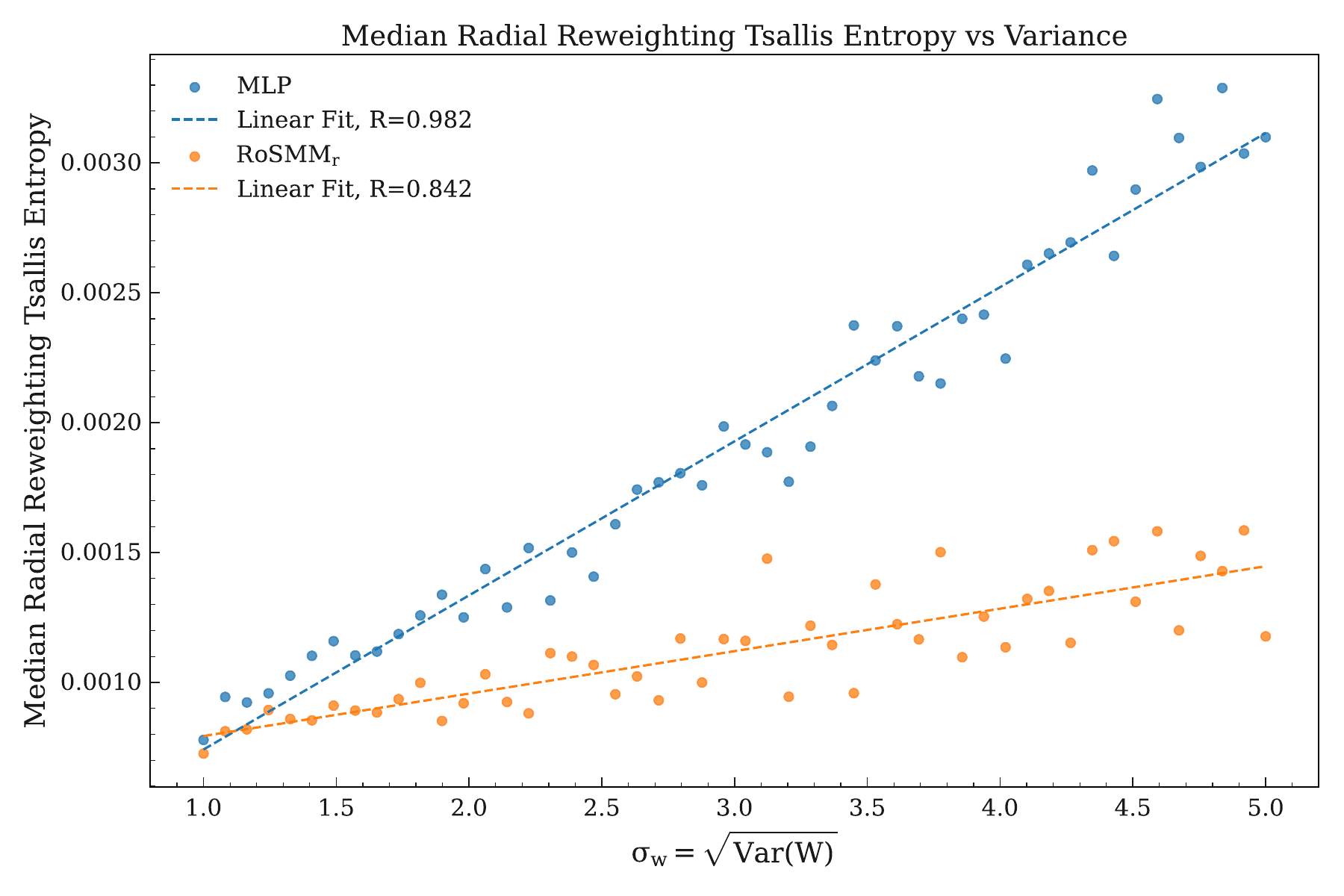}
\caption{The median Tsallis relative entropy ($D_{S_{2}}$) between the target and reweighted reference radial distributions for each value of $\sigma_w$ and both models. Linear fits and correlation coefficients for each plot are also shown.}\label{fig:ts_comparisons}
\end{figure}

The measure is computed using the radial distribution, since this feature encapsulates all of the information about the distribution. Once again, there appears to be no relationship between the performance of either model and the negative weight fraction\footnote{See Figure \ref{fig:entropy_plots} in the Appendix for plots demonstrating the relationship between the Tsallis entropy, $\sigma_{w}$, and $\eta$}. However, there is a relationship between the performance and the variance of the weights. To capture this effect more clearly, Figure \ref{fig:ts_comparisons} shows the median entropy for each value of $\sigma_{w}$ and a linear fit for each model. Here the relationship appears to be approximately linear between the reweighting performance and $\sigma_{w}$ with the performance of the simple classifier degrading 3.6 times faster than the $\textrm{RoSMM}_r$.

Based upon both studies, negative weights do not inherently pose a problem to model optimization, or performance. However, the variance of the weights directly impacts both optimization and performance. Therefore, as summarised by Equation \ref{eq:variance_gradient}
, the degraded performance of neural based likelihood ratio estimation tasks associated with negative weights does not stem from the signed nature of the weight. Rather, in practice, as the negative weight fraction of a sample increases, so does the weight variance. It is this correlated weight variance increase that underpins the degraded performance of optimisation tasks. In any case, the ratio of signed mixtures models demonstrate improved performance over an MLP classifier when negative weights are present even if the likelihood ratio is nonnegative.



\subsection{Results}

\begin{figure}[t]
\centering
\includegraphics[width=0.49\linewidth]{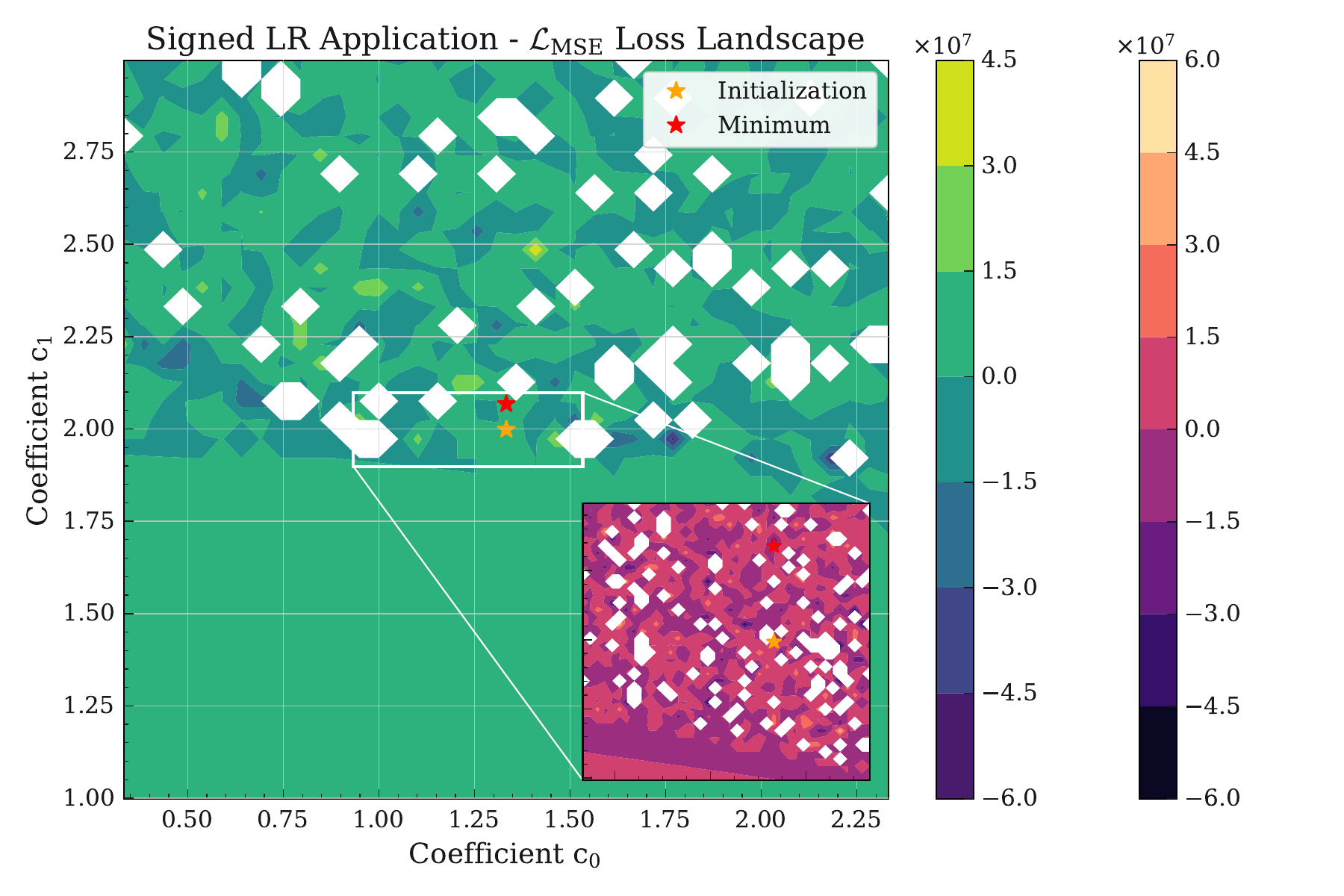}
\includegraphics[width=0.49\linewidth]{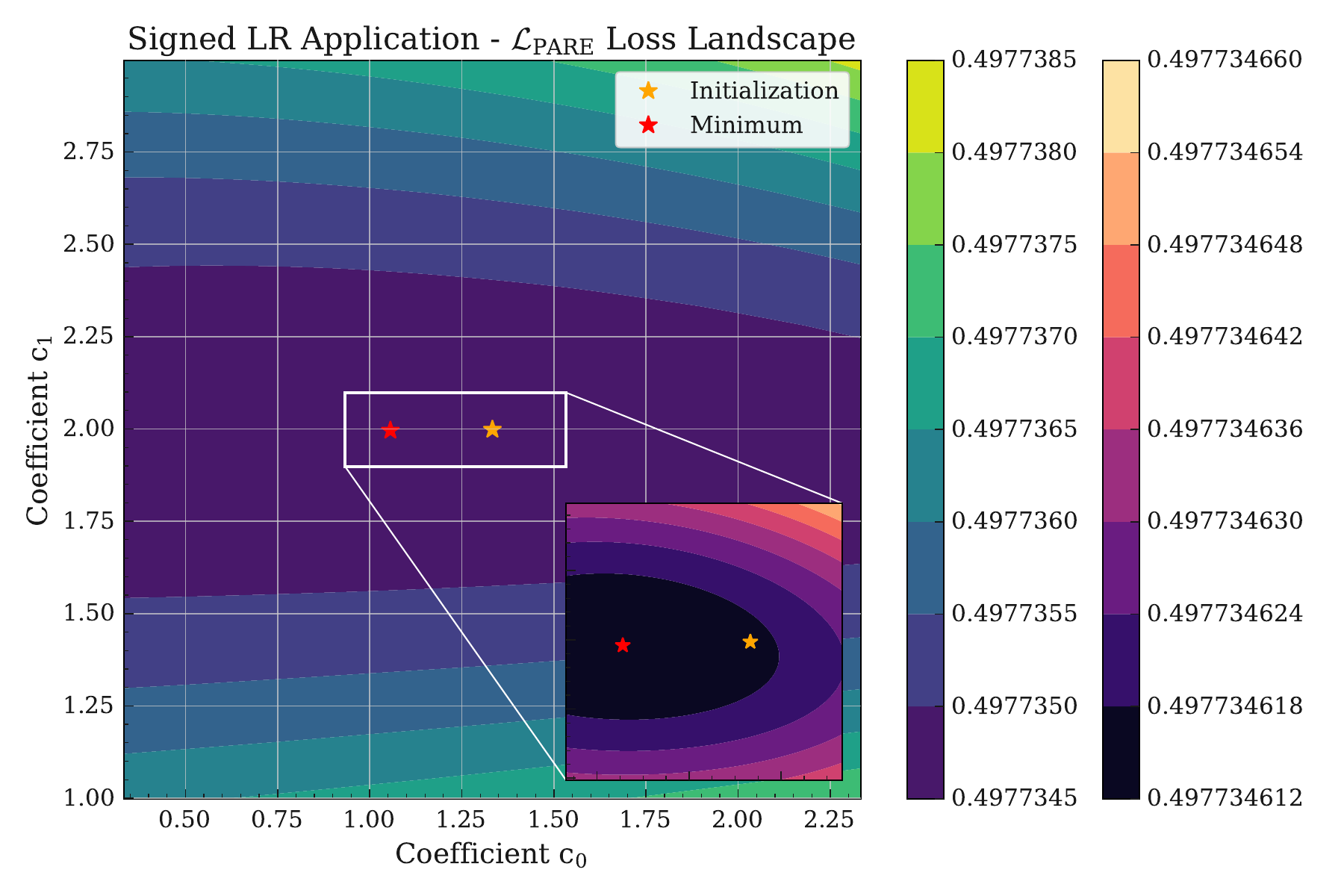}
\caption{The loss landscapes for optimizing the coefficients of the ratio of signed mixtures model with the sub-ratio models fixed for the signed likelihood ratio application. The left figure shows the MSE loss landscape and the right figure shows the PARE loss landscape.}\label{fig:toy_quasi_loss_landscape}
\end{figure}

Now we return to the Nonnegative and Signed LR test cases mentioned previously in Section \ref{sec:ToyGaussianDataGen}. For the ratio of signed mixtures model with coefficient optimisation ($\textrm{RoSMM}_{c}$), we plot the two-dimensional loss landscape according to Equation \ref{eq:mix_loss1} in Figure \ref{fig:toy_quasi_loss_landscape} for the Signed LR case. The instability of the MSE loss landscape is demonstrated in contrast to the smooth convexity of the new PARE loss. The coefficient loss landscapes for the Nonnegative test case can be found in Appendix \ref{app:toy_problem_results}.

To evaluate the models, we used the holdout testing datasets described in the previous section. The results for each type of model and application are shown in Table \ref{tab:toy_results}, using distance measures to define closure between the reference distribution weighted by a likelihood ratio model, and the target distributions. The definition of the different types of distance measures are given in Appendix \ref{app:DistMeasures}. The results also include the optimal likelihood ratio, which is calculated analytically using Equation \ref{eq:GaussMixModel}, as a performance benchmark. 

\begin{figure}[t]
\centering
\begin{subfigure}[t]{1.0\textwidth}
    \includegraphics[width=0.95\linewidth]{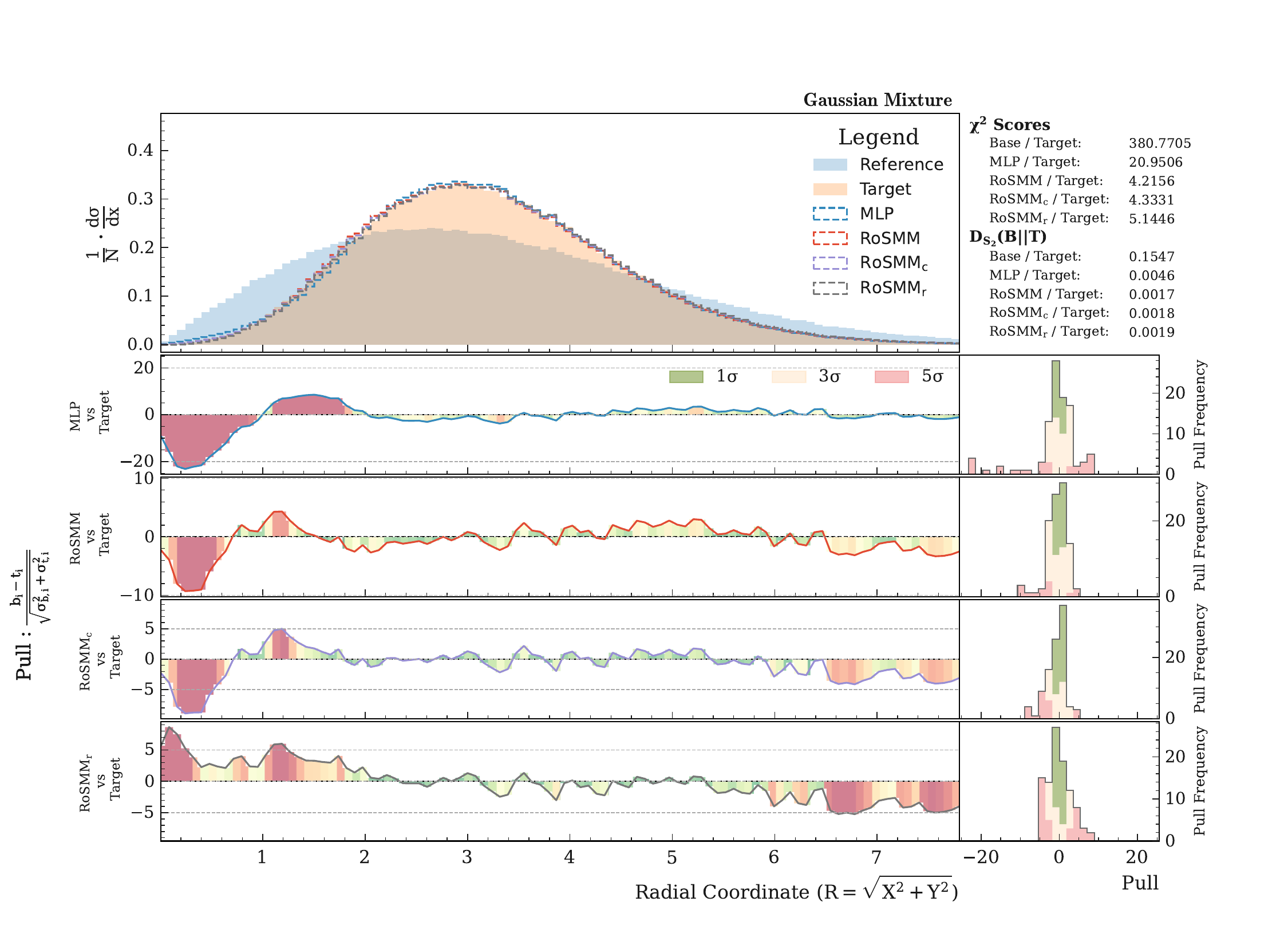}
    \caption{}\label{fig:toy_prob_radial_closure}
\end{subfigure}
\begin{subfigure}[t]{1.0\textwidth}
\includegraphics[width=0.95\linewidth]{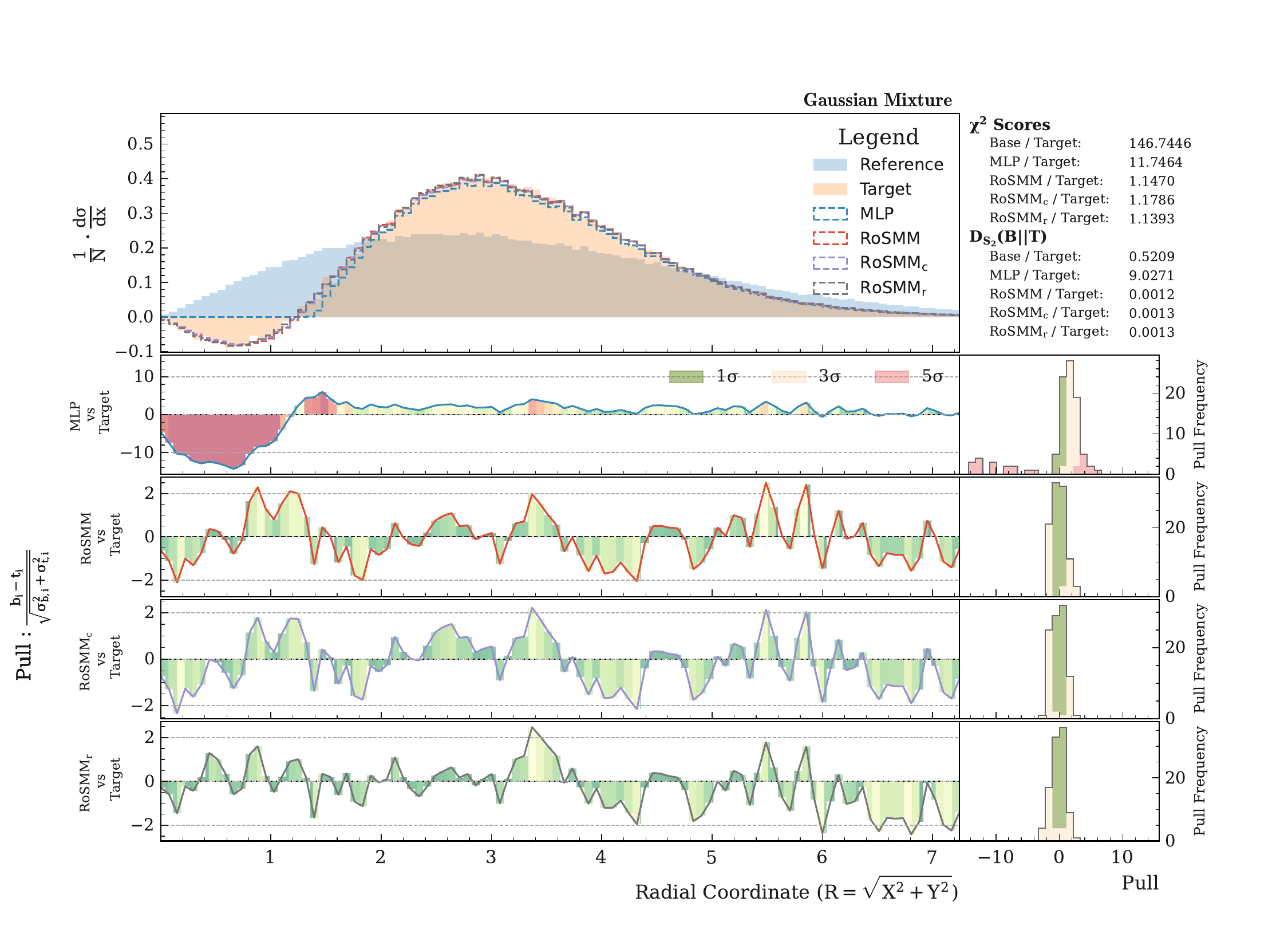}
\caption{}\label{fig:toy_quasi_radial_closure}
\end{subfigure}
\caption{The reweighting closure plots for the Gaussian mixture radial component distributions where the reference distribution is mapped to the target distribution using the different likelihood ratio estimation models. This figure shows the results for (a) nonnegative likelihood ratio application, and (b) signed likelihood ratio application.}\label{fig:toy_radial_closure}
\end{figure}

Figures \ref{fig:toy_prob_radial_closure} \& \ref{fig:toy_quasi_radial_closure} demonstrate the closure performance of the different likelihood ratio estimation models for the radial distribution, using the approximate likelihood ratio as a weight that maps the reference sample to the target. These plots demonstrate improved performance using the ratio of signed mixtures model for both the Nonnegative and Signed LR test cases, even for this simplified problem. In the Nonnegative LR test case the standard RoSMM performs the best, while in the Signed LR test case the RoSMM with full tuning ($\textrm{RoSMM}_{r}$) performs the best.

Figure \ref{fig:toy_quasi_radial_closure} demonstrates that the simple classifier trained using the BCE loss cannot learn the mapping to negative density regions. Consequently, the entire low radial value region that overlaps with the negative domain of the target distribution is mapped to zero; an example of the failure for traditional probabilistic ML models at expressing negative domains. The result of this is that although the remaining radial domain space is learnt, due to normalisation constraints the mapped distribution has the incorrect magnitude. The $x-y$ domain space closure plots, i.e. the input variables, can be found in Appendix \ref{app:toy_problem_results}.

\begin{table*}[t]
\caption{Signed Gaussian Mixture Model Results}\label{tab:toy_results}
\centering
\begin{tblr}{colspec={p{8cm}||>{\centering\arraybackslash}m{3.75cm}|>{\centering\arraybackslash}m{3.75cm}}}
 \hline
 \SetCell[c=3]{c} \textbf{Nonnegative Likelihood Ratio Test Case}\\
 \hline
 \textbf{Likelihood Ratio Estimator} & \textbf{Radial $\chi^2$ Score} & \textbf{Radial $D_{S_{2}}$} \\
 \hline
 MLP Classifier & 21.0 & $4.6\cdot 10^{-3}$ \\
 RoSMM & 4.22 & $1.7\cdot 10^{-3}$  \\
 $\textrm{RoSMM}_c$ - coefficient tuning & 4.33 & $1.8\cdot 10^{-3}$  \\
 $\textrm{RoSMM}_r$ - full tuning & 5.14 & $1.9\cdot 10^{-3}$ \\
 Analytically Optimal & 0.996 & $1 \cdot 4 \cdot 10^{-4}$ \\
 \hline
 \SetCell[c=3]{c} \textbf{Signed Likelihood Ratio Test Case}\\
 \hline
 \textbf{Likelihood Ratio Estimator}  & \textbf{Radial $\chi^2$ Score} & \textbf{Radial $D_{S_{2}}$} \\
 \hline
 MLP Classifier & 11.7 & 9.03 \\
 RoSMM & 1.15 & $1.2 \cdot 10^{-3}$  \\
 $\textrm{RoSMM}_c$ - coefficient tuning & 1.18 & $1.3 \cdot 10^{-3}$ \\
 $\textrm{RoSMM}_r$ - full tuning & 1.14 & $1.3 \cdot 10^{-3}$ \\
 Analytically Optimal & 0.958 & $4 \cdot 10^{-4}$ \\
 \hline
\end{tblr}
\end{table*}

\newgeometry{top=0.5cm,bottom=0.5cm}
\restoregeometry

\section{Application to Particle Physics: Standard Model Effective Field Theory in High Energy Physics}
\label{sec:QP-SMEFT}
In this section, proton-proton collisions at the LHC are simulated, in which pairs of Higgs bosons produced via gluon-gluon fusion ($gghh$) at a center of mass energy of $13$~TeV are created under two hypotheses; the Standard Model (SM) and an extended theory of the Standard Model using the Standard Model Effective Field Theory (SMEFT) \cite{Grzadkowski_2010} paradigm.  The latter is a low energy description of new physics at a new physics scale $\Lambda$. SMEFTs construct as an expansion in inverse powers of $\Lambda$, unknown contact interactions as operators $O_{i}$ of canonical dimension larger than four \cite{BUCHMULLER1986621,Grzadkowski_2010,Brivio_2019}. The corresponding amplitude that describes the $gg\rightarrow hh$ process is given by:
\begin{align}\label{eq:smeft_amplitude}
    \mathcal{M}_{gghh} =  \mathcal{M}_{\textrm{SM}} + \mathcal{M}_{\textrm{dim6}} + \mathcal{M}_{\textrm{dim6}^{2}},
\end{align}
where $\mathcal{M}_{\textrm{SM}}$ is the pure SM contribution, $\mathcal{M}_{\textrm{dim6}}$ is the single dimension-6 opertor insertions, and $\mathcal{M}_{\textrm{dim6}^{2}}$ is the double dimension-6 operation insertions; see Ref. \cite{Heinrich_2022} for more details. Depending on the adopted truncation scheme, the negative interference patterns with the SM, e.g. $2\textrm{Re}\{\mathcal{M}_{\textrm{SM}}^{*}\mathcal{M}_{\textrm{dim6}}\}$, can dominate in localised regions of phase space, thereby creating a quasiprobabilistic distribution. This is discussed further in the following sub-section.

\subsection{Data Generation}
Higgs boson pairs produced ($hh$) via gluon-gluon fusion ($gg$) from proton-proton collisions at a center of mass energy of $\sqrt{s} = 13$ TeV are simulated at Next-to-Leading order (NLO) in the strong coupling constant $\alpha_{S}$ (QCD) using the \textsc{PowhegBox-V2} MC generator \cite{Heinrich_2022} for the matrix element calculations. The \textsc{NNPDF3.0NLO} parton distribution function set is used \cite{Ball_2015} with the $h_{\textrm{damp}}$ parameter, which controls the high transverse momentum ($p_{\textrm{T}}$) radiation against which the $hh$ system recoils, set to $250$ GeV. The events were interfaced with \textsc{Pythia8} version 8.310 \cite{Sj_strand_2015} to model the parton shower, hadronisation, and underlying event utilising the \textsc{NNPDF3.0Nlo} parton distribution function set. The Higgs bosons are both forced to decay to muons, forming a final state with four muons from resonant Higgs decays, and additional parton final states in the form of gluons and quarks. 

\begin{figure}[h!]
\centering
\includegraphics[scale=0.75]{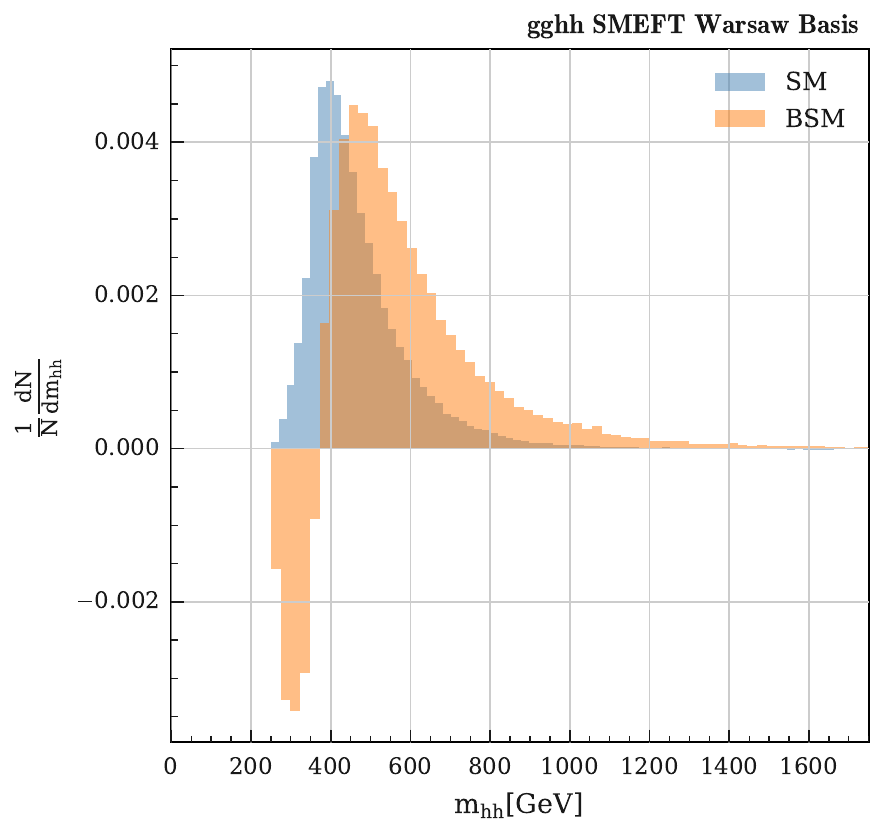}
\caption{Comparison of the reference sample (Standard Model) and target model (BSM) distributions of $m_{hh}$ in GeV. The data shown here is generated at the hard scattering level using \textsc{PowhegBox-V2} and does not include effects from the \textsc{Pythia8} generator.}\label{fig:smeft_distributions}
\end{figure}

Two samples were generated, a reference sample representing the Standard Model $gg\to hh$ process at NLO in QCD, and a target sample in the form of a beyond the Standard Model (BSM) $gg\to hh$ process generated using the Standard Model Effective Field Theory (SMEFT) \cite{Grzadkowski_2010,Brivio_2019,BUCHMULLER1986621} basis at NLO in QCD. The parameter configurations of each sample are given by Table \ref{tab:ggHH_configs}, where for the BSM target the transition amplitude of equation \ref{eq:smeft_amplitude} is truncated according to equation 2.7(a) of Ref. \cite{Heinrich_2022}. Consequently, $|\mathcal{M}_{gghh}|^{2}$ is expanded to include terms of $\Lambda^{-2}$, which includes the interference term $2\textrm{Re}\{\mathcal{M}_{\textrm{SM}}^{*}\mathcal{M}_{\textrm{dim6}}\}$. This interference term dominates at small di-Higgs invariant masses ($m_{hh}$), resulting in a quasiprobabislitic distribution. This is shown in Figure \ref{fig:smeft_distributions}, in which a comparison of the di-Higgs invariant mass distribution between the SM (reference) and BSM (target) is shown. For the hard scattering simulation the sampling is done in this $m_{hh}$ feature space, meaning all negative weights correspond to regions of negative density, and equivalently for positive weights/densities. Consequently, this is the most meaningful high-level feature, and so is used to evaluate the performance of the likelihood ratio estimation models, akin to the radial features in the signed Gaussian mixture model application. Figure \ref{fig:smeft_weights} in the appendix shows the distribution of weights for the training sample and the distribution of $W^2 - W$ to demonstrate the effect of the weights on training loss variance.

Each sample is composed of sixteen features used in training:
\begin{itemize}
\item Jet four-vector momentum: $(p_{T,j}, \eta_{j}, \phi_{j}, m_{j})$
\item Muon four-vector momentum (x4): $(p_{T,\mu}, \eta_{\mu}, \phi_{\mu})$,
\end{itemize}
where $p_{T}$, $\eta$, $\phi$, and $m$ for each object are the transverse mass, pseudo-rapdity, azimuthal, and mass of each object. The jet is selected based on descending $p_{T}$ ordering of all jets in the event. The muons are sorted according to descending $p_{T}$ ordering with 4 muons selected in each event. Muon masses were excluded from the training feature selection because this quantity is fixed in the simulator to always be 105.66 MeV. Then the following minimal selections were applied to the data:
\begin{itemize}
    \item Number of jets $\geq 1$
    \item Number of muons $\geq 4$
    \item $p_{T,j}, m_{j}, p_{T,\mu,0}, p_{T,\mu,1}, p_{T,\mu,2}, p_{T,\mu,3} > 0$
    \item $|\eta_{j}|, |\eta_{\mu,0}|, |\eta_{\mu,1}|, |\eta_{\mu,2}|, |\eta_{\mu,3}| < 5$
\end{itemize}
Additionally, to reduce problems with lack of support due to outliers, each $p_{T}$ feature was cut at the upper 99.8 percentile, which was computed using only the reference sample. This translated to additional cuts at:
\begin{itemize}
    \item $p_{T,j} < 318$ GeV
    \item $p_{T,\mu,0} < 507$ GeV
    \item $p_{T,\mu,1} < 395$ GeV
    \item $p_{T,\mu,2} < 225$ GeV
    \item $p_{T,\mu,3} < 153$ GeV
\end{itemize}
After these filters were applied to the data, 65\% of  the samples were used for training, 15\% for validation, and 20\% for testing, which corresponds to sample sizes per class (reference and target) of 776,369 for training, 179,162 for validation, and 238,883 for testing.

\begin{table*}
\caption{Proton-Proton collision $gg\to hh$ samples at $13.0$ TeV configurations}
\label{tab:ggHH_configs}
\centering
\begin{tblr}{colspec={p{4.5cm}||c|c|c|c|c}}
 \hline
 \SetCell[c=1]{c}{\centering \textbf{Sample}} & \SetCell[c=5]{c}{\textbf{SMEFT Wilson Coefficients}} \\
 \hline
 & $C_{H,kin}$ & $C_{H}$ & $C_{uH}$ & $C_{HG}$ & $\Lambda$ [TeV] \\
 \hline
 Standard Model (Ref.) & 0.0 & 0.0 & 0.0 & 0.0 & 0.0  \\
 \hline
 BSM (Target) & 13.5 & 2.64 & 12.6 & 0.0387 & 1.0  \\
 \hline
\end{tblr}
\end{table*}

\subsection{Machine Learning Setup}

All models were constructed and trained using PyTorch with the Adam optimizer \cite{NEURIPS2019_9015}. Training is stopped after a certain number (labelled ``patience'') of sequential epochs where the validation loss is greater than the lowest validation loss across all epochs. An epoch is defined as either the entire training dataset or $10^5$ training samples, whichever is smaller. The number of parameters in the models were chosen such that each model had roughly the same number of total parameters and expressibility. The architectures and model-specific settings are
\begin{itemize}
    \item MLP Classifier
    \begin{itemize}
        \item Inputs: 16
        \item 3 Hidden Layers: 128, 256, and 128 nodes (Linear with ReLU)
        \item Output: 1 (Sigmoid)
        \item Loss: Binary Cross Entropy
        \item Learning Rate: $10^{-4}$
        \item Batch Size: 64
        \item Patience: 20 epochs
    \end{itemize}
    \item Ratio of Signed Mixtures Model - composed of two sub-ratios and two mixture coefficients and trained with batch sizes of 128 for the positive-to-positive sub-ratio and 64 for the positive-to-negative sub-ratio. The final optimization step uses a learning rate of $10^{-4}$, batch size of $512$, and a patience of 10 epochs.
    Each sub-ratio has the following settings:
    \begin{itemize}
        \item Inputs: 16
        \item 3 Hidden Layers: 128, 128, and 128 nodes (Linear with ReLU)
        \item Output: 1 (Sigmoid)
        \item Loss: Binary Cross Entropy
        \item Learning Rate: $10^{-3}$
        \item Patience: 15 epochs
    \end{itemize}
\end{itemize}
Three flavours of the ratio of signed mixtures model are trained:
\begin{itemize}
\item $\textrm{RoSMM}$ - Ratio of Signed Mixtures Model using equations \ref{eq:rhat} \& \ref{eq:coeff_estimate}
\item $\textrm{RoSMM}_{c}$ - Ratio of Signed Mixtures Model with coefficient tuning according to equation \ref{eq:mix_loss1}
\item $\textrm{RoSMM}_{r}$ - Ratio of Signed Mixtures Model with coefficient and sub-density ratio tuning
\end{itemize}
where in all three cases the pre-trained sub-density ratio models according to Algorithm \ref{alg:subdensities}, are the same. The parameters $t_0 = 25619$ and $t_1 = 58$ were chosen with the same motivation as in the previous application.


\subsection{Results}

Since the reference dataset for this problem only included positive weights, the coefficient $c_0$ was fixed to $1$, and only the two sub-density ratios, $r_{++}$ and $r_{+-}$, needed to be learned. For the ratio of signed mixtures approach with only optimization for the coefficient $c_1$, the one-dimensional loss landscape according to Equation \ref{eq:mix_loss1} is shown in the Appendix.

For the fully optimized ratio of signed mixtures model the sub-density ratios can be optimized along with the coefficient. In this application, the total ratio is decomposed into
\begin{equation}
    r_{q}(\mathbf{x};c_1) = c_1 r_{++}(\mathbf{x}) + (1-c_1)r_{+-}(\mathbf{x})
\end{equation}
which provides a simpler optimization problem than the Gaussian mixture model from the previous section with four sub-density ratios.

\begin{figure}[!t]
\centering
\includegraphics[scale=0.5]{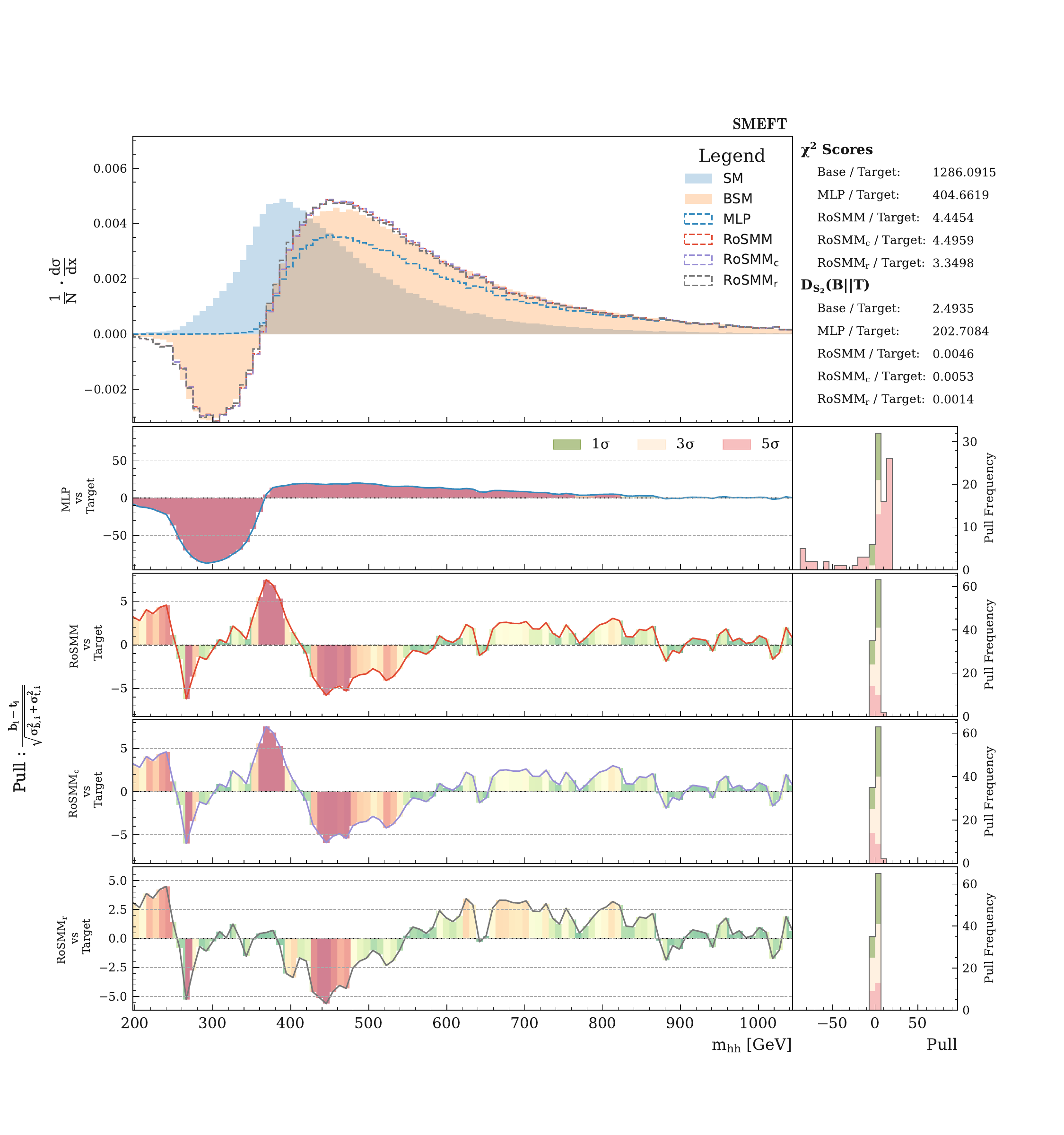}
\caption{The reweighting closure plots for the di-Higgs invariant mass ($m_{hh}$) where the reference (Standard Model) distribution is mapped to the target (SMEFT) distribution using the different likelihood ratio estimation models.}\label{fig:smeft_mhh_reweighting}
\end{figure}

The holdout testing datasets described in the previous section are used to evaluate each model. The results of each distance measure for each type of model and feature are shown in Tables \ref{tab:smeft_x2_results} and \ref{tab:smeft_tsallis_results} (see Appendix \ref{app:DistMeasures} for definitions). Based on these metrics, the simple classifier fails drastically for many features since, by construction, it cannot capture the information in the regions with negative density. However, in comparison the various ratio of signed mixtures models are able to learn the entirety of the likelihood ratio well with only minor differences in performance.

Figure \ref{fig:smeft_mhh_reweighting} demonstrates the closure performance of the different likelihood ratio estimation models for the mass of the two Higgs system using the approximate likelihood ratio as a weight that maps the reference sample to the target. Just like in the Gaussian mixture application, this plot shows that the simple classifier cannot learn the mapping to negative density regions. Instead, it maps those areas to zero and then learns the remaining shape of the distribution, but it fails to capture the proper magnitude even in the positive density region due to the normalization constraint. From this plot the fully optimized ratio of signed mixtures model appears to perform best overall, only slightly better than the other ratio of signed mixtures models. The same closure plots for each of the input variables for the models are included in the appendix.

\begin{table*}[!t]
\caption{SMEFT Results - $\chi^2$ Scores} \label{tab:smeft_x2_results}
\centering
\begin{adjustbox}{width=0.99\textwidth}
\begin{tblr}{colspec={p{4.25cm}||c|c|c|c|c|c|c|c},stretch=0.5}
 \hline
 \SetCell[r=2]{m}{\centering \textbf{LRE Model}} & \SetCell[c=8]{c} \textbf{Feature} &&&&&&&\\
 \hline
 & 1st Muon $p_T$ & 2nd Muon $p_T$ & 3rd Muon $p_T$ & 4th Muon $p_T$ & Jet $p_T$ & Jet $m$ & $hh$ $p_T$ & $m_{hh}$ \\
 \hline \hline
 MLP Classifier & 113 & 110 & 33.3 & 11.8 & 6.66 & 1.78 & 3.04 & 405 \\
 \hline
 RoSMM & \textbf{1.78} & \textbf{1.76} & \textbf{1.10} & 1.23 & 2.20 & 1.65 & 2.08 & 4.45 \\
 \hline
 $\textrm{RoSMM}_c$ & 1.79 & 1.80 & \textbf{1.10} & \textbf{1.21} & 2.21 & 1.65 & 2.08 & 4.50 \\
 \hline
 $\textrm{RoSMM}_r$  & 2.18 & 2.02 & 1.27 & 1.54 & \textbf{1.91} & \textbf{1.17} & \textbf{1.94} & \textbf{3.35} \\
 \hline
\end{tblr}
\end{adjustbox}
\end{table*}

\begin{table*}[!t]
\caption{SMEFT Results - $D_{S_{2}}$ Values} \label{tab:smeft_tsallis_results}
\centering
\begin{adjustbox}{width=0.99\textwidth}
    \begin{tblr}{colspec={p{4.25cm}||c|c|c|c|c|c|c|c},stretch=0.5}
    \hline
    \SetCell[r=2]{m}{\centering \textbf{LRE Model}} & \SetCell[c=8]{c} \textbf{Feature} &&&&&&&\\
    \hline
    & 1st Muon $p_T$ & 2nd Muon $p_T$ & 3rd Muon $p_T$ & 4th Muon $p_T$ & Jet $p_T$ & Jet $m$ & $hh$ $p_T$ & $m_{hh}$ \\
    \hline \hline
    MLP Classifier & 0.270 & 0.256 & 0.0611 & 0.0222 & 0.0190 & \textbf{0.0038} & 0.0137 & 203 \\
    \hline
    RoSMM & 0.0022 & 0.0083 & \textbf{0.0032} & \textbf{0.0036} & 0.205 & 0.0071 & \textbf{0.0042} & 0.0046 \\
    \hline
    $\textrm{RoSMM}_c$  & \textbf{0.0021} & 0.0095 & \textbf{0.0032} & \textbf{0.0036} & 0.0824 & 0.0072 & 0.0045 & 0.0053 \\
    \hline
    $\textrm{RoSMM}_r$  & 0.0036 & \textbf{0.0058} & 0.0036 & 0.0048 & \textbf{0.0038} & 0.0056 & 0.0074 & \textbf{0.0014} \\
    \hline
    \end{tblr}
\end{adjustbox}
\end{table*}

\section{Conclusions}
\label{sec:Conclusions}
We identified the specific challenges encountered in neural likelihood ratio estimation when confronted with negatively weighted events and introduced a novel loss function that yields an estimate of the quasiprobabilistic likelihood ratio through a generalization of the likelihood ratio trick. We have also introduced a new model architecture based on the ratio of signed mixture models that allows us to reformulate the task of estimating the quasiprobabilistic likelihood ratio in terms of purely positive weights. Finally, we have demonstrated these approaches on a pedagogical example and a realistic particle physics example. 

The approaches introduced here perform well and address a real-world problem encountered at the Large Hadron Collider. Namely, one wishes to implement an importance sampling procedure where both the proposal and target distributions are quasiprobabilistic and the likelihood ratio needed to compute importance weights must be estimated from weighted samples because likelihoods (densities) are implicitly defined by complex simulators with intractable likelihoods. 

The techniques described here face the same limitations as traditional neural likelihood ratio estimation, in addition to new challenges. These include the need for overlapping support between $q(\mathbf{x}|Y=0)$ and $q(\mathbf{x}|Y=1)$ (or the four combinations encountered in the subdensity ratios $r_{\pm\pm}$). Even when the distributions technically have the same support, if the likelihood ratios are far from one (very large or vanishingly small), then numerical stability and variance can be major impediments to learning. Sample efficiency is also an issue when the the two distributions are very similar and the ratio is very close to unity. Sample efficiency is an even bigger concern when learning subdensity ratios $r_{\pm\pm}$ because one needs a sufficient number of negatively weighted samples. The sample efficiency of neural ratio estimation can be improved with loss functions that leverage augmented training data as described in Refs.~\cite{Brehmer_2020,Stoye:2018ovl}, but adapting those techniques into the quasiprobabilistic setting remains a topic of future work.

\begingroup
\renewcommand
\refname{}
\printbibliography
\endgroup

\section*{Appendix}
\label{Appendix}

\subsection{Notation}
\label{app:Notation}
Here we list some commonly used notation used in this paper with explicit definitions to avoid any confusion:
\begin{align*}
    \mathbb{E}_{X}\left[f(X)\right] &= \int f(x)p(x)dx\\
    \mathbb{E}_{X,Y}\left[f(X,Y)\right] &= \int\int f(x,y)p(x,y)dxdy \\
    \mathbb{E}_{X|Y=y}\left[f(X,Y)\right] &= \mathbb{E}\left[f(X,Y)|Y=y\right] = \int f(x,y)p(x|y)dx
\end{align*}
where $p(\cdots)$ is the probability density function and $f$ is an arbitrary function.

\subsection{Likelihood Ratio Estimation}
\label{app:LRE}

In many areas of study it is often intractable to write down or compute the likelihood ratio directly, so machine learning techniques are often employed to estimate the likelihood ratio \cite{cranmer2016approximating,Brehmer_2020,hermans2020likelihoodfree}. We focus on one particular method, the ``likelihood ratio trick'' using a binary classifier to estimate the likelihood ratio \cite{cranmer2016approximating}.

\paragraph{Unweighted Binary Classification}

The traditional setting for learning a binary classifier is with unweighted data. We present the well-known results for this setting.

Let $X \sim p(\mathbf{x})$ and $Y$ be random variables that take values in $\mathcal{X}$ and $\mathcal{Y} = \{0,1\}$, respectively. $X$ represents observed data and $Y$ represents the supervisory class labels. The standard approach to learning a probabilistic classification model $s : \mathcal{X} \to [0,1]$ is to find the optimal function $s^*(\mathbf{x})$ that minimizes the binary cross entropy loss
\begin{align*}
    s_{\textrm{BCE}}^* = \argmin_{s} -\mathbb{E}_{X,Y}\left[Y\log (s(X)) + (1-Y)\log(1-s(X))\right]
\end{align*}
which has a well-known solution given by
\begin{align}\label{eq:bce_loss_optimal}
    s_{\textrm{BCE}}^*(\mathbf{x}) = \mathbb{E}_{Y|X=\mathbf{x}}\left[Y\right] = \frac{p(\mathbf{x}|Y=1)}{p(\mathbf{x}|Y=0) + p(\mathbf{x}|Y=1)}
\end{align}
if the training dataset has equal numbers of samples from each class. In practice, this solution is usually realized using a universal approximating function $s(\mathbf{x};\theta)$ where it holds that $s^*(\mathbf{x}) \equiv s(\mathbf{x};\theta^*)$ for the optimal parameters $\theta^*$. 

It then follows that the likelihood ratio can be written in terms of the optimal classifier as follows:
\begin{equation}
    r^*(\mathbf{x}) = \frac{s_{\textrm{BCE}}^*(\mathbf{x})}{1 - s_{\textrm{BCE}}^*(\mathbf{x})}
\end{equation}
which is referred to as the ``ratio trick''.

\paragraph{Weighted Binary Classification}
\label{sec:WeightedClassifiers}

In the case of weighted data, we use the same construction as in the Introduction but limit ourselves to probabilistic distributions. We define the proposal distribution as $\tilde{X} \sim p_{\rm proposal}(\mathbf{x})$ and the target distribution as $X \sim p_{\rm target}(\mathbf{x})$ such that both distributions are defined on the same domain $\mathcal{X}$ but with the support of $p_{\rm proposal}$ containing the support of $p_{\rm target}$. This construction is used to extend the results of unweighted classification to weighted classification in the natural way.

First, we introduce a tool that is useful for many proofs in this paper.
\blem[]\label{lem:reweighting}
Let $f: \mathcal{X} \times \mathcal{Y} \to \mathbb{R}$ be any function that can be written in the form $f(\mathbf{x},\mathbf{y}) = \sum_j g_j(\mathbf{x})h_j(\mathbf{y})$ for some functions $g_j:\mathcal{X} \to \mathbb{R}$ and $h_j : \mathcal{Y} \to \mathbb{R}$ then
    \begin{equation}
        \mathbb{E}_{\tilde{X},Y,W}\left[f(\tilde{X}, Y)W\right] = \mathbb{E}_{{X},Y}[f(X,Y)]
    \end{equation}
\elem
as long as the weights for each class are properly normalized, i.e. $\mathbb{E}_{W|Y}[W] = 1$.

Then the following proposition is a consequence of this lemma.
\bprop[]\label{thm:bce_loss}
The optimal binary classifier $s^*:\mathcal{X} \to (0,1)$ for the weighted binary cross entropy (weighted-BCE) loss function
trained with a class-balanced dataset and properly normalized weights is
\begin{align*}
    s_{\mathrm{BCE}}^*(\mathbf{x}) 
    &= \argmin_{s}\mathbb{E}_{\tilde{X},Y,W}\left[W\cdot \mathcal{L}_{\mathrm{BCE}}(s(\tilde{X}),Y)\right] \\
    &= \mathbb{E}_{Y|X=\mathbf{x}}\left[Y\right] \\
    &= \frac{p_{X}(\mathbf{x}|Y=1)}{p_{X}(\mathbf{x}|Y=0) + p_{X}(\mathbf{x}|Y=1)}
\end{align*}
\eprop

\brem
The mean-squared error loss function has the same optimal function.
\erem
So incorporating the weights into the training enables the classifier to learn the optimal classifier between the target distributions. As before, this can also give an estimate for the likelihood ratio between the true data distributions using the ratio trick.
\subsection{Signed Probability Measures}
\label{app:HahnJordan}
We introduce some key ideas involving signed measures and signed probability spaces from Khrennikov, but encourage the reader to view Ref \cite{Interpretations_of_Probability} for more details. Signed probability spaces are constructed in the same way as traditional probability spaces except the probability measure can be signed rather than nonnegative. Notably, they can violate the first Kolmogorov axiom, but still satisfy the second and (usually) third Kolmogorov axioms of probability theory. Random variables defined on such a space are called \textit{quasiprobabilistic}.

To motivate the signed mixture model solution of Section \ref{sec:SMM}, where the analog of the likelihood ratio for quasiprobabilities is constructed, we consider the Hahn-Jordan decomposition \cite{Interpretations_of_Probability}. The Hahn-Jordan decomposition implies that for any signed probability measure $\mathbf{P}$ there exist two nonnegative measures $\boldsymbol{\mu}^\pm$ such that $\mathbf{P} = \boldsymbol{\mu}^+ - \boldsymbol{\mu}^-$. Therefore, any quasiprobabilistic density can be decomposed into the mixture $q(\mathbf{x}) = cp_+(\mathbf{x}) + (1-c)p_-(\mathbf{x})$, for nonnegative density functions $p_\pm$ corresponding to the measures $\boldsymbol{\mu}^\pm$, and normalization constant $c \in [1,\infty)$. 

When dealing with datasets $\{\mathbf{x}_i,w_i\}^{N}_{i=1}$, a pragmatic question is raised of determining a suitable partition strategy that allows the user to estimate $p_{\pm}$ from the dataset alone. In the limit that the weights $w_i$ of each realisation $\mathbf{x}_{i}$ equate exactly to the likelihood ratio $q(\mathbf{x}_{i})/p(\mathbf{x}_{i})$, then the adopted strategy of defining $p_{\pm}$ by the sign of the weights (see equation \ref{eq:ProbabilityPartition}) yields the Hahn-Jordan decomposition.


\subsection{$\mathcal{L}_{\textrm{PARE}}$ Loss Considerations}
\label{app:loss_considerations}
In order to guide choices for the parameters $t_0,t_1$ we provide some recommendations and considerations. Recall the definitions from Section \ref{sec:Losses}, but consider the ratio $r_q$ as the independent variable rather than the input feature $\mathbf{x}$:
\begin{align*}
    r_{\textrm{pole}} &= -(t_0/t_1)^2 \\
    s^*_{\textrm{PARE}}(r_q;\mathbf{t}) &= \frac{t_0 + t_1 r_{{q}}}{t_0^2 + t_1^2 r_{{q}}} \\
\end{align*}
and define the two functions
\begin{align*}
    f(r_q;\mathbf{t}) &\equiv \mathcal{L}_{\textrm{PARE}}(s^*_{\textrm{PARE}}(r_q;\mathbf{t}),0;\mathbf{t}) = (1-t_0 s^*_{\textrm{PARE}}(r_q;\mathbf{t}))^2 \\
    g(r_q;\mathbf{t}) &\equiv \mathcal{L}_{\textrm{PARE}}(s^*_{\textrm{PARE}}(r_q;\mathbf{t}),1;\mathbf{t}) = (1-t_1 s^*_{\textrm{PARE}}(r_q;\mathbf{t}))^2
\end{align*}
such that
\begin{align*}
    \mathcal{L}_{\textrm{PARE}}(s^*_{\textrm{PARE}}(r_q;\mathbf{t}), y; \mathbf{t}) = (1-y)f(r_q;\mathbf{t}) + yg(r_q;\mathbf{t})
\end{align*}
For our recommendations, we consider the loss stable if each term in the loss is roughly of order one, or equivalently if each function $f,g$ is of order one.

To approximate when the loss becomes too large, consider the Maclaurin series of $f$ and $g$:
\begin{equation}
\begin{split}
    f(r_q;\mathbf{t}) 
    &= \sum_{k=2}^\infty (-1)^{k} (k-1)|r_{\textrm{pole}}|^{-(k-1)}(1-|r_{\textrm{pole}}|^{-1/2})^2 r_q^k \\
    &\approx \sum_{k=2}^\infty (-1)^{k} (k-1)|r_{\textrm{pole}}|^{-(k-1)} r_q^k \\
\end{split}
\end{equation}
\begin{equation}
\begin{split}
    g(r_q;\mathbf{t}) 
    &= \sum_{k=0}^\infty (-1)^{k+1} (k+1)|r_{\textrm{pole}}|^{-k}(1-|r_{\textrm{pole}}|^{-1/2})^2 r_q^k \\
    &\approx \sum_{k=0}^\infty (-1)^{k+1} (k+1)|r_{\textrm{pole}}|^{-k}r_q^k
\end{split}
\end{equation}
where the approximation is made using the assumption that $r_{\textrm{pole}}$ is large so $(1-|r_{\textrm{pole}}|^{-1/2})^2 \approx 1$. Hence to leading order $f$ is approximately stable until $r_q \sim \pm\sqrt{|r_{\textrm{pole}}|}$, while $g$ is approximately stable until $r_q \sim \pm{|r_{\textrm{pole}}|}$. Therefore it is recommended to choose $t_0$ and $t_1$ such that any expected values of $r_q$ are not too close to $\pm\sqrt{|r_{\textrm{pole}}|}$. In particular, if an initial estimate $\hat{r}_q$ for the likelihood ratio exists (such as an RoSMM model) we recommend choosing both parameters to be large positive integers such that
\begin{equation}
    -|t_0/t_1| < \inf_{\mathbf{x} \in \mathcal{X}} \hat{r}_q(\mathbf{x})
\end{equation}

It is also important to consider that this choice can come at the risk of numerical stability for the loss as well, so the parameters should not be chosen such that the pole is arbitrarily large. To improve numerical stability, we suggest choosing both parameters to be integers but not so large that fluctuations in $f$ and $g$ over the dataset are less than numerical precision. In particular, it is recommended that both parameters be positive integers to improve precision. This should be a good general approach but leaves some room for hyperparameter exploration.


\subsection{Distance Measures}
\label{app:DistMeasures}
With the focus of this paper on likelihood ratio estimation, and its application as a reweighting method that maps between probability measures, $r : P_{B}(X) \mapsto P_{T}(X)$, the performance of the proposed signed mixture model for quasiprobabilistic distributions is rooted in determining an appropriate measure of distance between two distributions \cite{Basic2011}. 

The distance measures defined within this section are used to numerically summarise the level of closure between a reference and target distribution. The distance measures will be provided in a binned and continuous form, due to the fact that the distance measures will be evaluated using 1-dimensional binned representations of the reference ($B \mapsto b_{i}$) and target ($T \mapsto t_{i}$) distributions:
\begin{align*}
    b_{i} = \frac{\mathbf{B}[X \in [x_{i}, x_{i}+\delta x] ]}{\delta x}, t_{i} = \frac{\mathbf{T}[X \in [x_{i}, x_{i}+\delta x] ]}{\delta x},
\end{align*}
where $\delta x$ is the interval of each bin. The yield of $i^{\textrm{th}}$ bin of a normalised histogram formed from a MC sampling of the density functions is therefore given by the sum of weights; $b_{i} = \sum^{N^{i}_{b}}_{j} w^{b}_{j}$ and $t_{i} = \sum^{N^{i}_{t}}_{j} w^{t}_{j}$, where $N^{i}_{b/t}$ represents the number of unweighted events in the $i^{\textrm{th}}$ bin.

Unfortunately there is no singular distance measure that will entirely encapsulate all aspects of distance between two distributions, therefore multiple distance measures will be provided.

\subsubsection{{$\chi^{2}$ Statistic}}

The $\chi^{2}$ statistic (divergence) for two density functions ($b(x)$ \& $t(x)$) is defined as:
\begin{align}
D_{\chi^{2}}(B,T) = \int \frac{b^{2}(x)}{t(x)} dx - 1.
\end{align}
For two binned representations of a density function, with a total of $N_{\textrm{bins}}$ discrete bins, its discrete form is given by \cite{gagunashvili2006comparison}:
\begin{align}
    D_{\chi^{2}}(B,T) = \sum^{N_{\textrm{bins}}}_{i} \frac{(W_{B}t_{i} - W_{T}b_{i})^{2}}{W^{2}_{B}\sigma^{2}_{T,i} + W^{2}_{T}\sigma^{2}_{B,i}},
\end{align}
where $W_{B} = \sum^{N_{b}}_{j} w^{b}_{j}$ and $W_{T} = \sum^{N_{t}}_{j} w^{t}_{j}$ are the total sum of weights of each sample, and $\sigma^{2}_{P,i} = \sum^{N^{i}_{b}}_{j} (w^{p}_{j})^{2}$ and $\sigma^{2}_{T,i} = \sum^{N^{i}_{t}}_{j} (w^{t}_{j})^{2}$ are the weight variances of the reference and target distributions for the $i^{\textrm{th}}$ bin, respectively. Given that the $\chi^{2}$ measure is dependent on the binning, the reduced-$\chi^{2}$ form is used by normalising according to the number of degrees of freedom ($N_{\textrm{bins}}$), yielding $\chi^{2}_{\textrm{ndof}} = \chi^{2}/(N_{\textrm{bins}}-1)$.

The $\chi^{2}_{\textrm{ndof}}$ statistic has a natural interpretation in statistics as how unlikely the distribution of B was drawn from the population of T. This is valuable due to the fact that formally $D_{\chi^{2}_{\textrm{ndof}}}(B,T) = 0$ if and only if $B=T$, however for finite sample sizes the $\chi^{2}_{\textrm{ndof}}$ statistic converges to unity (1.0) due to the fact that an unbiased sample drawn from a density function will deviate from the true density function due to stochastic variance. Therefore, on average the yield inside each bin will deviate from the true density fucntion by one standard deviation of the weight variance. Therefore, during the evaluation of the mapping performance of the mixture model, a $D_{\chi^{2}_{\textrm{ndof}}} = 1.0$ value is seen as a perfect score.

\subsubsection{{Tsallis Entropy \& Relative Entropy}}
\label{app:DistMeasures_TS}
Tsallis entropy is the generalised form of the standard Boltzmann-Gibbs entropy originating from statistical thermodynamics, that gave rise to Shannon entropy in information theory. Tsallis entropy is defined as:
\begin{align} \label{eq:TsallisEntropy}
    S_{q}(Q) = \int \ln_{\alpha}(q(x)) dx &\mapsto  \frac{1}{\alpha-1} \left[ 1 - \sum^{N_{\textrm{bins}}}_{j} q^{\alpha}_{j} \right],
\end{align}
where $\ln_{\alpha}(.)$ is the $\alpha$-deformed logarithm, and $Q$ is a generic signed probability measure. The continuous and discrete forms are both shown for completeness. Using this generalised form, the concept of information based relative entropy can be defined as:
\begin{align}
    D_{S_{\alpha}}(B||T) &= \int b(x) \ln_{\alpha}\left( \frac{b(x)}{t(x)} \right) dx           \mapsto \frac{1}{\alpha-1} \left[ 1 - \sum^{N_{\textrm{bins}}}_{j} b^{\alpha-1}_{j} t_{j} \right]
\end{align}
The $\alpha$ parameter, sometimes known as the entropic index (measure of non-extensivity), is seen in information theory as a measure of incomplete information in the system. In the limit of $\alpha \rightarrow 1$ one obtains the standard formalism of the Kullbleck-Leibler divergence, meanwhile $\alpha =2$ (used in this paper) is a natural representation of quantum entropy for quasiprobabilistic distributions \cite{PhysRevE.62.4665,Manfredi_2022}. This form of information entropy is key for the applications within this paper, because assigning $\alpha = 2$ yields a form of entropy that is safe against the negative domains of quasiprobabilistic densities ($q(x) < 0$ for some $x$).

\subsection{Two-Dimensional Gaussian Distributions}
\label{app:2D_gaussians}
The building blocks of the Guassian mixture models are the two-dimensional Gaussian probability distributions of the form $\mathcal{N}(0, \sigma^2 I_2)$. The probability density function (pdf) is parameterised by the following:
\begin{align}
    p(x,y;\sigma) &= \frac{1}{2\pi\sigma^2}\exp\left({-\frac{1}{2}\frac{x^2+y^2}{\sigma^2}}\right) = p(x;\sigma)p(y;\sigma),
\end{align}
which can be expressed in polar coordinates $(r, \phi)$ using the standard coordinate transform, modifying the density function to become:
\begin{align}\label{eq:camel_polar}
    p(r,\phi;\sigma) = \frac{r}{2\pi\sigma^2}\exp\left({-\frac{1}{2}\left(\frac{r}{\sigma}\right)^2}\right) = p(r;\sigma)p(\phi),
\end{align}
where $p(r;\sigma)$ and $p(\phi)$ determine the independent distributions over $r$ and $\phi$, respectively. They are given by:
\begin{align}
    p(r;\sigma) &= \frac{r}{\sigma^2}\exp\left({-\frac{1}{2}\left(\frac{r}{\sigma}\right)^2}\right), \\
    p(\phi) &= 1/2\pi,
\end{align}
where it should be noted that all of the sampling information is encoded in the radial distribution. 

We are able to define the following cumulative density function (cdf) and quantile function for the radial distribution
\begin{align}
    F(r;\sigma) &= 1 - \exp\left({-\frac{1}{2}\left(\frac{r}{\sigma}\right)^2}\right) \\
    Q(z;\sigma) &= \sqrt{-2\sigma^2 \log\left(1-z\right)}
\end{align}
which allows us to sample from this distribution directly using the probability integral transform (PIT).

Similarly, we can derive the Gaussian mixture radial cumulative distribution function:
\begin{align*}
    &F_X(r;c,\sigma_1,\sigma_2) \\
    &= 1 - c\exp\left(-\frac{1}{2}\left(\frac{r}{\sigma_1}\right)^2\right) + (c-1)\exp\left(-\frac{1}{2}\left(\frac{r}{\sigma_2}\right)^2\right)
\end{align*}
but there is no analytical quantile function, it must be solved numerically.

\subsection{Proofs and Derivations}
\label{app:Proofs}
\label{app:ConditionalWeightProposal}
\bpf[Equation \ref{eq:p_q_Expw}]
We wish to show
\begin{equation}
    \int w \,p_{\rm proposal}(w|\mathbf{x}) \, dw =  \frac{p_{\rm target}(\mathbf{x})}{p_{\rm proposal}(\mathbf{x})} \; .
\end{equation}
We will use Eq.~\ref{eq:p_w_given_x}
\begin{equation}
    p_{\rm proposal}(w|\mathbf{x}) = \int  p_{\rm proposal}(\mathbf{z}|\mathbf{x}) \, \delta\left(w - \frac{p_{\rm target}(\mathbf{x},\mathbf{z})}{p_{\rm proposal}(\mathbf{x}, \mathbf{z})}\right)\, d\mathbf{z} \;.
\end{equation}
Substituting $p_{\rm proposal}(w|\mathbf{x})$, integrating over $w$ to impose the the delta-function constraint, and employing the relationship $p(x,z)=p(x)p(z|x)$ we arrive at the following derivation:
\begin{eqnarray}
    \int w p_{\rm proposal}(w|\mathbf{x}) dw 
    &=& \int \int w p_{\rm proposal}(\mathbf{z}|\mathbf{x})  \, \delta\left(w - \frac{p_{\rm target}(\mathbf{x},\mathbf{z})}{p_{\rm proposal}(\mathbf{x}, \mathbf{z})}\right) dw d\mathbf{z} \\ \nonumber
    & = & \int \frac{p_{\rm target}(\mathbf{x},\mathbf{z})}{p_{\rm proposal}(\mathbf{x}, \mathbf{z})} p_{\rm proposal}(\mathbf{z}|\mathbf{x}) d\mathbf{z} \\ \nonumber
    & = & \frac{p_{\rm target}(\mathbf{x})}{p_{\rm proposal}(\mathbf{x})} \;.
\end{eqnarray}
\epf

\bprop[]
\begin{equation*}
    \mathbb{E}_{\tilde{X},W}\left[f(\tilde{X})W\right] = \mathbb{E}_{X}\left[f(X)\right]
\end{equation*}
holds for all functions $f$ if and only if
\begin{equation*}
    \mathbb{E}_{W|\tilde{X}=\mathbf{x}}\left[W\right] = q_{\rm target}(\mathbf{x})/p_{\rm proposal}(\mathbf{x})
\end{equation*}
\eprop

\begin{proof}
The forward direction $``\Rightarrow"$:

Note that 
\begin{align*}
    \mathbb{E}_{\tilde{X},W}\left[f(\tilde{X})W\right] &= \int_{\mathcal{X}}d\mathbf{x} \int_{\mathbb{R}} f(\mathbf{x})w p_{\rm proposal}(\mathbf{x}, w) dw \\
    &= \int_{\mathcal{X}} f(\mathbf{x}) \left(\int_{\mathbb{R}} w p_{\rm proposal}(\mathbf{x}, w) dw \right) d\mathbf{x}
\end{align*}
and
\begin{align*}
    \mathbb{E}_{X}\left[f(X)\right] &= \int_{\mathcal{X}} f(\mathbf{x}) q_{\rm target}(\mathbf{x}) d\mathbf{x}
\end{align*}
So if these two equations hold for all functions $f$ then it must be true that
\begin{align*}
    q_{\rm target}(\mathbf{x}) &= \int_{\mathbb{R}} w p_{\rm proposal}(\mathbf{x}, w) dw \\
    &= \int_{\mathbb{R}} w p_{\rm proposal}(w|\mathbf{x})p_{\rm proposal}(\mathbf{x}) dw \\
    &= p_{\rm proposal}(\mathbf{x})\int_{\mathbb{R}} w p_{\rm proposal}(w|\mathbf{x}) dw
\end{align*}
So if we divide both sides by $p_{\rm proposal}(\mathbf{x})$ this implies
\begin{align*}
    q_{\rm target}(\mathbf{x})/p_{\rm proposal}(\mathbf{x}) &= \int_{\mathbb{R}} w p_{\rm proposal}(w|\mathbf{x}) dw = \mathbb{E}_{W|\tilde{X}=\mathbf{x}}\left[W\right]
\end{align*}

The reverse direction $``\Leftarrow"$:

Observe the following
\begin{align*}
    \mathbb{E}_{\tilde{X},W}\left[f(\tilde{X})W\right] &= \mathbb{E}_{\tilde{X}}\mathbb{E}_{W|\tilde{X}}\left[f(\tilde{X})W\right] \\
    &= \mathbb{E}_{\tilde{X}}\left[f(\tilde{X})\mathbb{E}_{W|\tilde{X}}\left[W\right]\right] \\
    &= \mathbb{E}_{\tilde{X}}\left[f(\tilde{X})q_{\rm target}(\tilde{X})/p_{\rm proposal}(\tilde{X})\right] \\
    &= \int_{\mathcal{X}} f(\mathbf{x})(q_{\rm target}(\mathbf{x})/p_{\rm proposal}(\mathbf{x})) p_{\rm proposal}(\mathbf{x}) d\mathbf{x} \\
    &= \int_{\mathcal{X}} f(\mathbf{x})q_{\rm target}(\mathbf{x}) d\mathbf{x} \\
    &= \mathbb{E}_{X}\left[f(X)\right]
\end{align*}
\end{proof}


\bpf[Lemma \ref{lem:reweighting}]
Consider the following
\begin{align*}
    \mathbb{E}_{\tilde{X},Y,W}\left[f(\tilde{X}, Y)W\right] &= \mathbb{E}_{\tilde{X},Y,W}\Big[\sum_j g_j(\tilde{X})h_j(Y)W\Big] \\
    &= \sum_j \mathbb{E}_{\tilde{X},Y,W}\left[g_j(\tilde{X})h_j(Y)W\right]
\end{align*}
where for each $j$ we have
\begin{align*}
    \mathbb{E}_{\tilde{X},Y,W}\left[g_j(\tilde{X})h_j(Y)W\right] &= \mathbb{E}_{Y}\mathbb{E}_{\tilde{X},W|Y}\left[g_j(\tilde{X})h_j(Y)W\right] \\
    &= \mathbb{E}_{Y}\left[h_j(Y)\mathbb{E}_{\tilde{X},W|Y}\left[g_j(\tilde{X})W\right]\right] \\
    &= \mathbb{E}_{Y}\left[h_j(Y)\mathbb{E}_{X|Y}\left[g_j(X)\right]\right] \\
    &= \mathbb{E}_{Y}\mathbb{E}_{X|Y}\left[g_j(X)h_j(Y)\right] \\
    &= \mathbb{E}_{X,Y}\left[g_j(X)h_j(Y)\right]
\end{align*}
where it's important to note that there is an implicit dependence on the fact that $\mathbb{E}_{W|Y}[W] = 1$ for the equality from lines two to three to hold. For instance, if the function $g_j:\mathcal{X} \to \mathbb{R}$ is chosen to be $g_j(\mathbf{x}) = 1$ for all $\mathbf{x} \in \mathcal{X}$ then this requirement becomes obvious. Putting this all together we get that

\begin{align*}
    \mathbb{E}_{\tilde{X},Y,W}\left[f(\tilde{X}, Y)W\right] &= \sum_j \mathbb{E}_{\tilde{X},Y,W}\left[g_j(\tilde{X})h_j(Y)W\right] \\
    &= \sum_j \mathbb{E}_{X,Y}\left[g_j(X)h_j(Y)\right] \\
    &= \mathbb{E}_{X,Y}\Big[\sum_j g_j(X)h_j(Y)\Big] \\
    &= \mathbb{E}_{{X},Y}[f(X,Y)]
\end{align*}
\epf


\bpf[Equation \ref{eq:pare_loss}]
First, we want to eliminate the weight random variable $W$. To do this, we use Lemma \ref{lem:reweighting} once we show that the functions of $X$ and $Y$ are separable within the expectation. Observe that

\begin{align*}
    &\mathbb{E}_{\tilde{X},Y,W}\left[W\cdot\mathcal{L}_{\textrm{PARE}}(s(\tilde{X}),Y;\mathbf{t})\right] \\
    &= \mathbb{E}_{\tilde{X},Y,W}\left[W\cdot(1-s(\tilde{X})t_Y)^2\right] \\
    &= \mathbb{E}_{\tilde{X},Y,W}\left[W\cdot(1-2s(\tilde{X})t_Y + s^2(\tilde{X})t_Y^2)\right] \\
    &= \mathbb{E}_{X,Y}\left[1-2s(X)t_Y + s^2(X)t_Y^2\right] \\
    &= \mathbb{E}_{X,Y}\left[(1-s(X)t_Y)^2\right] \\
    &= \mathbb{E}_{X,Y}\left[\mathcal{L}_{\textrm{PARE}}(s(X),Y;\mathbf{t})\right]
\end{align*}
Now we can write the minimization problem as the following
\begin{align*}
    s_{\textrm{PARE}}^*(\mathbf{x};\mathbf{t}) &= \argmin_{s}\mathbb{E}_{\tilde{X},Y,W}\left[W\cdot\mathcal{L}_{\textrm{PARE}}(s(\tilde{X}),Y;\mathbf{t})\right] \\
    &= \argmin_{s}\mathbb{E}_{X,Y}\left[\mathcal{L}_{\textrm{PARE}}(s(X),Y;\mathbf{t})\right] \\
    &= \argmin_{s}\mathbb{E}_{X,Y}\left[(1-s(X)t_Y)^2\right] \\
    &= \argmin_{s}\mathbb{E}_{X}\mathbb{E}_{Y|X=\mathbf{x}}\left[(1-s(X)t_Y)^2\right]\\
    &= \argmin_{s}\int\mathbb{E}_{Y|X=\mathbf{x}}\left[(1-s(X)t_Y)^2\right]q_X(\mathbf{x})d\mathbf{x}
\end{align*}
where the integral can be extremized using the Euler-Lagrange equation. We define that
\begin{equation}
    \mathcal{L}(\mathbf{x},s) \equiv \mathbb{E}_{Y|X=\mathbf{x}}\left[(1-s(X)t_Y)^2\right]q_X(\mathbf{x})
\end{equation}
such that $s_{\textrm{PARE}}^*$ is the solution of the following (Euler-Lagrange) equation
\begin{align*}
    \frac{\partial \mathcal{L}}{\partial s} - \sum_{j=1}^n\frac{\partial}{\partial x_j}\left(\frac{\partial\mathcal{L}}{\partial (\nabla s)_j}\right) = 0
\end{align*}
or, since no derivatives of $s$ appear, simply the solution of
\begin{equation*}
    \frac{\partial \mathcal{L}}{\partial s} = 0
\end{equation*}
Therefore we have
\begin{align*}
    && \frac{\partial \mathcal{L}}{\partial s} = \frac{\partial}{\partial s}\mathbb{E}_{Y|X=\mathbf{x}}\left[(1-s(X)t_Y)^2\right]q_X(\mathbf{x}) = 0\\
    \implies&& \mathbb{E}_{Y|X=\mathbf{x}}\left[\frac{\partial}{\partial s}(1-s(X)t_Y)^2\right] = 0 \\
    \implies&& \mathbb{E}_{Y|X=\mathbf{x}}\left[-2t_Y(1-s_{\textrm{PARE}}^*(X;\mathbf{t})t_Y)\right] = 0 \\
    \implies&& 2s_{\textrm{PARE}}^*(\mathbf{x};\mathbf{t})\mathbb{E}_{Y|X=\mathbf{x}}\left[t_Y^2\right] -2\mathbb{E}_{Y|X=\mathbf{x}}\left[t_Y\right] = 0 \\
    \implies&& s_{\textrm{PARE}}^*(\mathbf{x};\mathbf{t}) = \frac{\mathbb{E}_{Y|X=\mathbf{x}}\left[t_Y\right]}{\mathbb{E}_{Y|X=\mathbf{x}}\left[t_Y^2\right]}
\end{align*}
wherever $q_X(x) \neq 0$. In the case of binary classification we can simplify this solution to be
\begin{align*}
    s_{\textrm{PARE}}^*(\mathbf{x};\mathbf{t}) &= \frac{\mathbb{E}_{Y|X=\mathbf{x}}\left[t_Y\right]}{\mathbb{E}_{Y|X=\mathbf{x}}\left[t_Y^2\right]} \\
    &= \frac{t_0 p(Y=0|X=\mathbf{x}) + t_1 p(Y=1|X=\mathbf{x})}{t_0^2 p(Y=0|X=\mathbf{x}) + t_1^2 p(Y=1|X=\mathbf{x})} \\
    &= \frac{t_0 q_{{X}}(\mathbf{x}|Y=0)p(Y=0) + t_1 q_{{X}}(\mathbf{x}|Y=1)p(Y=1)}{t_0^2 q_{{X}}(\mathbf{x}|Y=0)p(Y=0) + t_1^2 q_{{X}}(\mathbf{x}|Y=1)p(Y=1)}
\end{align*}
which in the case of a balanced training dataset we have
$$p(Y=0) = p(Y=1) = 1/2$$
so these factors cancel and we can write
\begin{align*}
    s_{\textrm{PARE}}^*(\mathbf{x};\mathbf{t}) &= \frac{t_0 q_{{X}}(\mathbf{x}|Y=0) + t_1 q_{{X}}(\mathbf{x}|Y=1)}{t_0^2 q_{{X}}(\mathbf{x}|Y=0) + t_1^2 q_{{X}}(\mathbf{x}|Y=1)} \\
    &= \frac{t_0 + t_1 r_{{q}}(\mathbf{x};\mathbf{c}^*)}{t_0^2 + t_1^2 r_{{q}}(\mathbf{x};\mathbf{c}^*)}
\end{align*}
\epf


\bpf[Equation \ref{eq:variance_gradient}]
There is a general proof which can be applied to this equation. Consider any function $f: \mathcal{X} \times \mathcal{Y} \to \mathbb{R}$ which can be decomposed as $f(X,Y) = \sum_j g_j(X)h_j(Y)$ for some functions $g_j:\mathcal{X}\to \mathbb{R}$ and $h_j:\mathcal{Y}\to\mathbb{R}$. Our goal is to show that any such function satisfies
\begin{align*}
    \text{Var}(W\cdot f(\tilde{X},Y)) - \text{Var}\left(f(X,Y)\right) = \mathbb{E}_{\tilde{X},Y,W}\left[(W^2-W)\cdot f(\tilde{X},Y)^2\right]
\end{align*}
which can be used to derive Equation \ref{eq:variance_gradient}. Recall that for any random variable $X$ it holds that
\begin{align*}
    \text{Var}(X) = \mathbb{E}\left[X^2\right] - \mathbb{E}\left[X\right]^2
\end{align*}
So we can write that
\begin{align*}
    &\text{Var}(W\cdot f(\tilde{X},Y)) \\
    &\quad = \mathbb{E}_{\tilde{X},Y,W}\left[W^2\cdot f(\tilde{X},Y)^2\right] - \mathbb{E}_{\tilde{X},Y,W}\left[W\cdot f(\tilde{X},Y)\right]^2
\end{align*}
and using Lemma \ref{lem:reweighting} we have
\begin{align*}
    &\text{Var}(f(X,Y)) \\
    &\quad = \mathbb{E}_{X,Y}\left[f(X,Y)^2\right] - \mathbb{E}_{X,Y}\left[f(X,Y)\right]^2 \\
    &\quad = \mathbb{E}_{\tilde{X},Y,W}\left[W\cdot f(\tilde{X},Y)^2\right] - \mathbb{E}_{\tilde{X},Y,W}\left[W\cdot f(\tilde{X},Y)\right]^2
\end{align*}
Plugging in these two results and usinng the linearity of expectations, we see that
\begin{align*}
    &\text{Var}(W\cdot f(\tilde{X},Y)) - \text{Var}\left(f(X,Y)\right) \\
    &\quad = \mathbb{E}_{\tilde{X},Y,W}\left[W^2\cdot f(\tilde{X},Y)^2\right] - \mathbb{E}_{\tilde{X},Y,W}\left[W\cdot f(\tilde{X},Y)^2\right] \\
    &\quad = \mathbb{E}_{\tilde{X},Y,W}\left[(W^2-W)\cdot f(\tilde{X},Y)^2\right]
\end{align*}

Now that this equation is proven, we want to apply it to the formula for SGD parameter updates. Recall that these updates are defined as
\begin{equation*}
    \theta^{t+1} = \theta^t - \gamma \nabla_\theta\mathcal{L}_{\textrm{batch}} (s(\tilde{X};\theta^t ),Y,W)\bigr|_{\theta = \theta^t}
\end{equation*}
where $\gamma$ is the SGD learning rate and $\mathcal{L}_{\textrm{batch}}$ is given by
\begin{equation*}
    \mathcal{L}_{\textrm{batch}} (s(\tilde{X};\theta^t ),Y,W) = \frac{1}{N_{\textrm{batch}}}\sum_{i=1}^{N_{\textrm{batch}}}W_i\cdot \mathcal{L}(s(\tilde{X}_i;\theta^t ),Y_i)
\end{equation*}
where $\mathcal{L}(s,y)$ is a loss function and $N_{\textrm{batch}}$ is the batch size.

Then it follows that
\begin{align*}
\text{Var}_{\tilde{X},Y,W}(\theta^{t+1}) &= \text{Var}\left(\theta^t - \gamma \nabla_\theta\mathcal{L}_{\textrm{batch}} (s(\tilde{X};\theta^t ),Y,W)\bigr|_{\theta = \theta^t}\right) \\
&= \gamma^2\text{Var}\left(\nabla_\theta\mathcal{L}_{\textrm{batch}} (s(\tilde{X};\theta^t ),Y,W)\bigr|_{\theta = \theta^t}\right) \\
&= \gamma^2\text{Var}\left(\frac{1}{N_{\textrm{batch}}}\sum_{i=1}^{N_{\textrm{batch}}}W_i\cdot \nabla_\theta\mathcal{L}(s(\tilde{X}_i;\theta^t ),Y_i)\bigr|_{\theta = \theta^t}\right) \\
&= \frac{\gamma^2}{N_{\textrm{batch}}^2}\text{Var}\left(\sum_{i=1}^{N_{\textrm{batch}}}W_i\cdot \nabla_\theta\mathcal{L}(s(\tilde{X}_i;\theta^t ),Y_i)\bigr|_{\theta = \theta^t}\right) \\
&= \frac{\gamma^2}{N_{\textrm{batch}}^2}\sum_{i=1}^{N_{\textrm{batch}}}\text{Var}\left(W_i\cdot \nabla_\theta\mathcal{L}(s(\tilde{X}_i;\theta^t ),Y_i)\bigr|_{\theta = \theta^t}\right) \\
&= \frac{\gamma^2}{N_{\textrm{batch}}}\text{Var}\left(W\cdot \nabla_\theta\mathcal{L}(s(\tilde{X};\theta^t ),Y)\bigr|_{\theta = \theta^t}\right)
\end{align*}
where we have used that each $X_i$, $Y_i$, and $W_i$ are i.i.d. for the last two lines. Similarly,
\begin{align*}
\text{Var}_{X,Y}(\theta^{t+1}) &= \frac{\gamma^2}{N_{\textrm{batch}}}\text{Var}\left(\nabla_\theta\mathcal{L}(s(X;\theta^t ),Y)\bigr|_{\theta = \theta^t}\right)
\end{align*}

For standard loss functions such as BCE and MSE it can be shown that the function
\begin{equation*}
    f(X,Y) = \nabla_\theta\mathcal{L}(s(X;\theta^t ),Y)\bigr|_{\theta = \theta^t}
\end{equation*}
satisfies the requirements of Lemma \ref{lem:reweighting}. Therefore, we can conclude that
\begin{equation*}
\begin{split}
    &\text{Var}_{\tilde{X},Y,W}\big(\theta^{t+1}\big) - \text{Var}_{X,Y}\big(\theta^{t+1}\big) \\
    &\quad = \frac{\gamma^2}{N_{\textrm{batch}}} \mathbb{E}_{\tilde{X},Y,W}\left[(W^2-W) \cdot \left(\nabla_\theta\mathcal{L}(s(\tilde{X};\theta), Y)\bigr|_{\theta = \theta^t}\right)^2\right]
\end{split}
\end{equation*}

\epf

\bpf[Proposition \ref{thm:bce_loss}]
First, we will show the proof for the binary cross entropy loss function, i.e. we wish to show that
\begin{align*}
    s_{\mathrm{BCE}}^*(\mathbf{x}) &= \argmin_{s}\mathbb{E}_{\tilde{X},Y,W}\left[W\cdot \mathcal{L}_{\textrm{BCE}}(s(\tilde{X}),Y)\right] \\
    &= \frac{p_X(\mathbf{x}|Y=1)}{p_X(\mathbf{x}|Y=0) + p_X(\mathbf{x}|Y=1)}
\end{align*}
Since
\begin{align*}
    \mathcal{L}_{\textrm{BCE}}(s(\tilde{X}),Y) = -Y\log s(\tilde{X}) - (1-Y)\log(1-s(\tilde{X}))
\end{align*}
we can apply Lemma \ref{lem:reweighting} to the weighted expectation such that
\begin{equation*}
    \mathbb{E}_{\tilde{X},Y,W}\left[W\cdot \mathcal{L}_{\textrm{BCE}}(s(\tilde{X}),Y)\right] = \mathbb{E}_{X,Y}\left[\mathcal{L}_{\textrm{BCE}}(s(X),Y)\right]
\end{equation*}
and hence
\begin{align*}
    s_{\mathrm{BCE}}^* &= \argmin_{s}\mathbb{E}_{\tilde{X},Y,W}\left[W\cdot \mathcal{L}_{\textrm{BCE}}(s(\tilde{X}),Y)\right] \\
    &= \argmin_{s}\mathbb{E}_{X,Y}\left[\mathcal{L}_{\textrm{BCE}}(s(X),Y)\right]
\end{align*}
Thus we can use the well-known result that the optimal classifier for this loss function with a class-balanced dataset, i.e. $p(Y=0)=p(Y=1)$, is
\begin{align*}
    s_{\mathrm{BCE}}^*(\mathbf{x}) &= \frac{p_X(\mathbf{x}|Y=1)}{p_X(\mathbf{x}|Y=0) + p_X(\mathbf{x}|Y=1)}
\end{align*}
\brem
The same result follows in the case of mean-squared error, which is defined as
\begin{align*}
    \mathcal{L}_{\textrm{MSE}}(s(\tilde{X}),Y) &= (Y - s(\tilde{X}))^2 \\
        &= Y^2 - 2Y\cdot s(\tilde{X}) + s(\tilde{X})^2
\end{align*}
using the exact same procedure as before.
\erem
\epf

\clearpage

\subsection{Additional Algorithms}\label{sec:appendix_algorithms}

\begin{algorithm}
\caption{Generating the signed Gaussian mixture datasets}\label{alg:camel}
\begin{algorithmic}
\Require $N \geq 1, a > 0, b < 0$
\State $P_a := |a| / (|a| + |b|)$
\State $P_b := 1 - P_a$
\State $\mathcal{D} := \{\}$
\While{size($\mathcal{D}$) $< N$}
    \State Draw $p, u_1, u_2 \sim \text{Unif}(0,1)$
    \State $\phi := 2\pi u_2$
    \If{$p < P_a$}
        \State $r := \sqrt{-2\sigma_1^2 \log(u_1)}$
        \State $w := 1$
    \Else
        \State $r := \sqrt{-2\sigma_2^2 \log(u_1)}$
        \State $w := -1$
    \EndIf
    \State $\mathbf{x} := (r\cos\phi, r\sin\phi)$
    \State $\mathcal{D} := \mathcal{D} \cup \{(\mathbf{x}, w)\}$
\EndWhile
\end{algorithmic}
\end{algorithm}

\clearpage

\subsection{Additional Plots: Model Optimisation and Negative Weights}

\begin{figure}[H]
\centering
\includegraphics[scale=0.34]{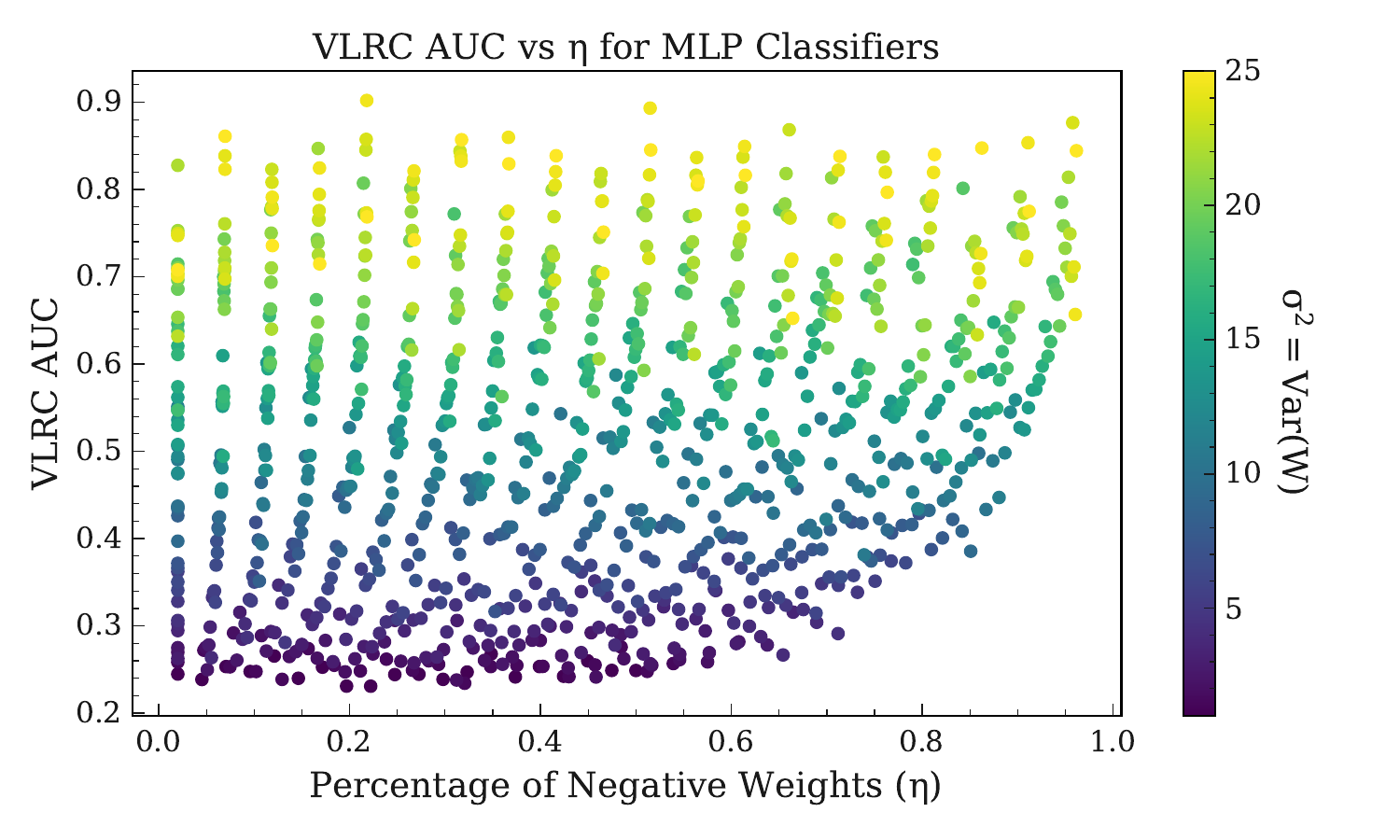}
\includegraphics[scale=0.34]{images/NegWeightStudy/vlrc_vs_var_plot.pdf}
\caption{Scatterplot demonstrating the relationships between the Area Under the Curve for the Validation Loss Residual Curves (VLRC AUC), the standard deviation of the weights $\sigma_w$, and the fraction of negative weights $\eta$. Both figures show the same data but projected onto different 2D planes where each datapoint represents a simple MLP classifier trained for a particular choice of $(\sigma_w, \eta)$.}\label{fig:vlrc_plots}
\end{figure}

\begin{figure}[H]
\centering
\includegraphics[scale=0.34]{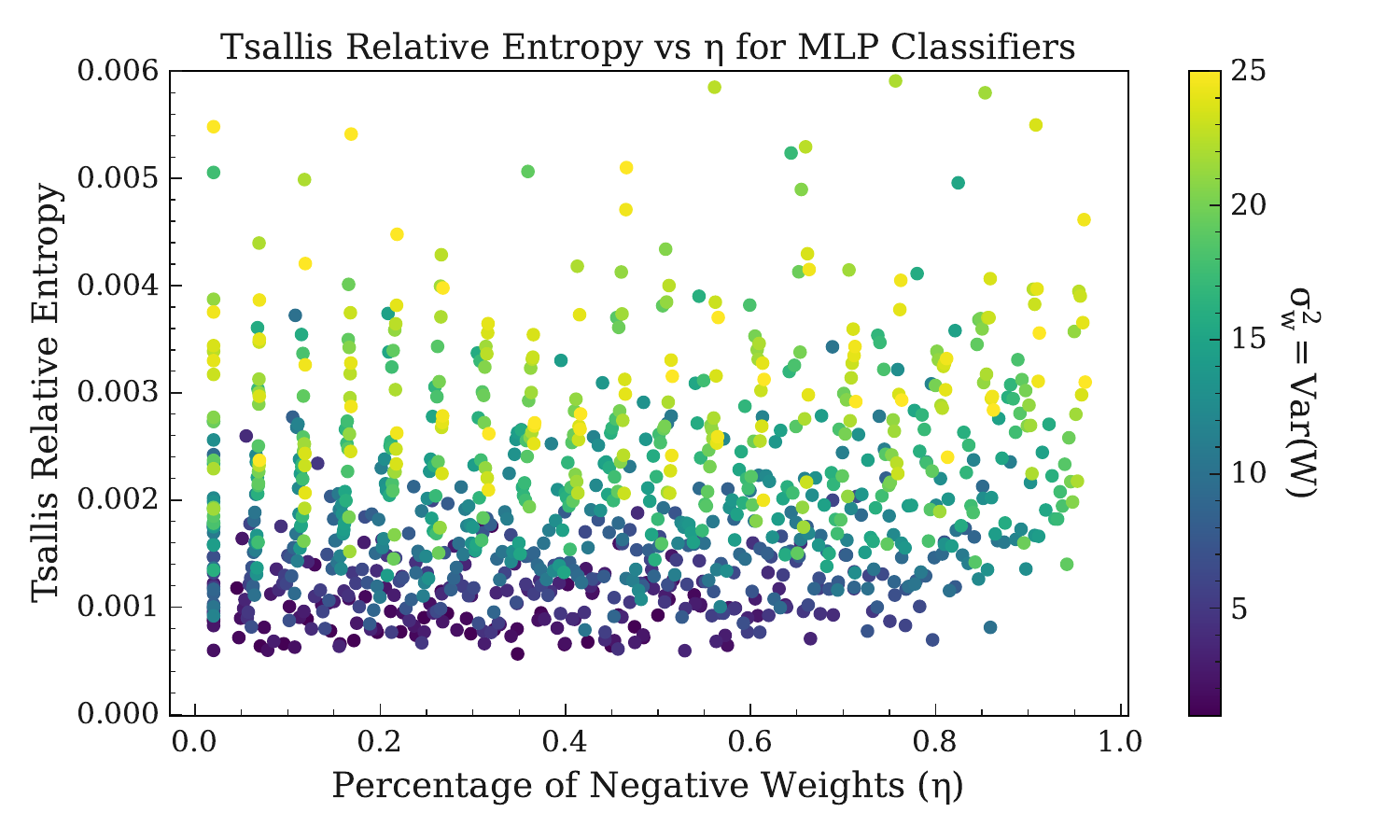}
\includegraphics[scale=0.34]{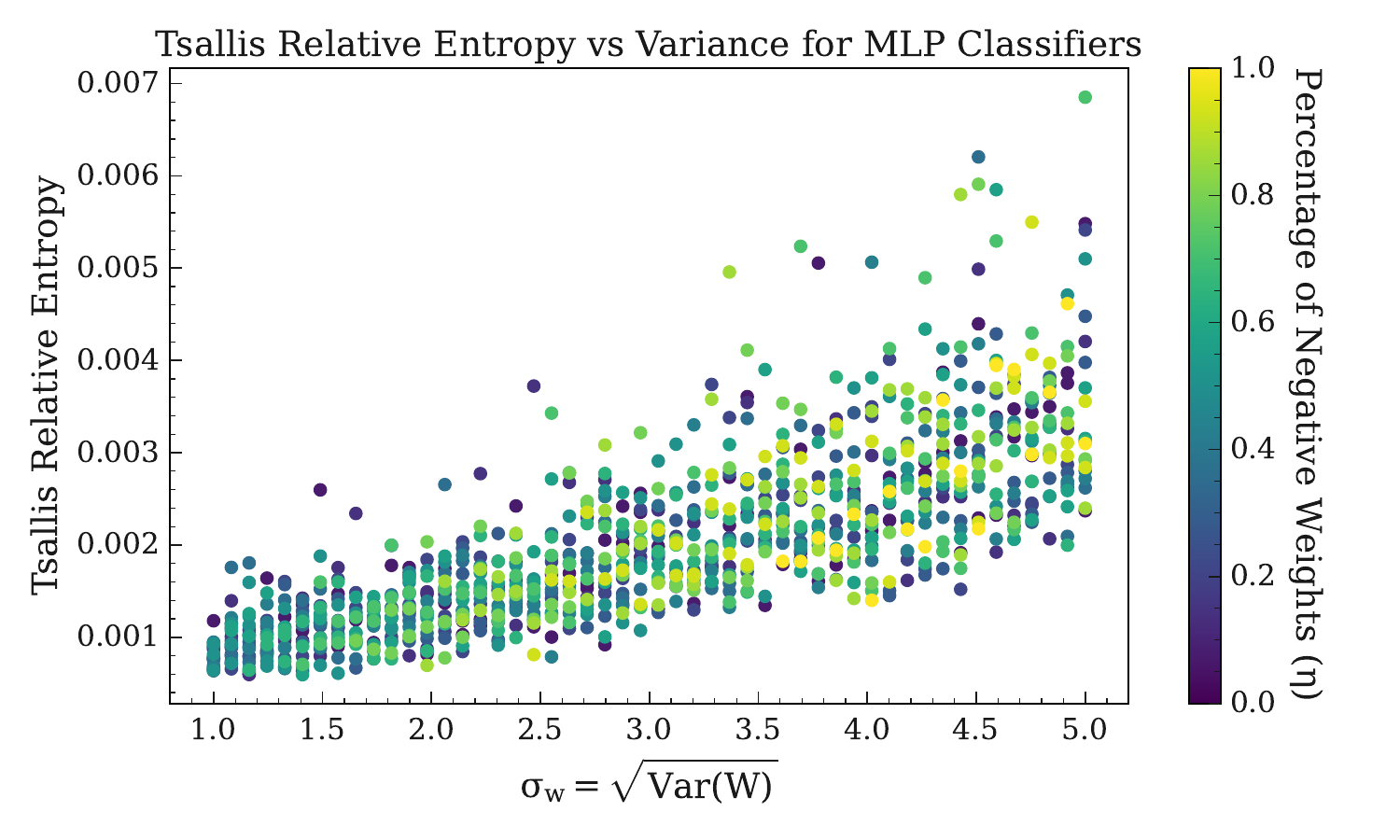}
\includegraphics[scale=0.34]{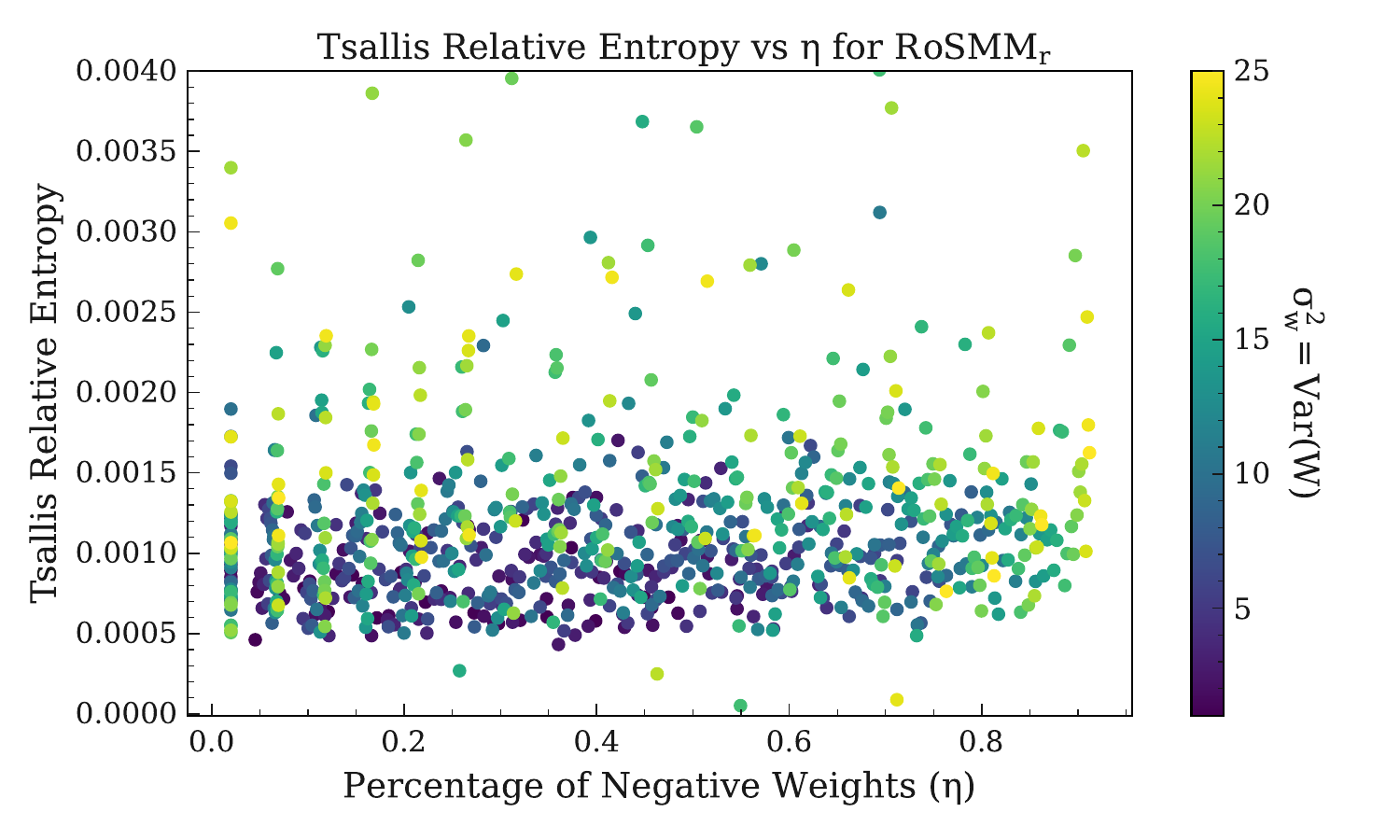}
\includegraphics[scale=0.34]{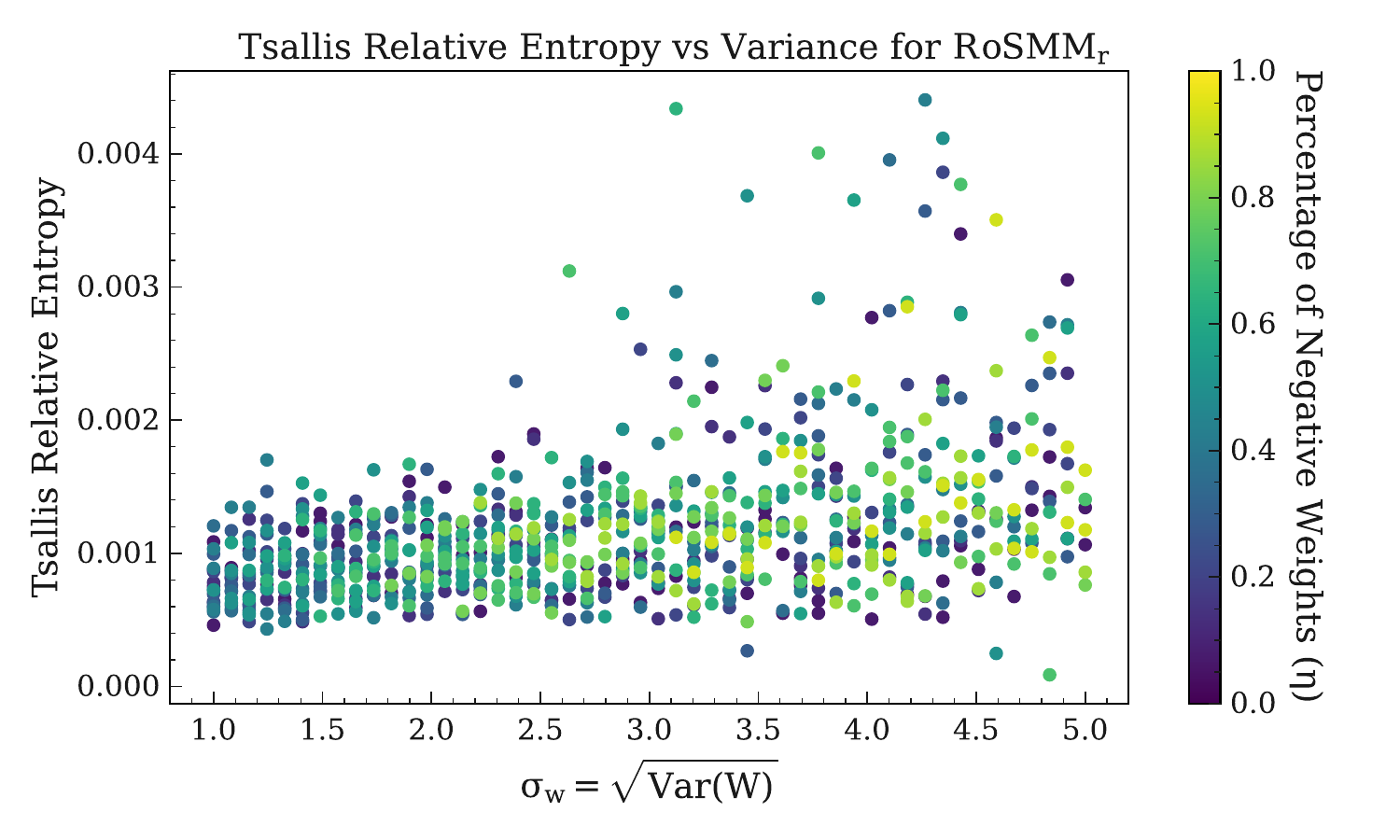}
\caption{Scatterplots demonstrating the relationships between the Tsallis entropy between the target and reweighted reference radial distributions, the standard deviation of the weights $\sigma_w$, and the fraction of negative weights $\eta$. For each type of model, both figures show the same data but projected onto different 2D planes. Each datapoint represents a simple MLP classifier or signed mixture model trained for a particular choice of $(\sigma_w, \eta)$.}\label{fig:entropy_plots}
\end{figure}


\subsection{Additional Plots: Signed Gaussian Mixture Models}
\label{app:toy_problem_results}

\begin{figure}[H]
\centering
\includegraphics[scale=0.28]{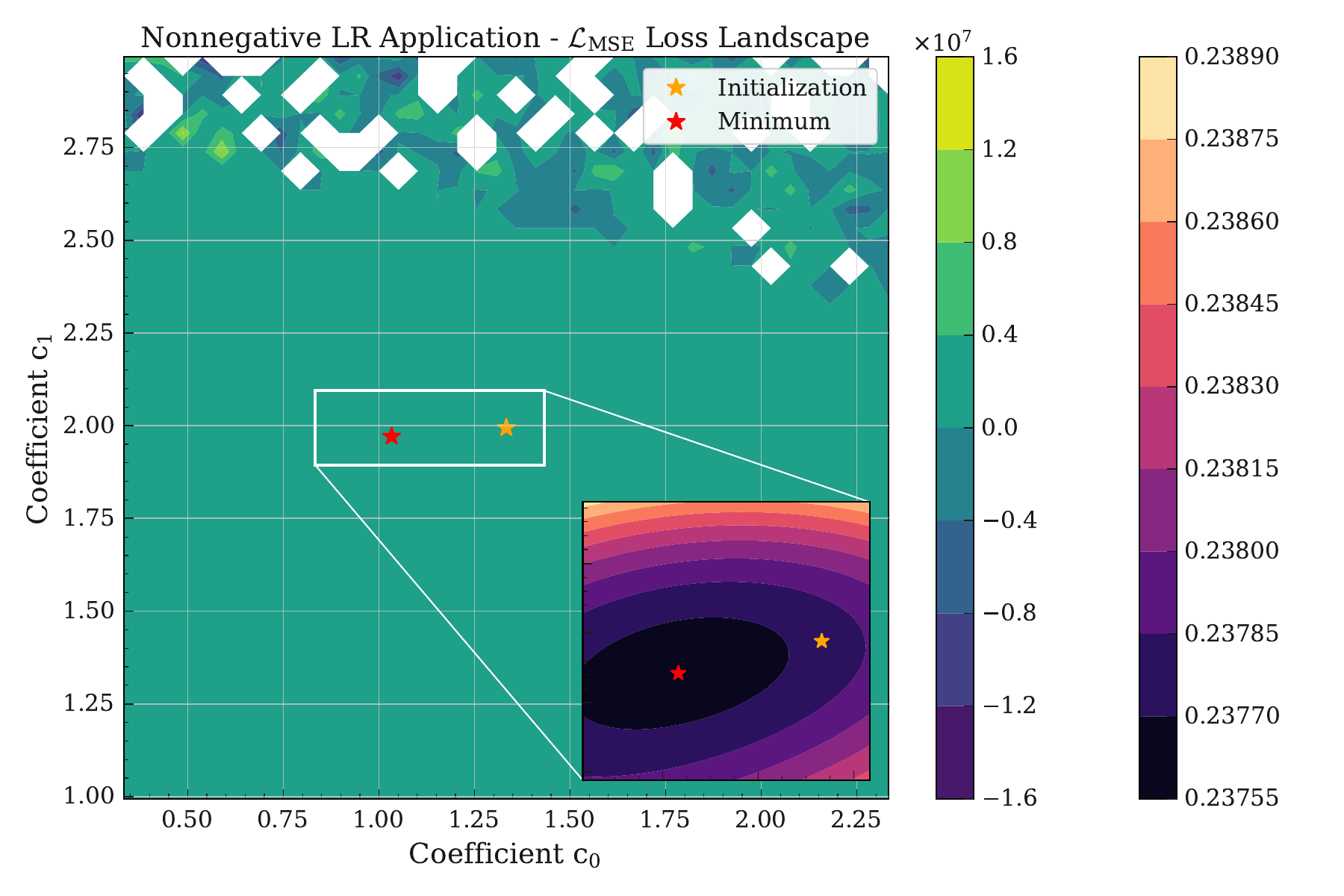}
\includegraphics[scale=0.28]{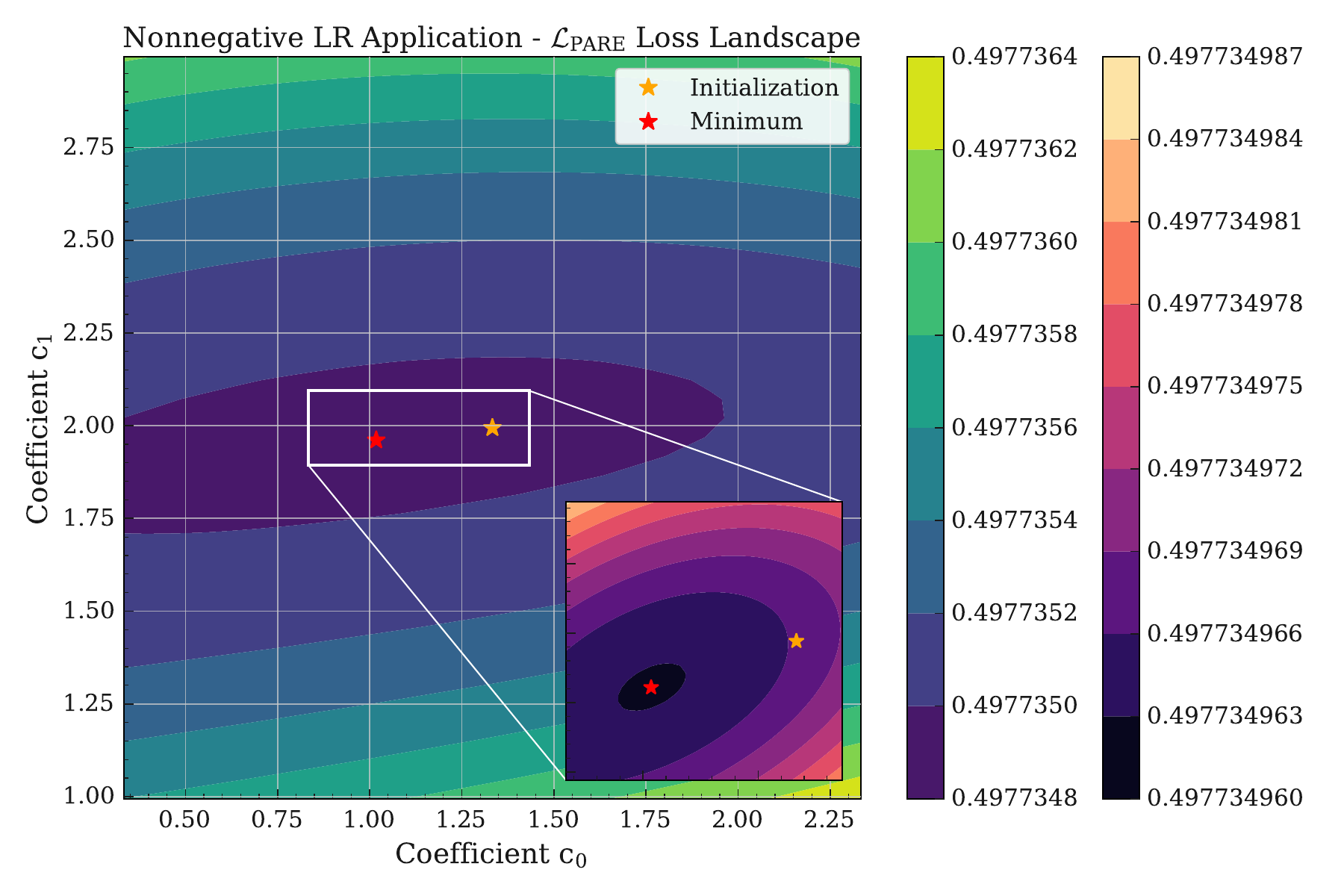}
\caption{The loss landscapes for optimizing the coefficients of the signed mixture model with the sub-ratio models fixed for the nonnegative likelihood ratio application. The left figure shows the MSE loss landscape and the right figure shows the PARE loss landscape.}
\end{figure}

\begin{figure}[H]
\centering
\includegraphics[scale=0.23]{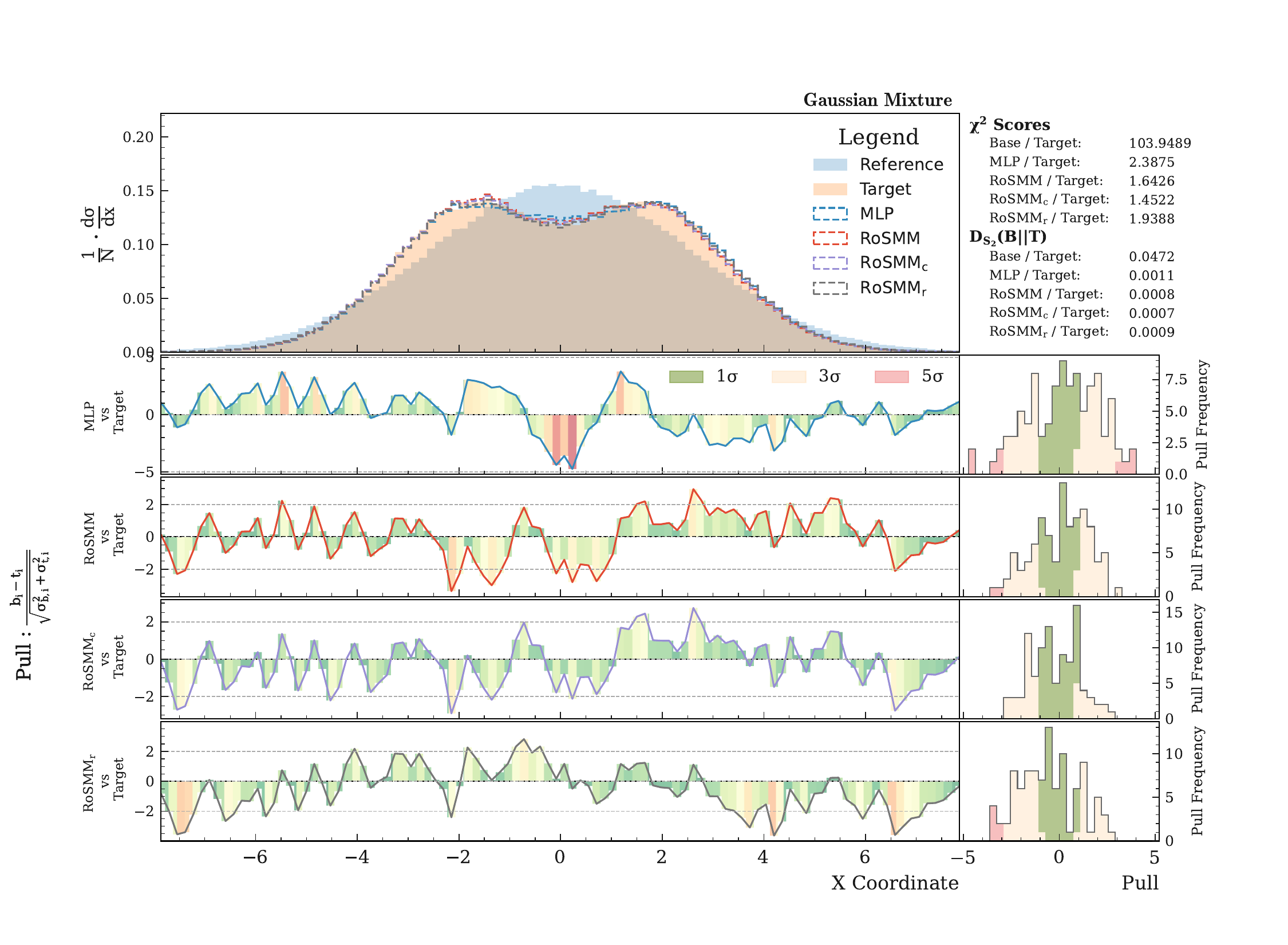}
\includegraphics[scale=0.23]{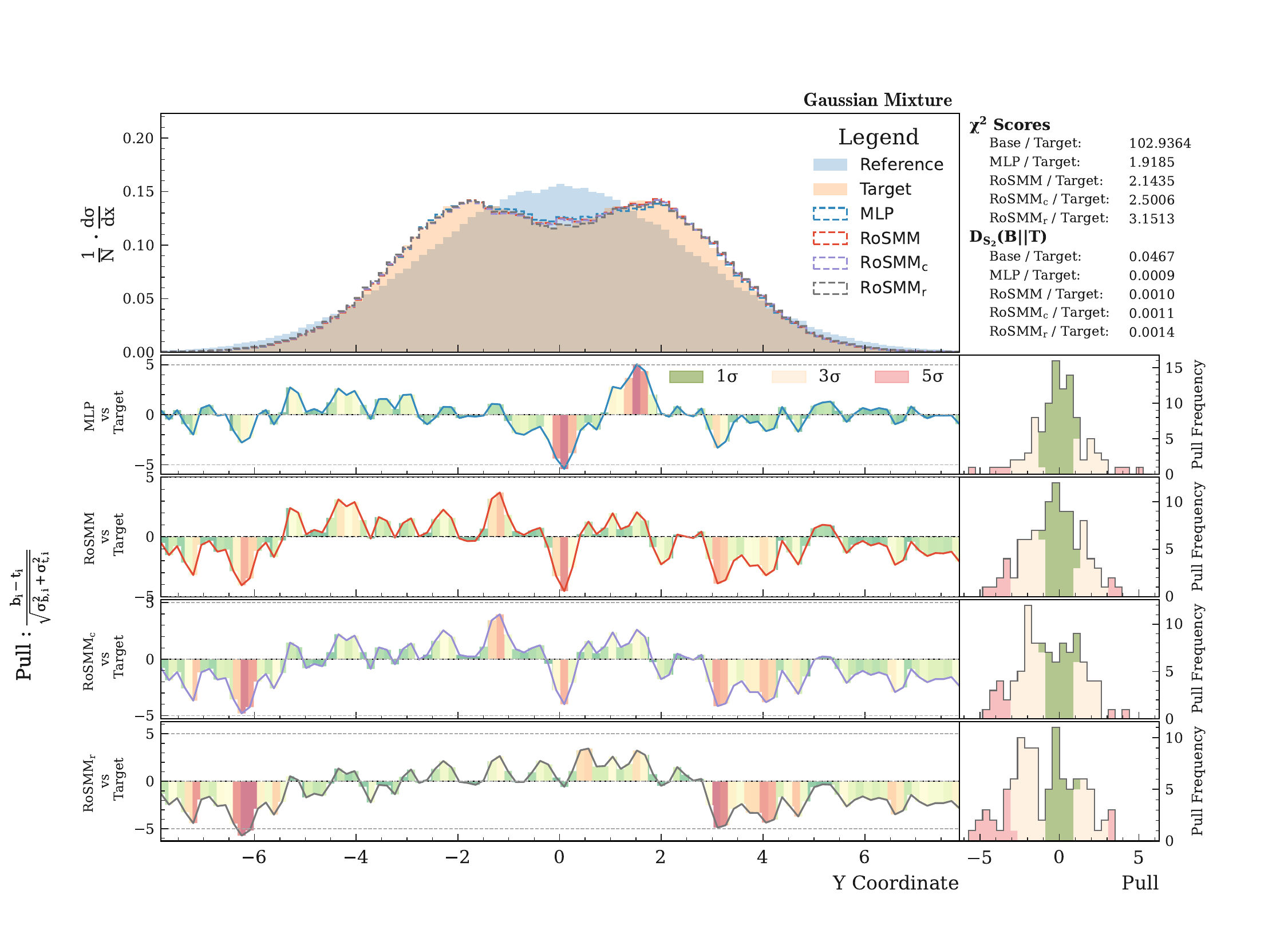}
\includegraphics[scale=0.23]{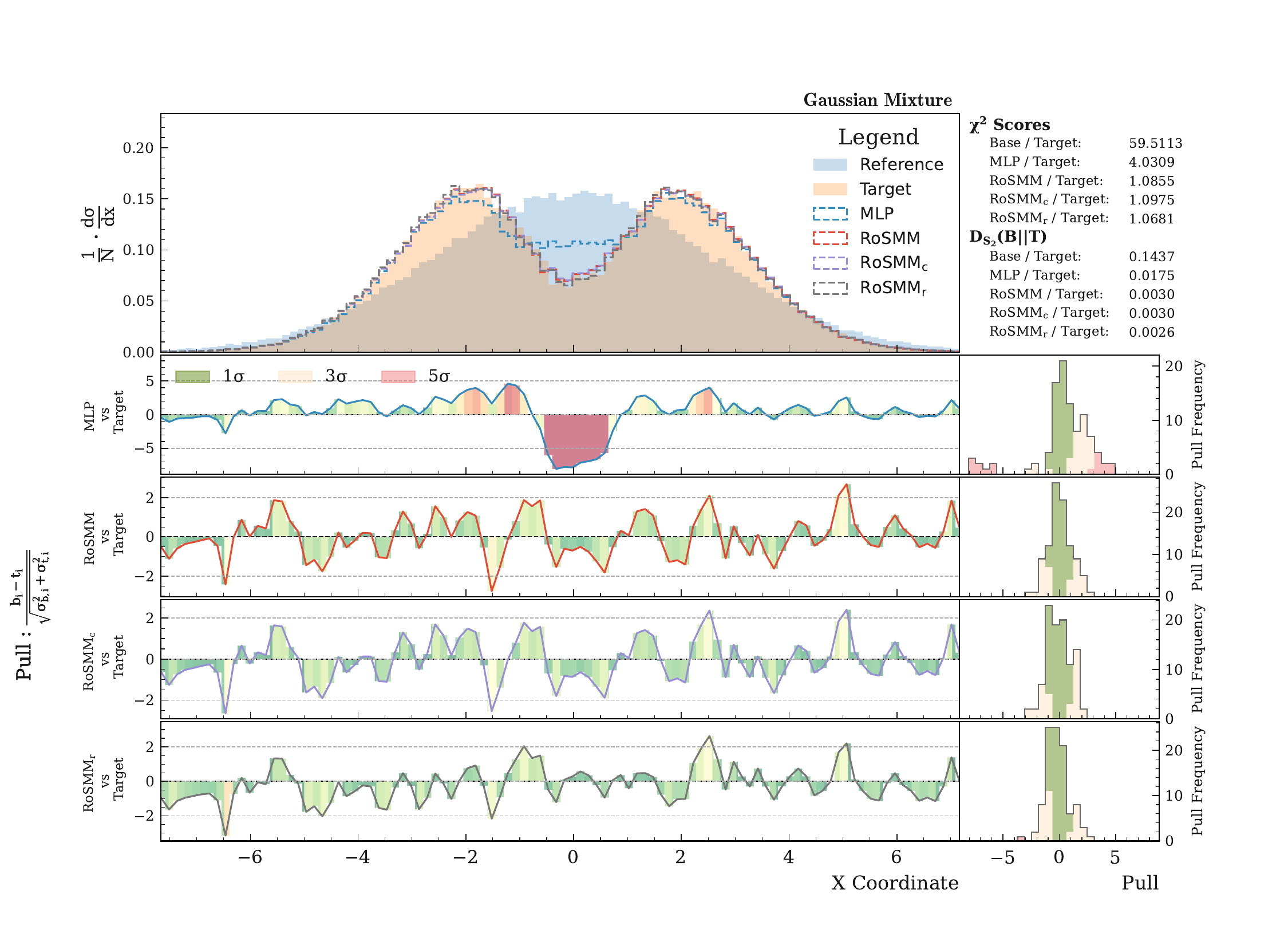}
\includegraphics[scale=0.23]{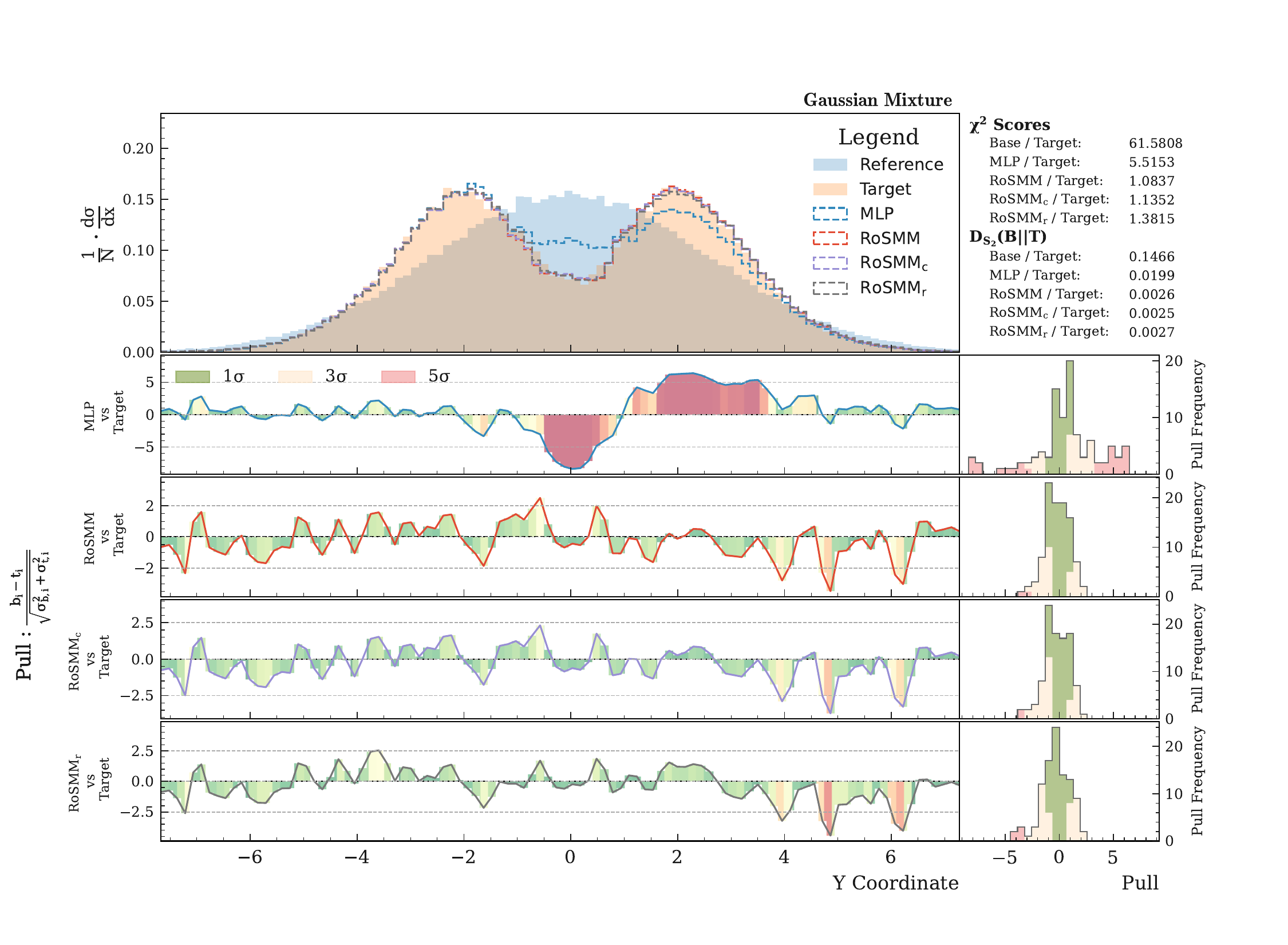}
\caption{Top row: reweighting closure plots for coordinates $X$ and $Y$ on the nonnegative Gaussian mixture model dataset using the different ratio estimation models. Bottom row: reweighting closure plots for coordinates $X$ and $Y$ on the signed Gaussian mixture model dataset using the different ratio estimation models.}
\end{figure}

\begin{figure}[H]
\centering
\includegraphics[scale=0.23]{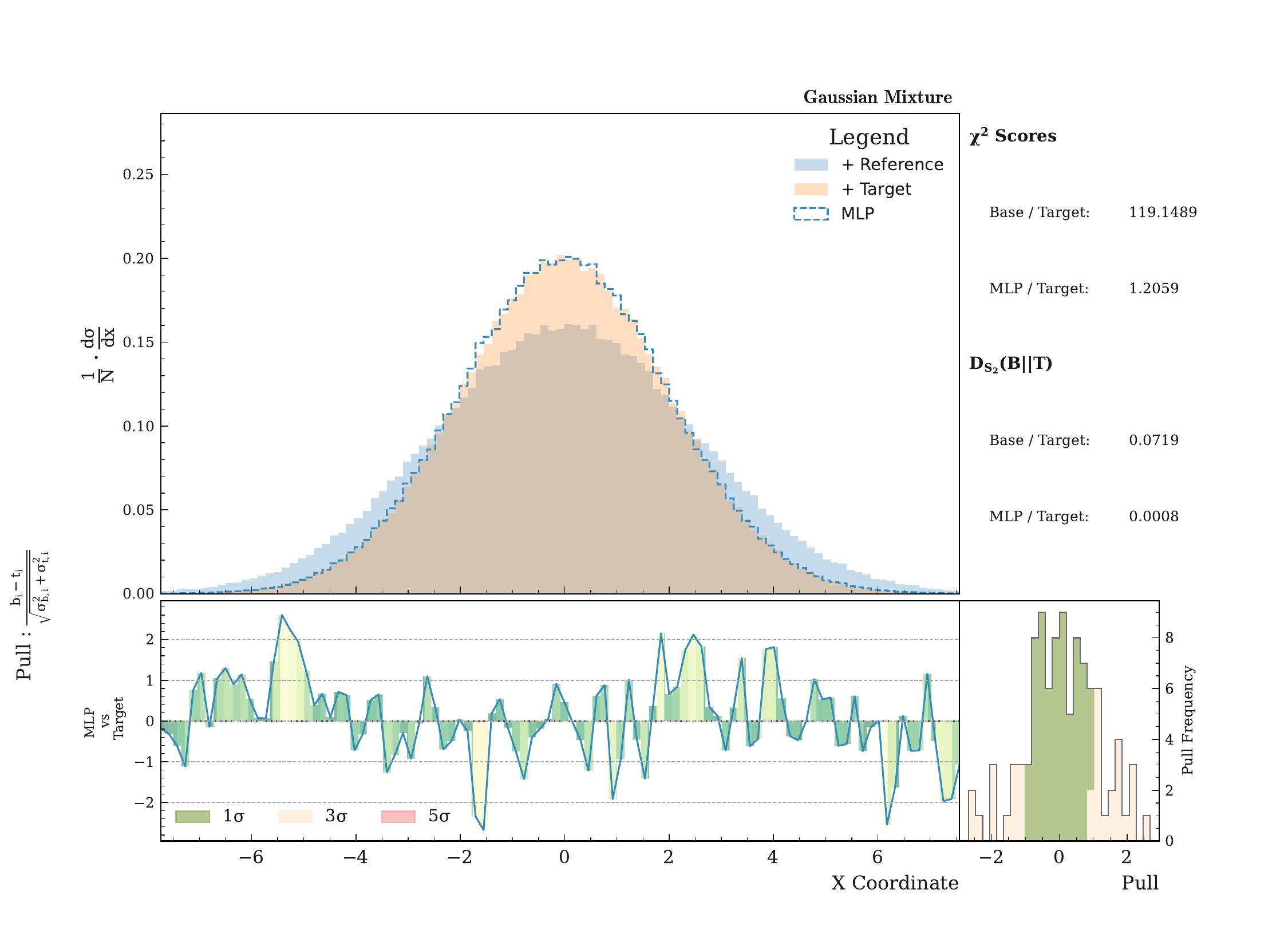}
\includegraphics[scale=0.23]{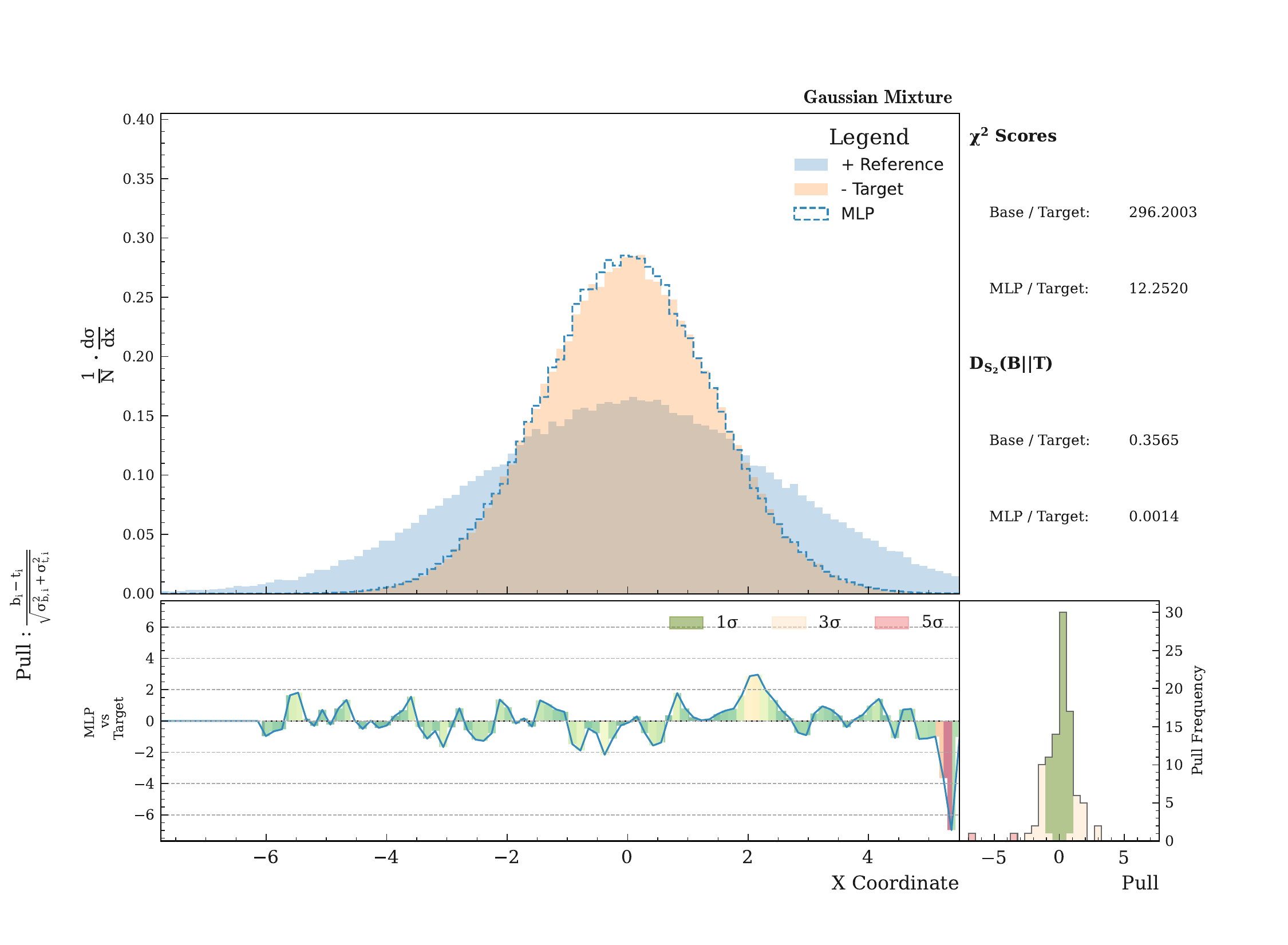}
\includegraphics[scale=0.23]{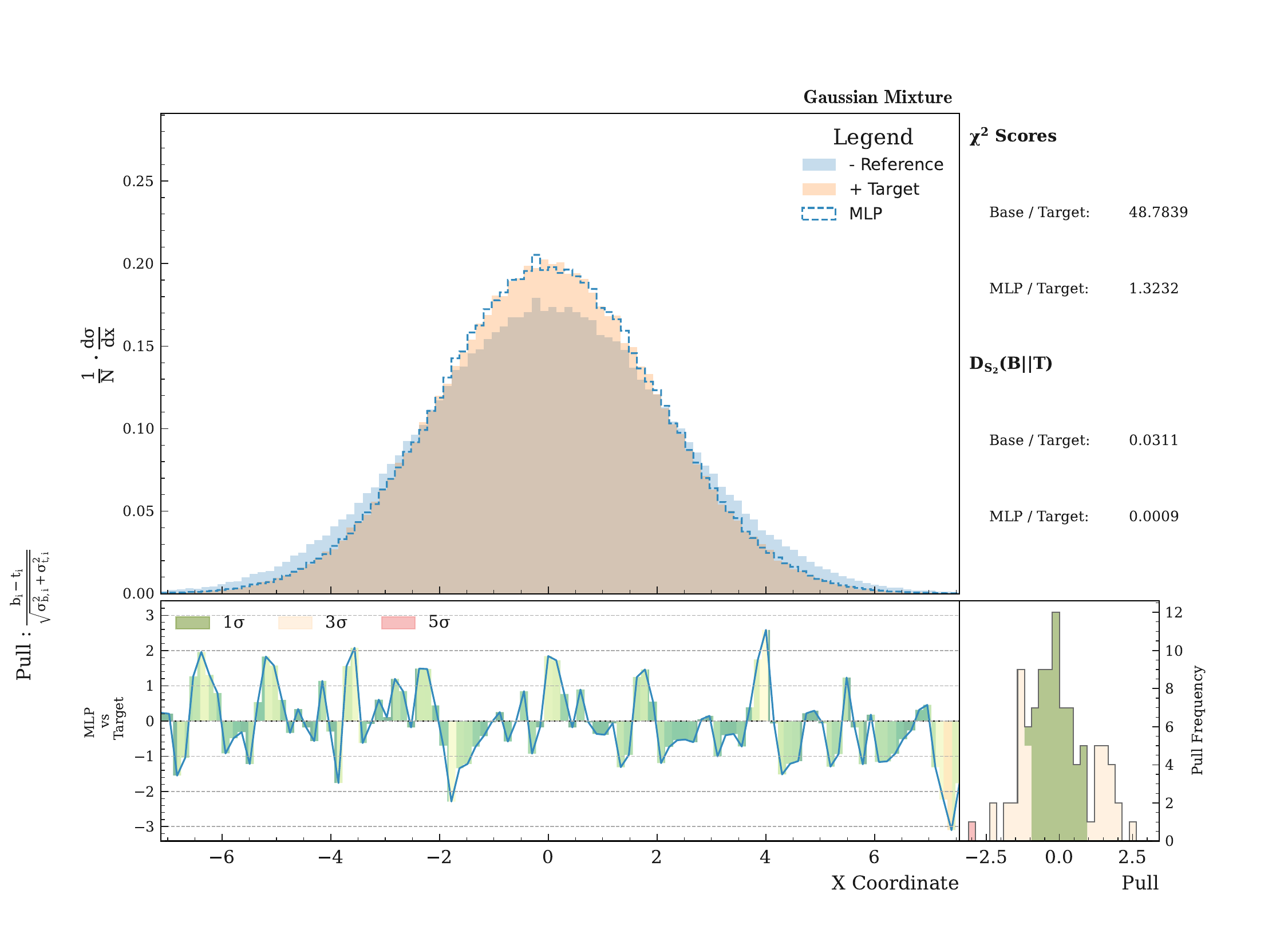}
\includegraphics[scale=0.23]{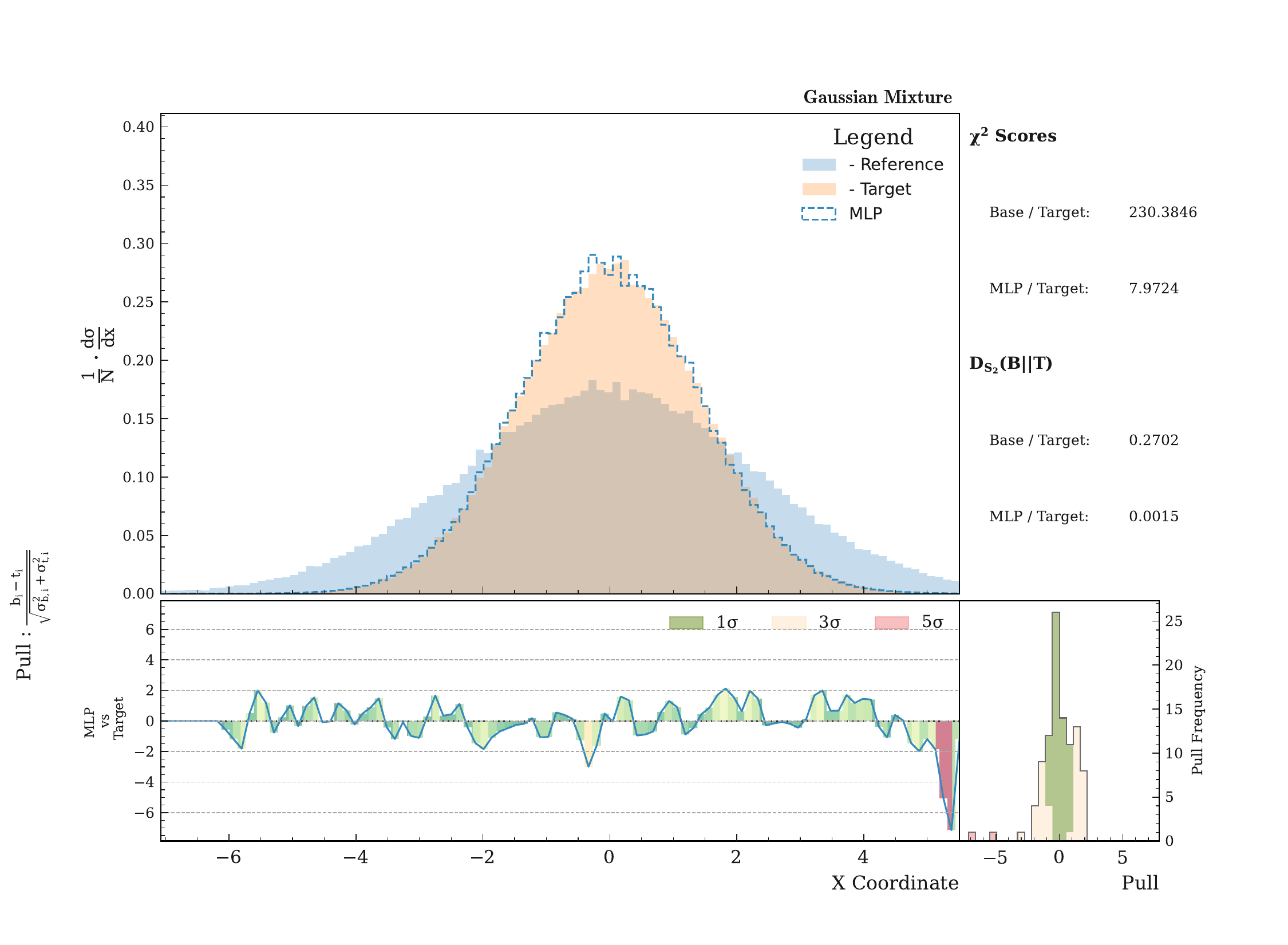}
\caption{Reweighting closure plots for the $X$-coordinate on the nonnegative Gaussian mixture model dataset for the different signed mixture model subdensity ratio estimation datasets and models $r_{\pm\pm}$.}
\end{figure}

\begin{figure}[H]
\centering
\includegraphics[scale=0.23]{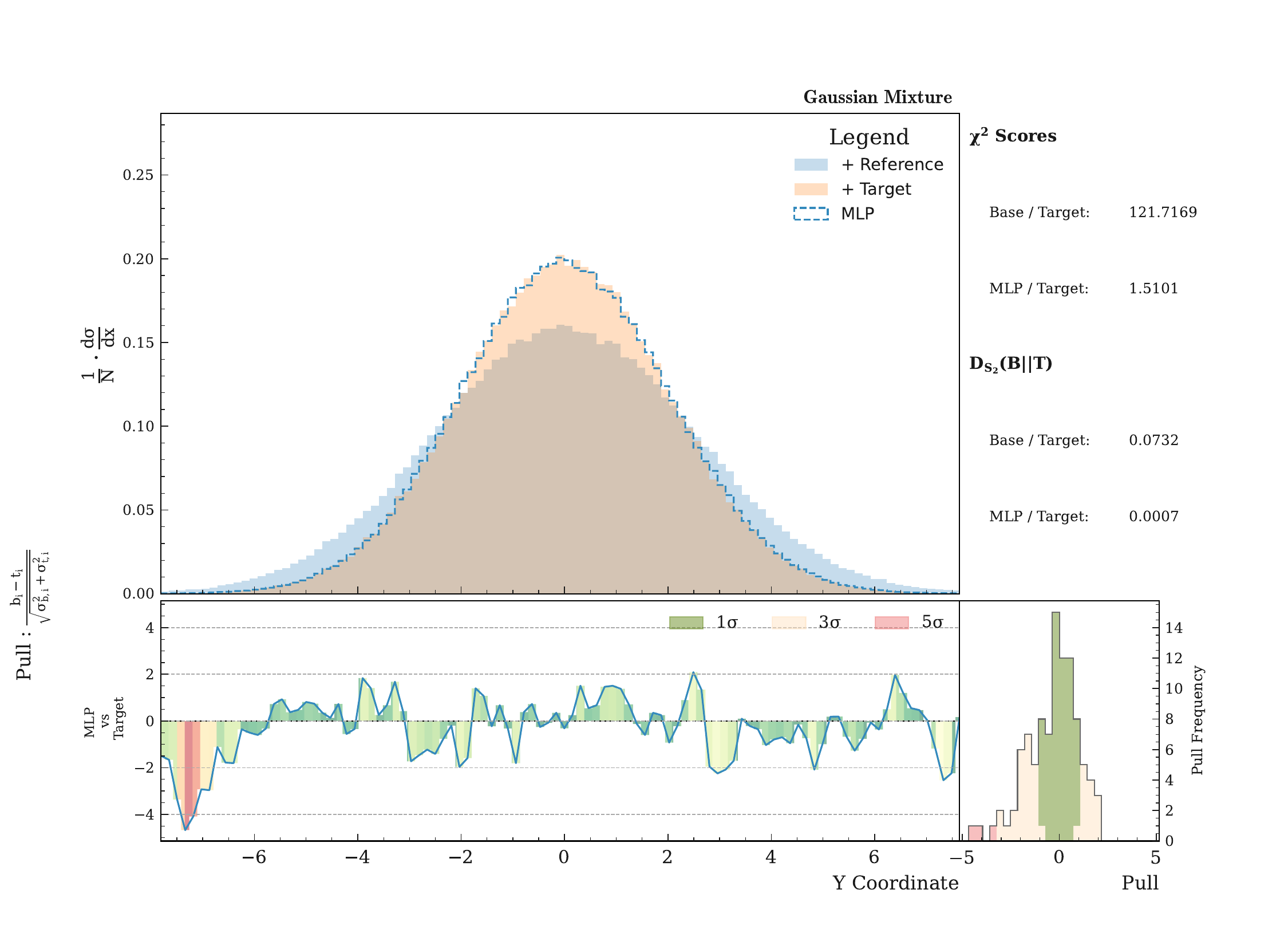}
\includegraphics[scale=0.23]{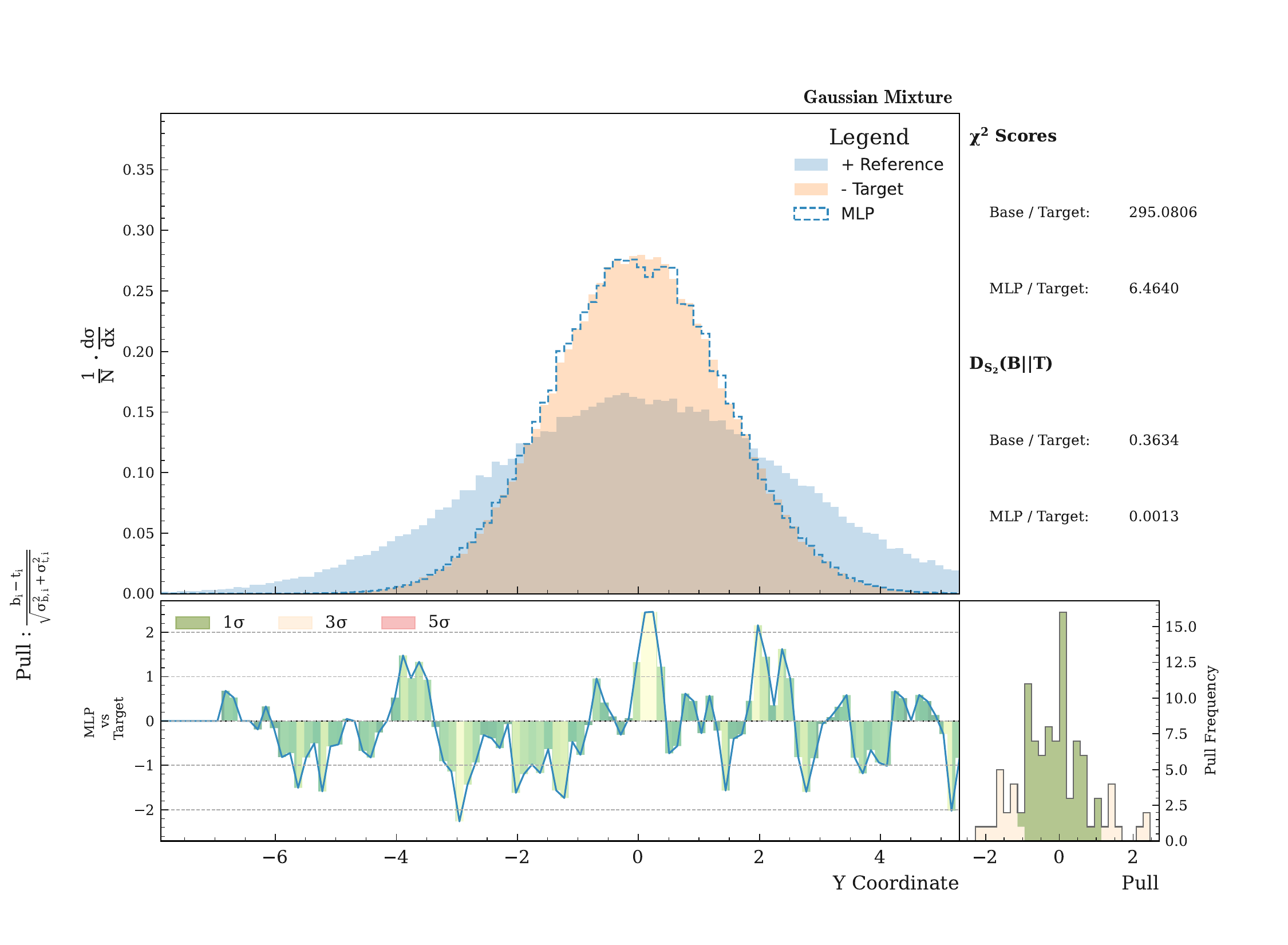}
\includegraphics[scale=0.23]{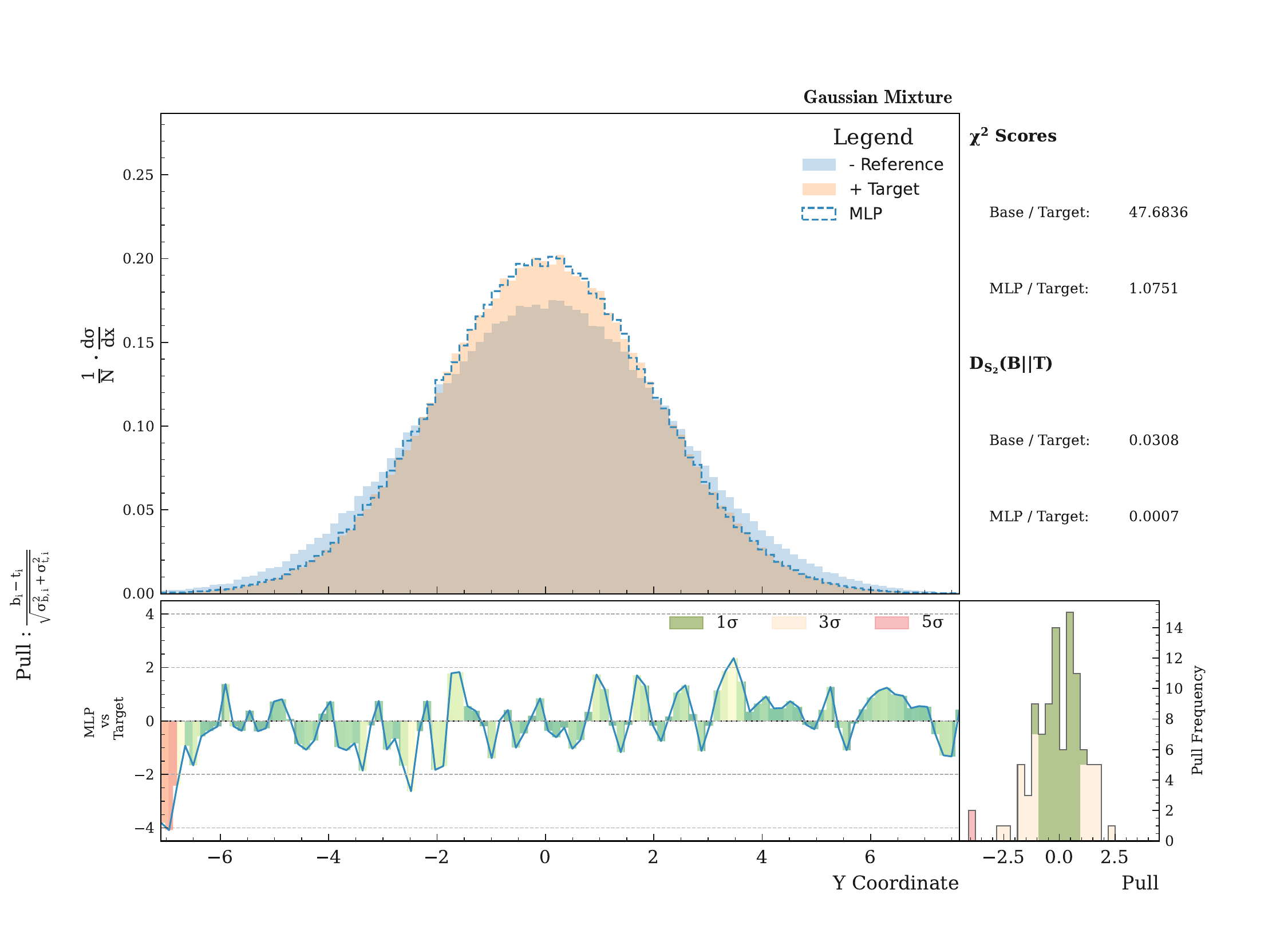}
\includegraphics[scale=0.23]{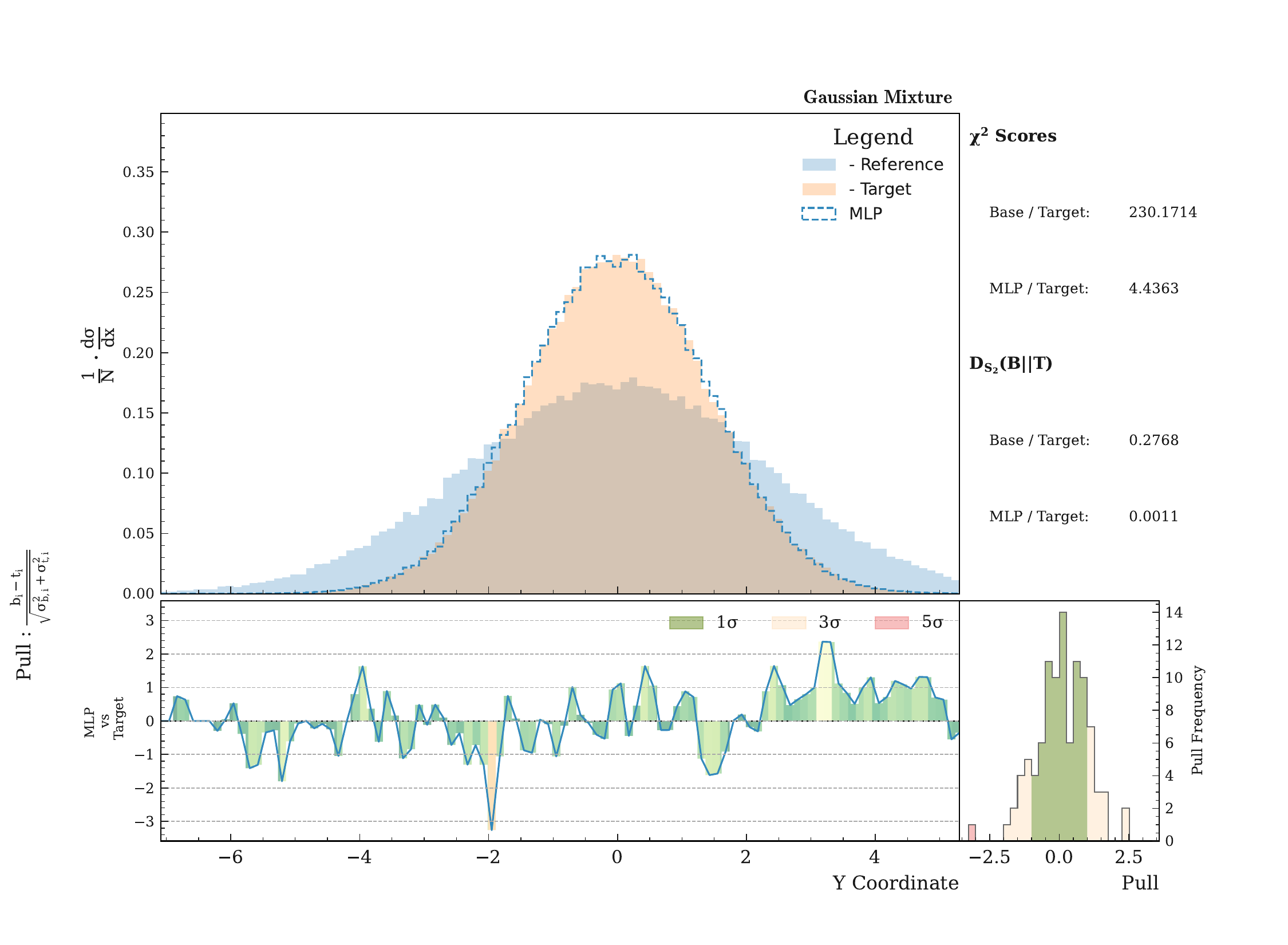}
\caption{Reweighting closure plots for the $Y$-coordinate on the nonnegative Gaussian mixture model dataset for the different signed mixture model subdensity ratio estimation datasets and models $r_{\pm\pm}$.}
\end{figure}

\begin{figure}[H]
\centering
\includegraphics[scale=0.23]{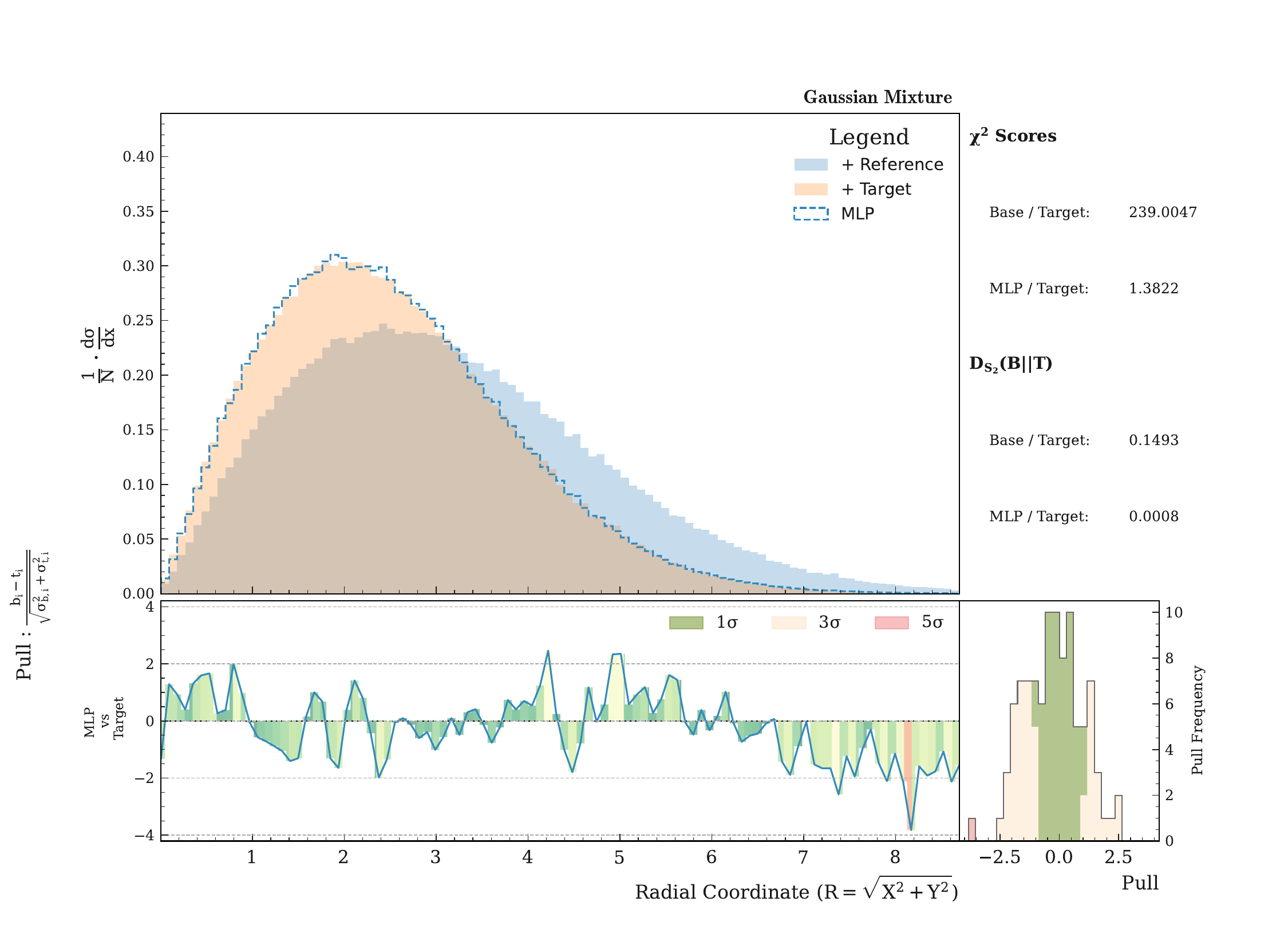}
\includegraphics[scale=0.23]{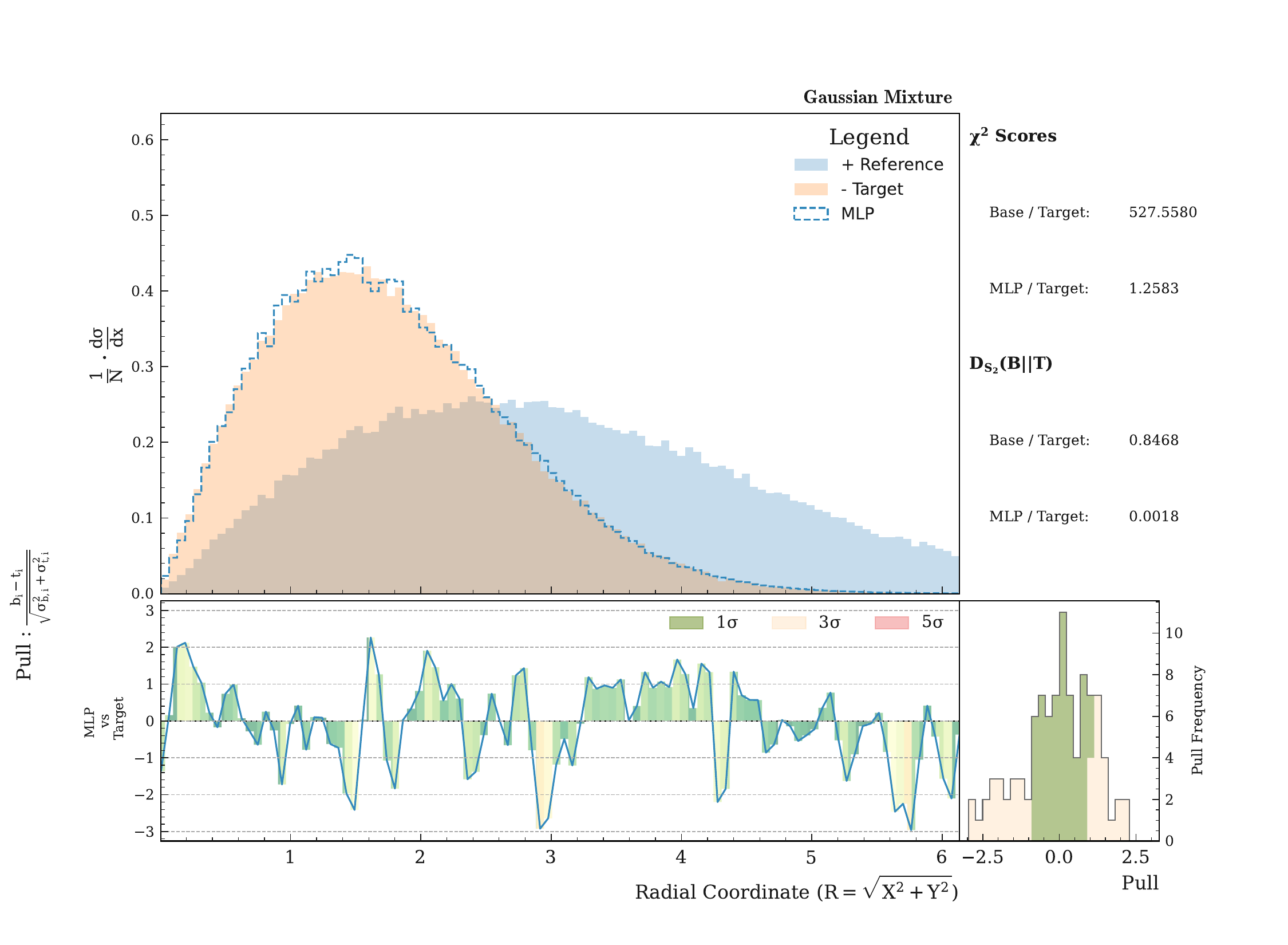}
\includegraphics[scale=0.23]{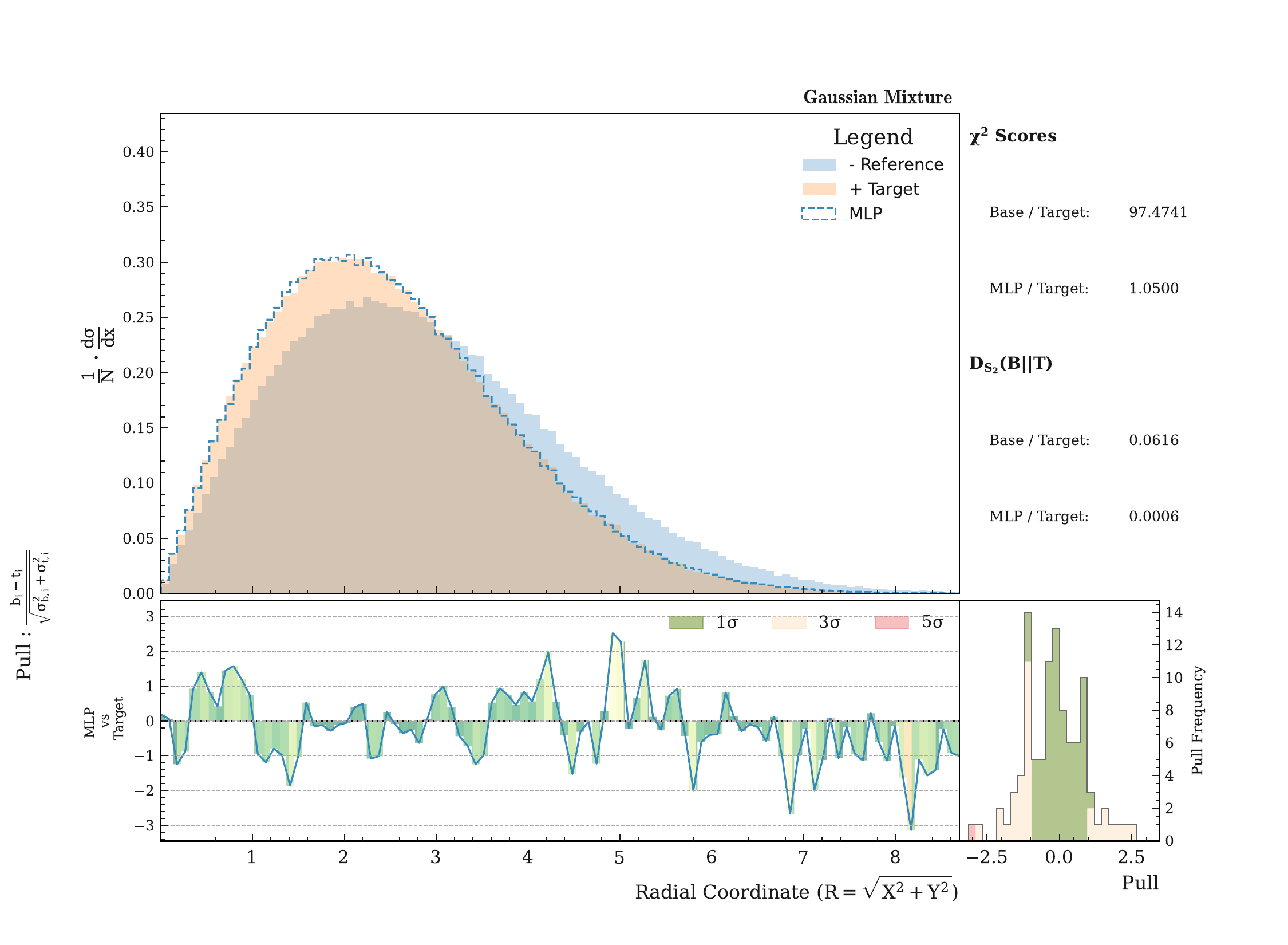}
\includegraphics[scale=0.23]{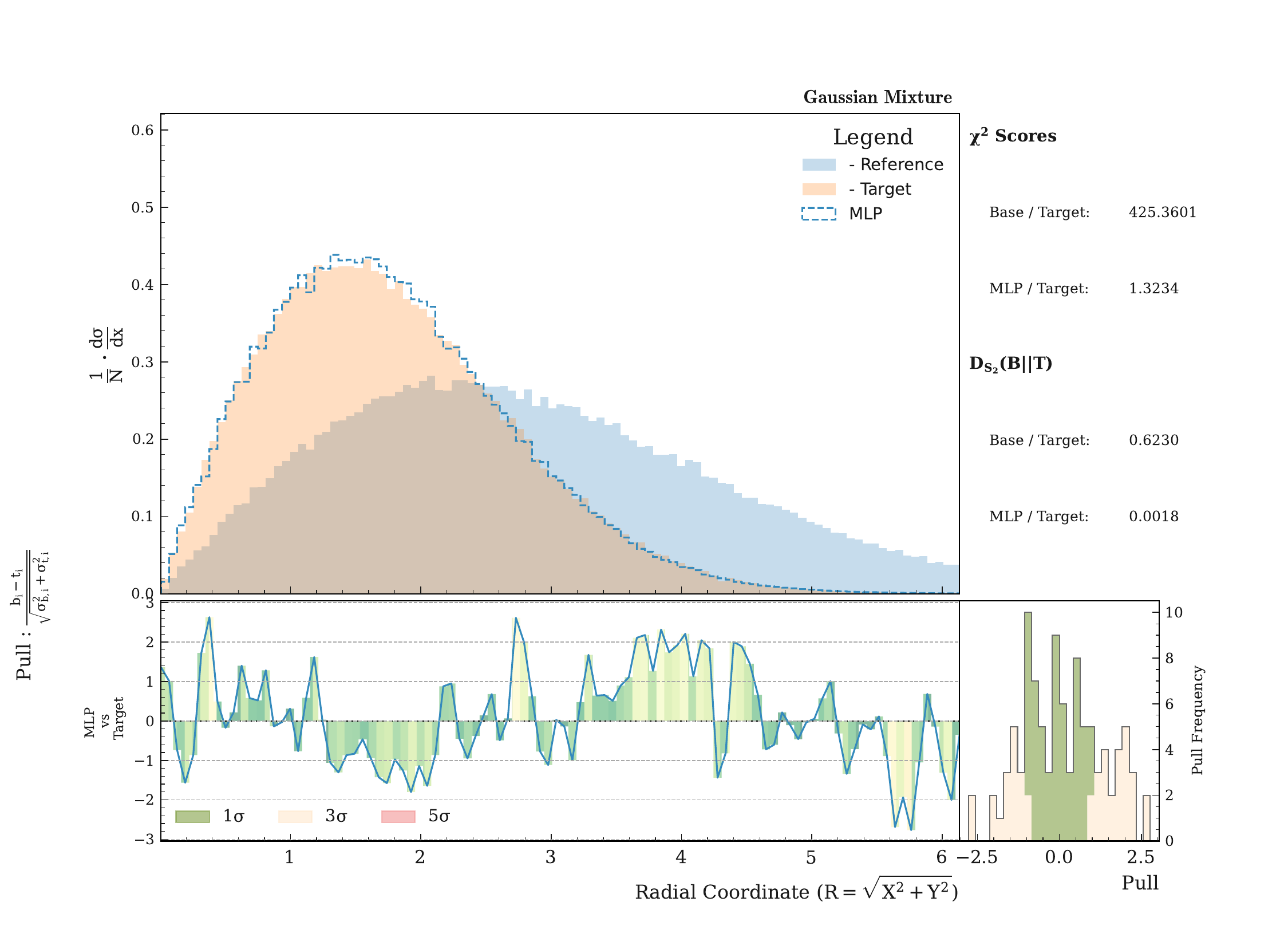}
\caption{Reweighting closure plots for the radial coordinate $R$ on the nonnegative Gaussian mixture model dataset for the different signed mixture model subdensity ratio estimation datasets and models $r_{\pm\pm}$.}
\end{figure}

\begin{figure}[H]
\centering
\includegraphics[scale=0.23]{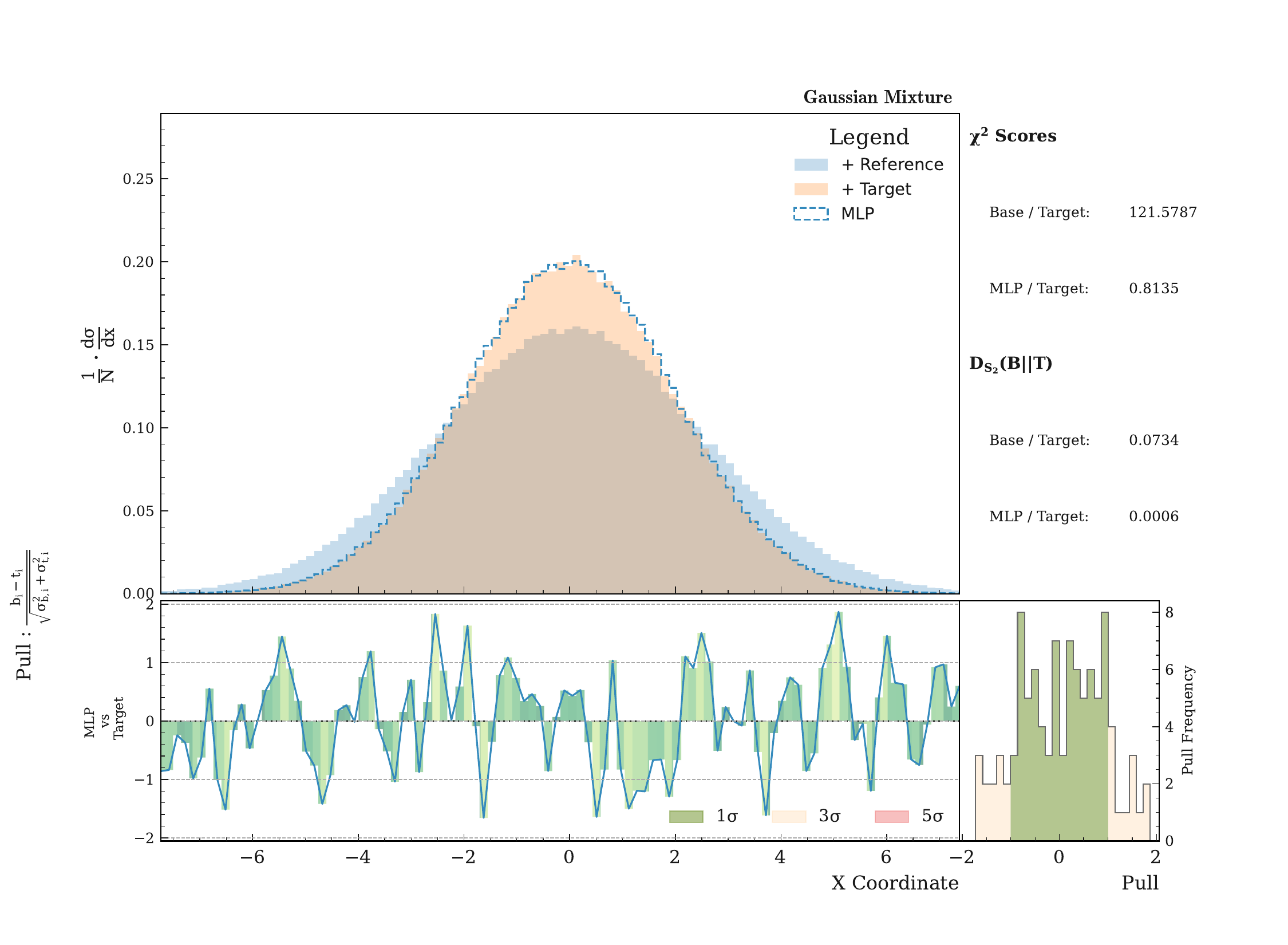}
\includegraphics[scale=0.23]{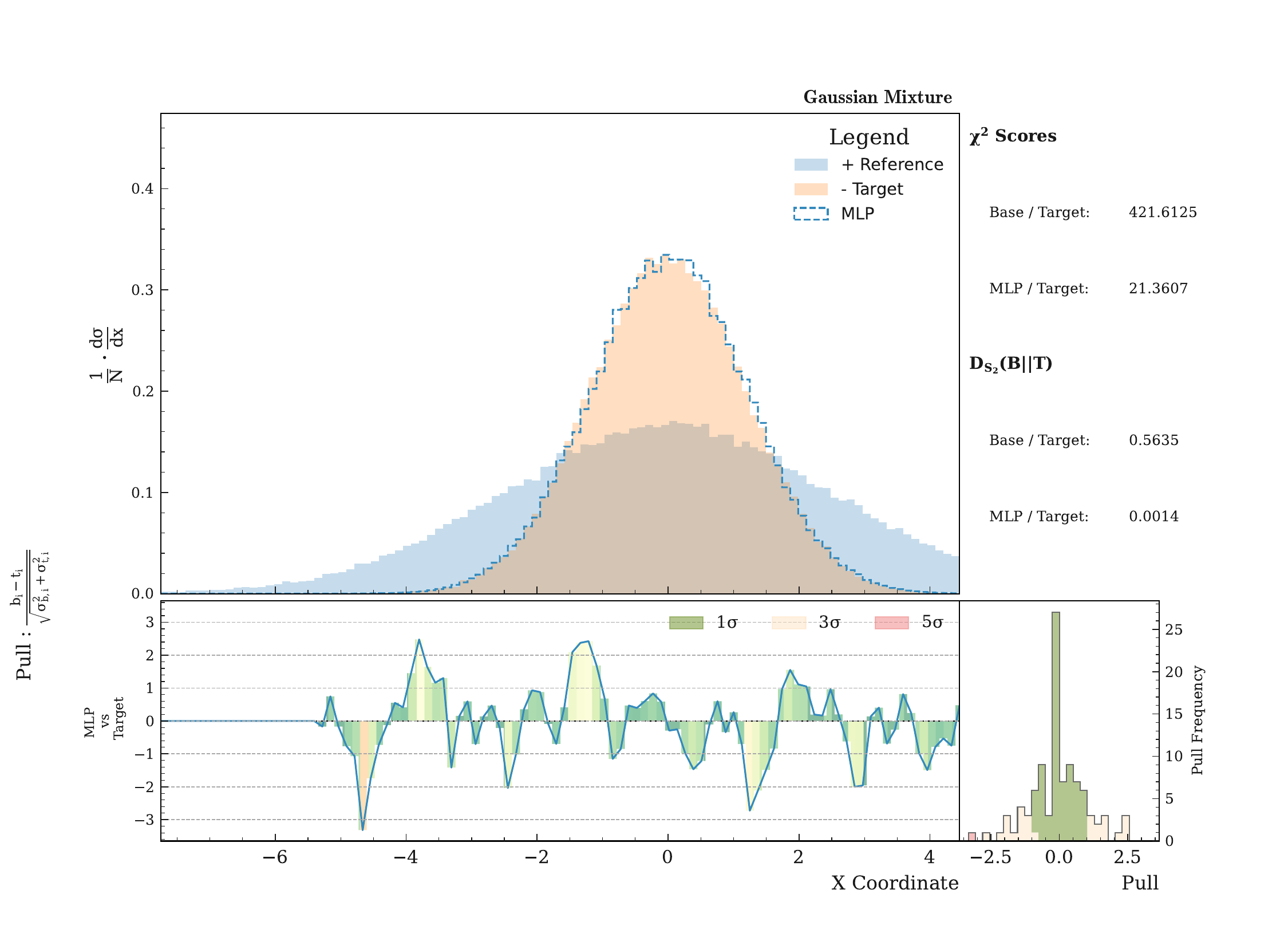}
\includegraphics[scale=0.23]{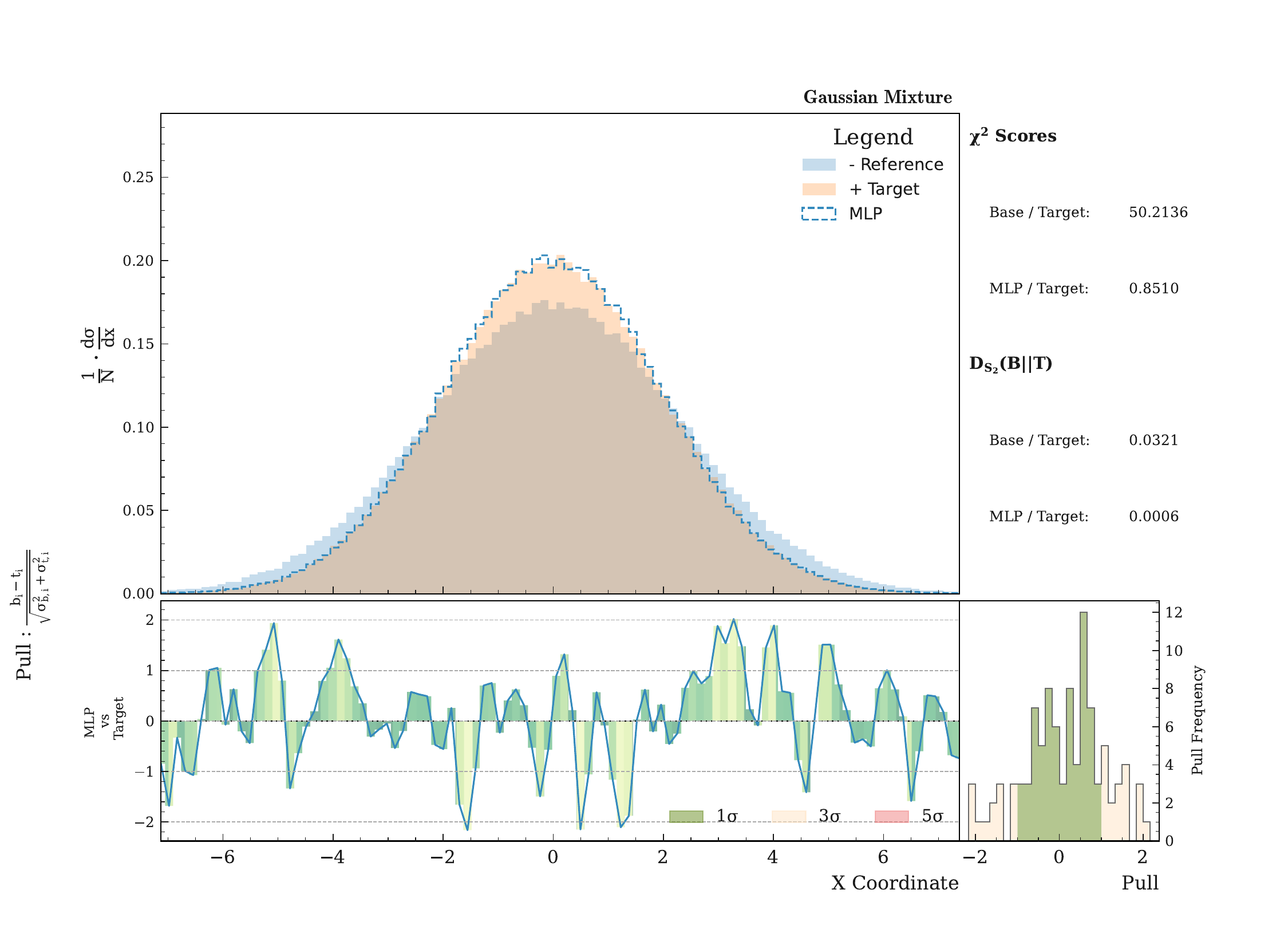}
\includegraphics[scale=0.23]{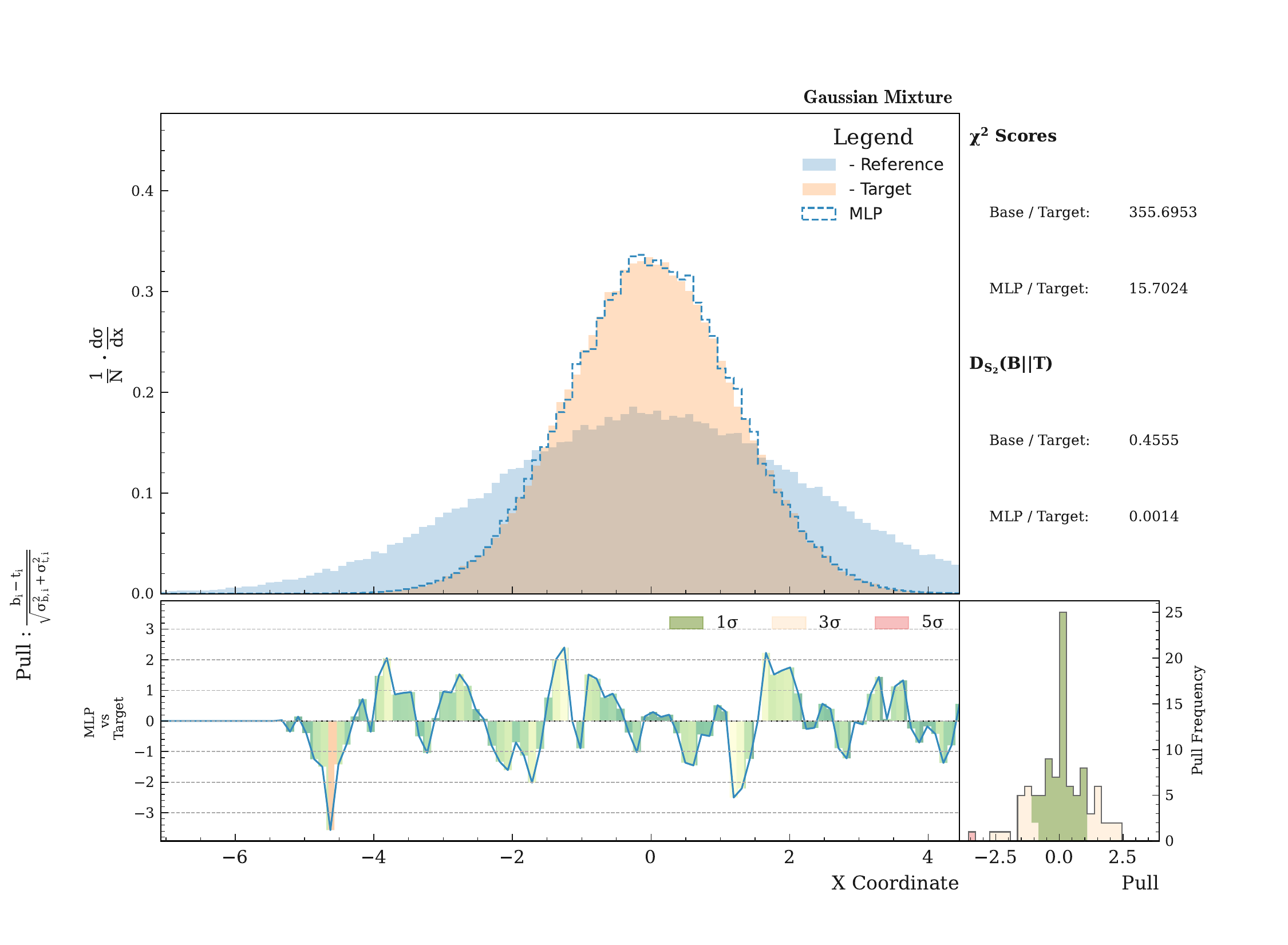}
\caption{Reweighting closure plots for the $X$-coordinate on the signed Gaussian mixture model dataset for the different signed mixture model subdensity ratio estimation datasets and models $r_{\pm\pm}$.}
\end{figure}

\begin{figure}[H]
\centering
\includegraphics[scale=0.23]{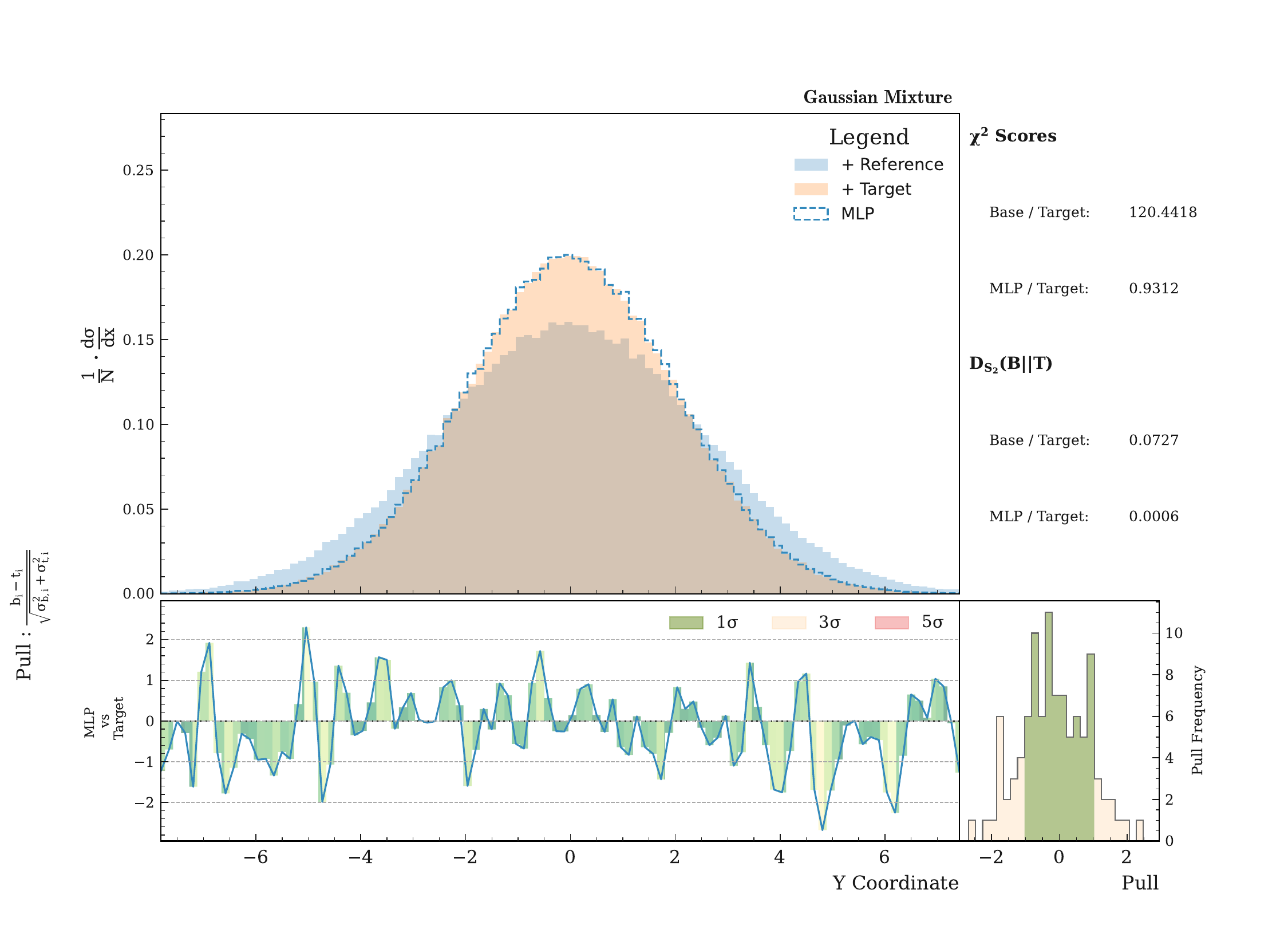}
\includegraphics[scale=0.23]{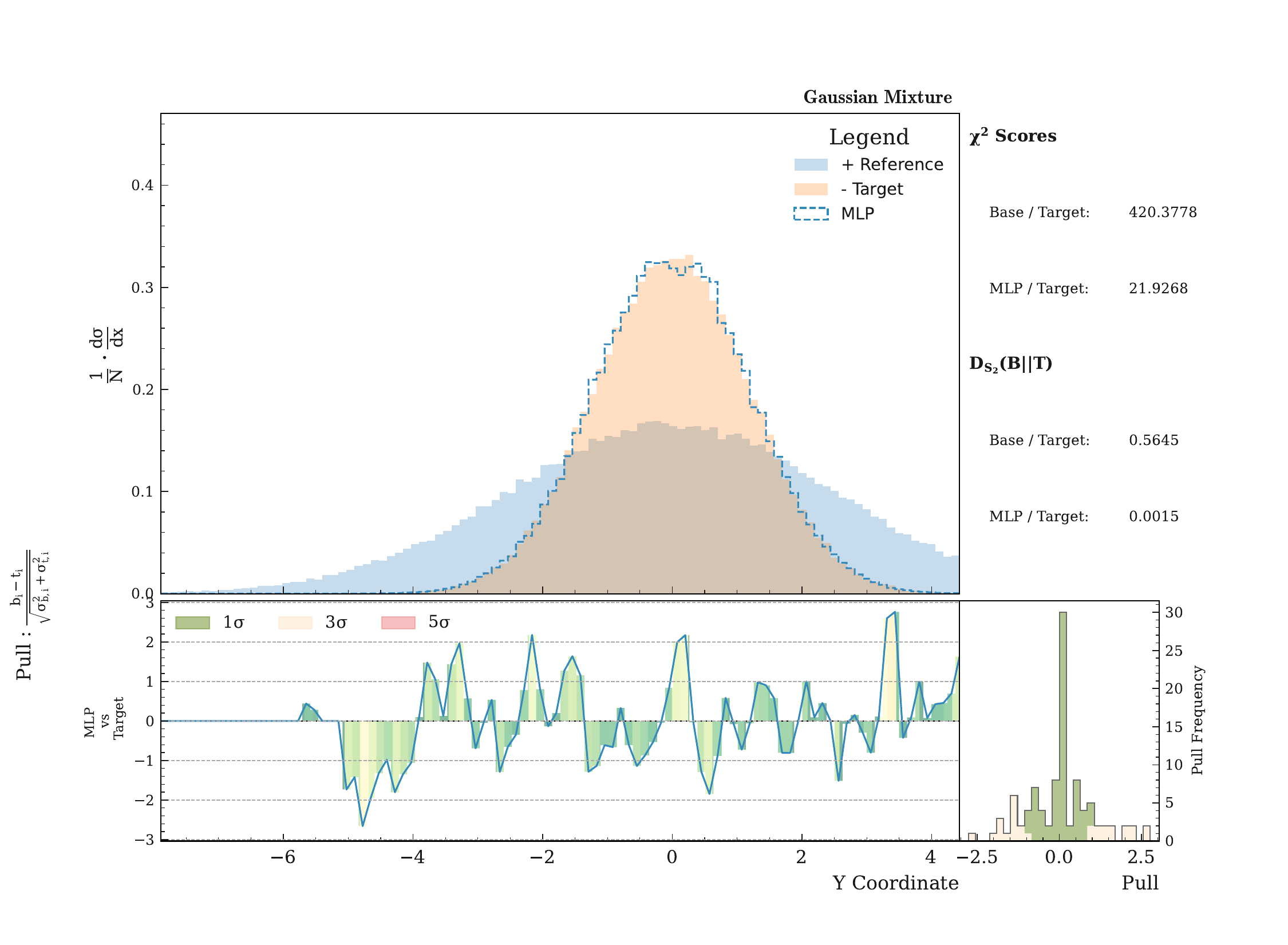}
\includegraphics[scale=0.23]{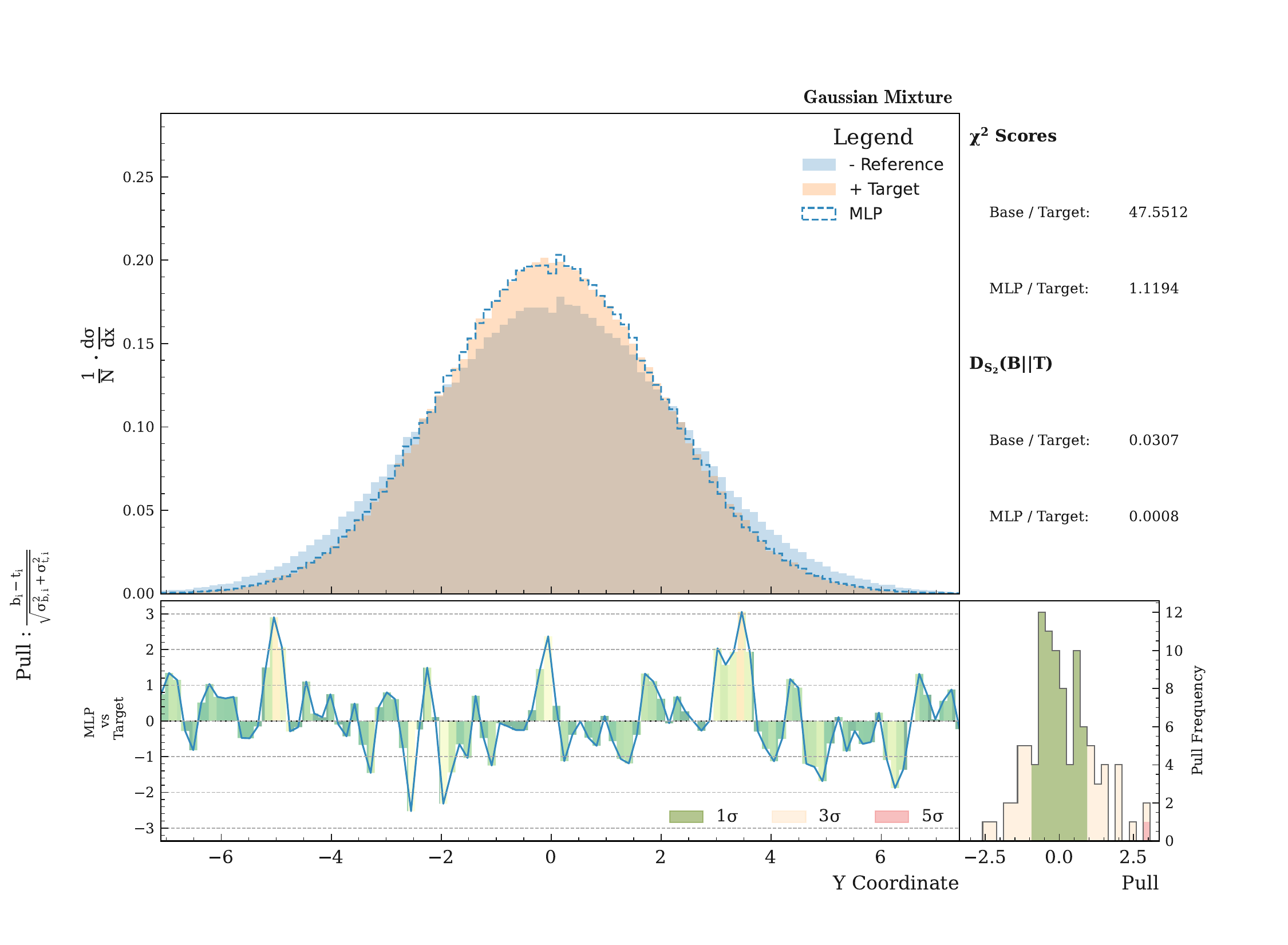}
\includegraphics[scale=0.23]{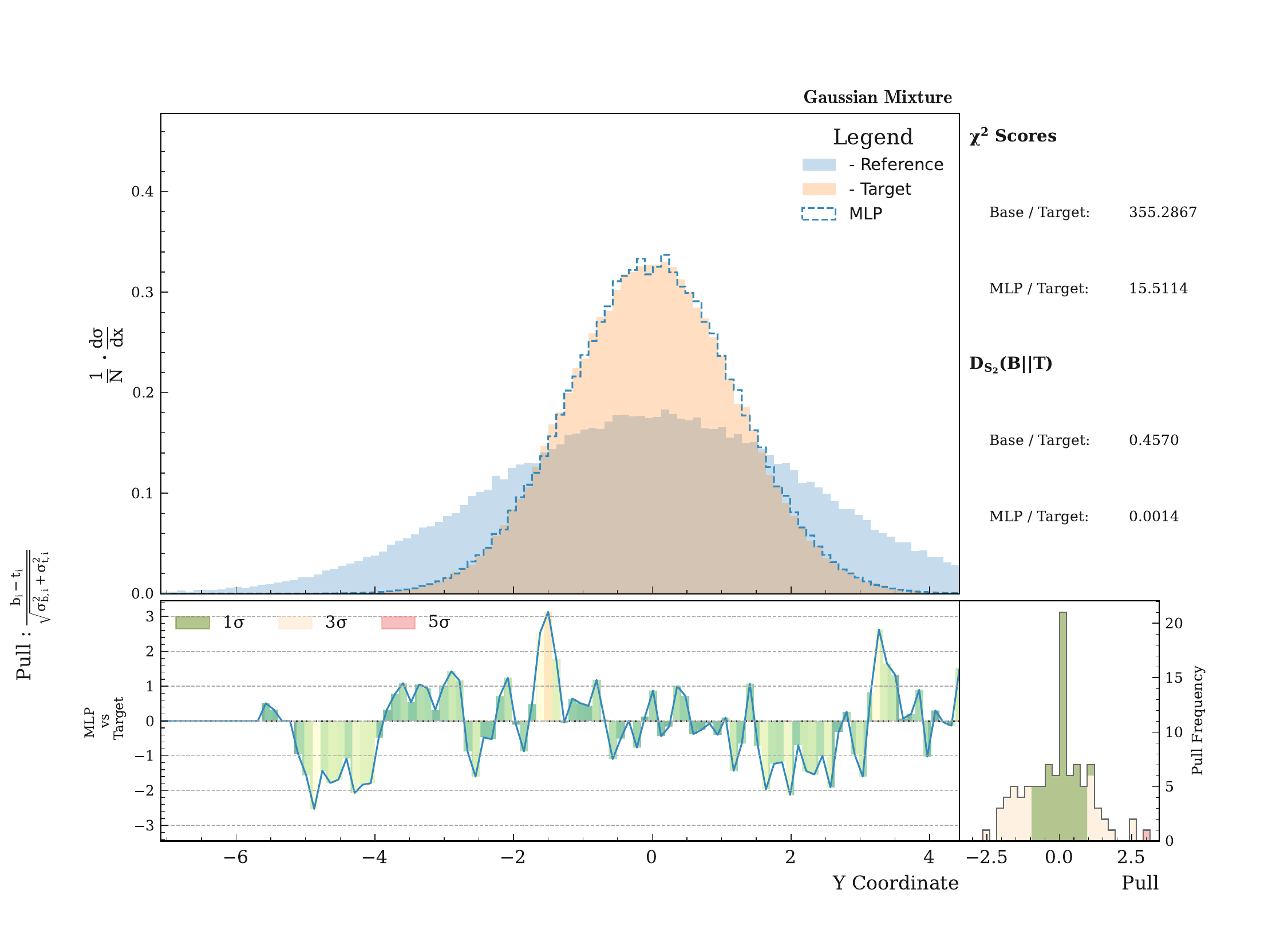}
\caption{Reweighting closure plots for the $Y$-coordinate on the signed Gaussian mixture model dataset for the different signed mixture model subdensity ratio estimation datasets and models $r_{\pm\pm}$.}
\end{figure}

\begin{figure}[H]
\centering
\includegraphics[scale=0.23]{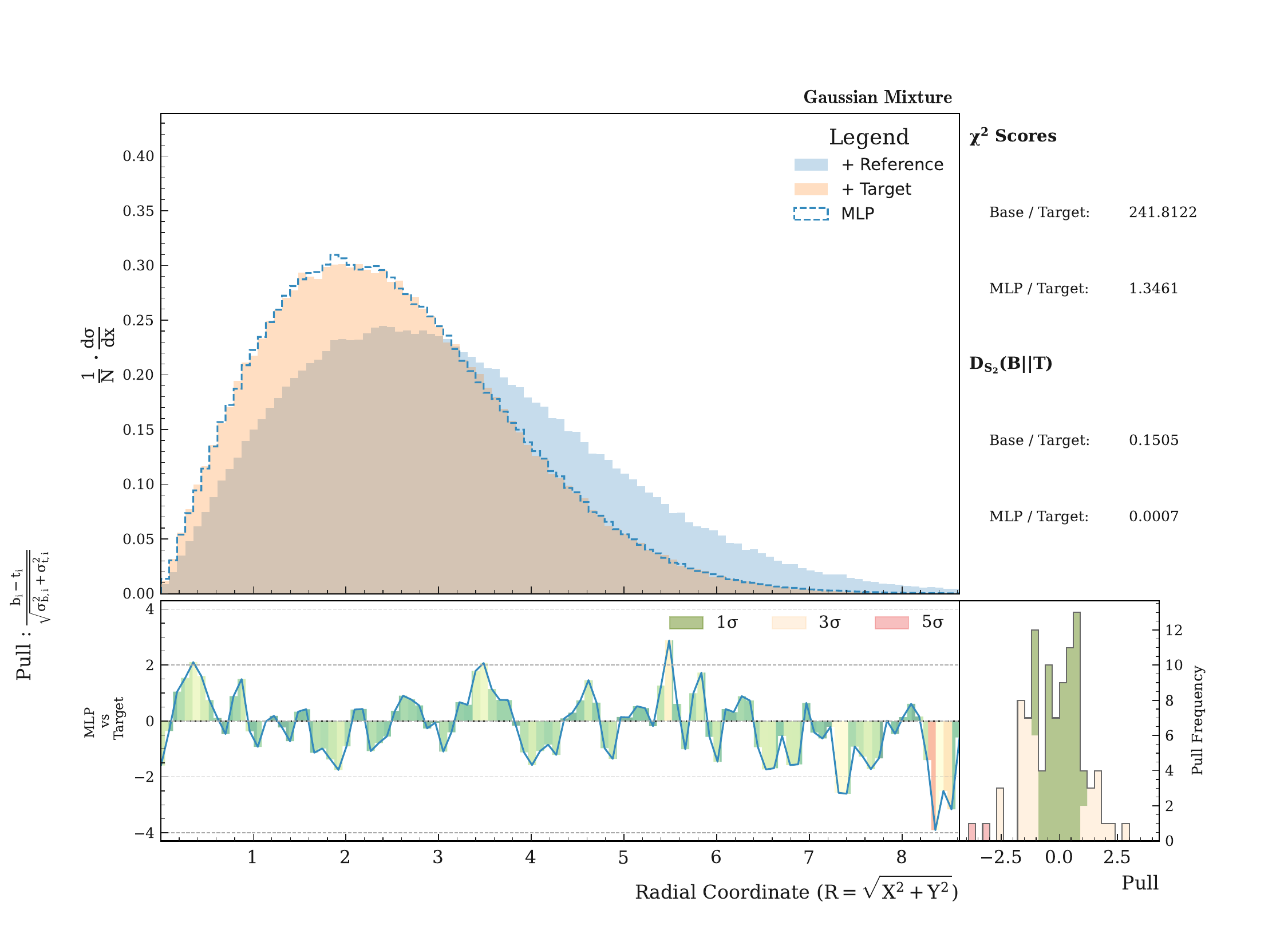}
\includegraphics[scale=0.23]{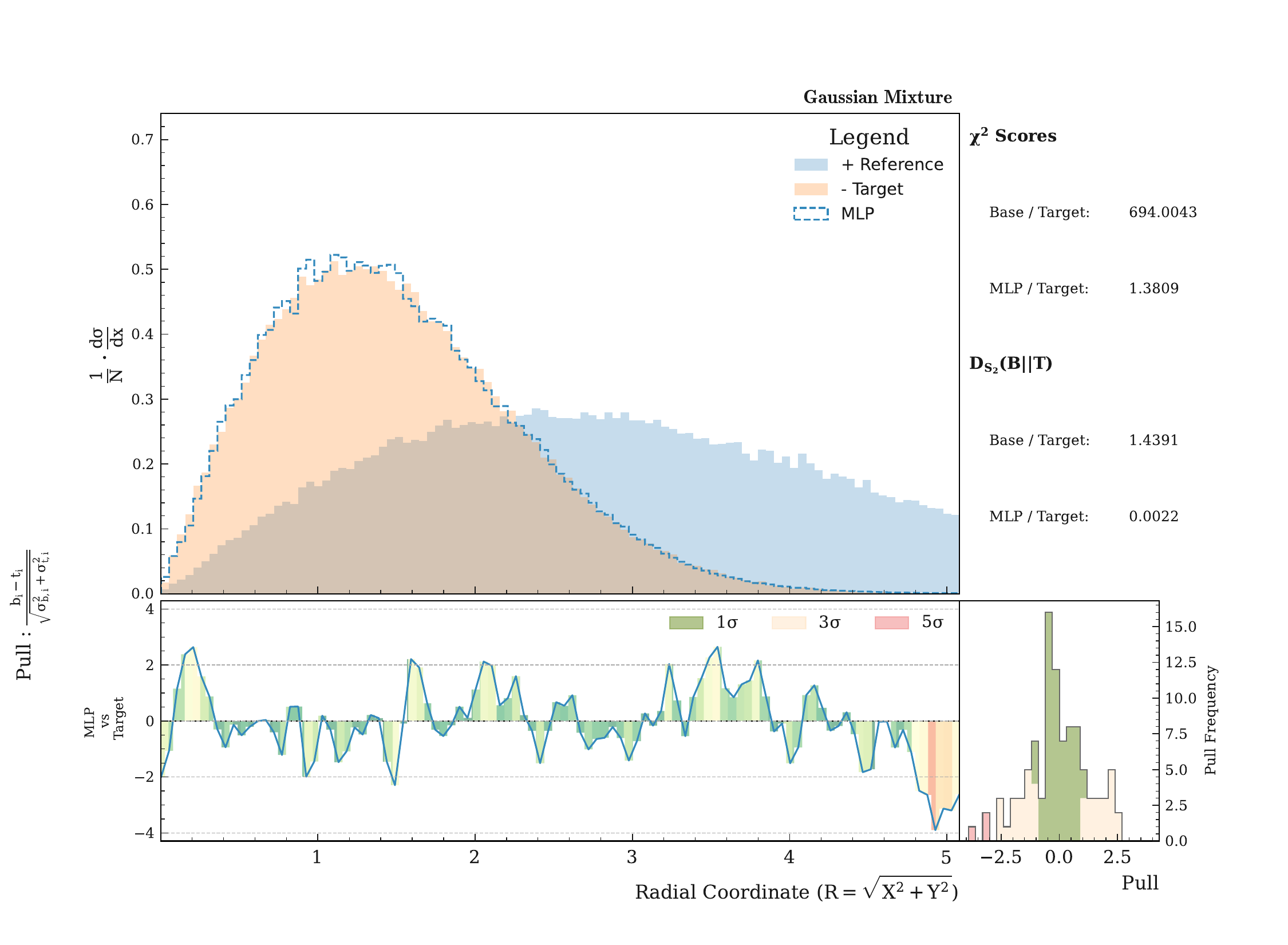}
\includegraphics[scale=0.23]{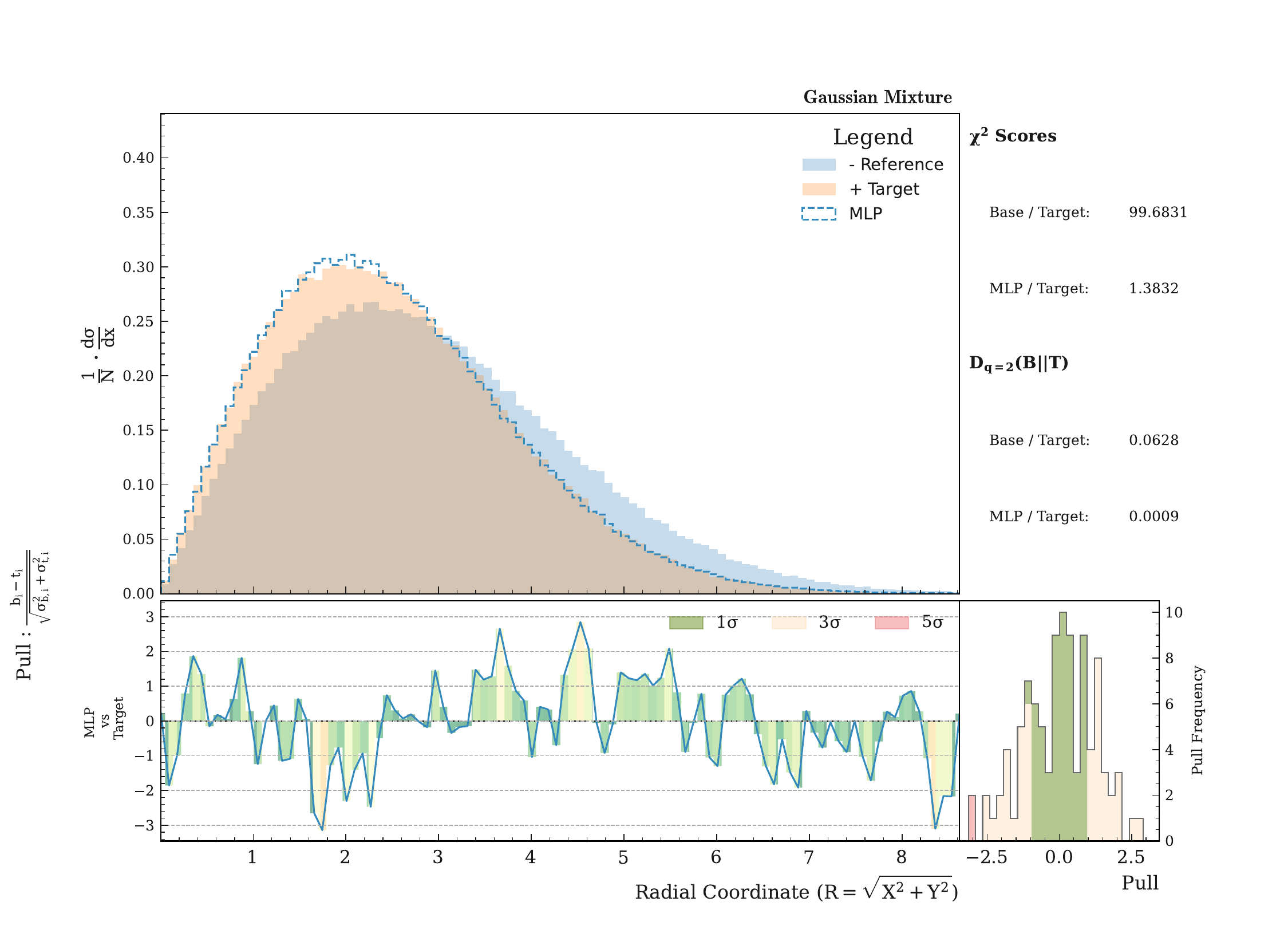}
\includegraphics[scale=0.23]{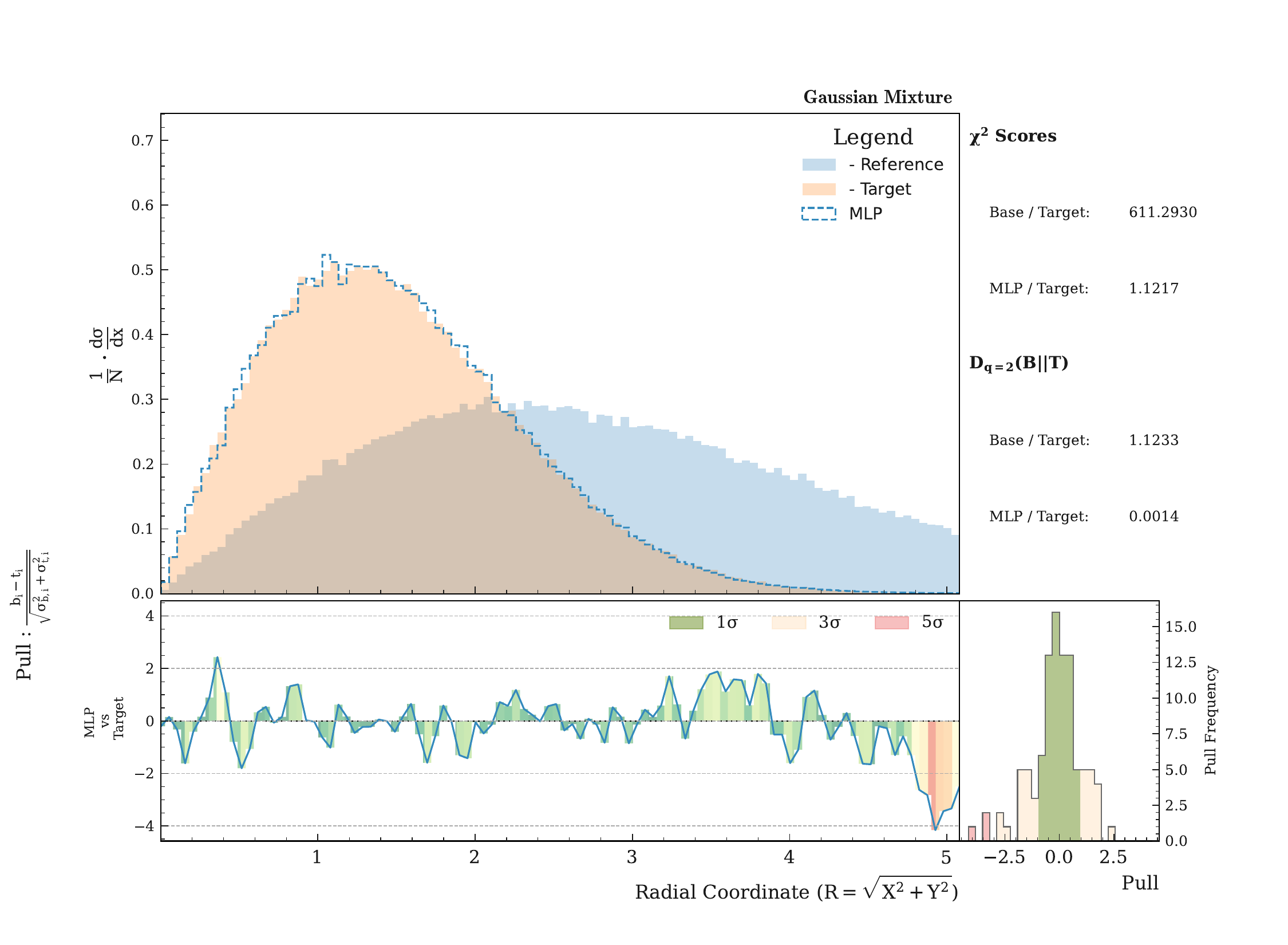}
\caption{Reweighting closure plots for the radial coordinate $R$ on the signed Gaussian mixture model dataset for the different signed mixture model subdensity ratio estimation datasets and models $r_{\pm\pm}$.}
\end{figure}


\subsection{Additional Plots: Application to Particle Physics}

\begin{figure}[H]
\centering
\includegraphics[scale=0.34]{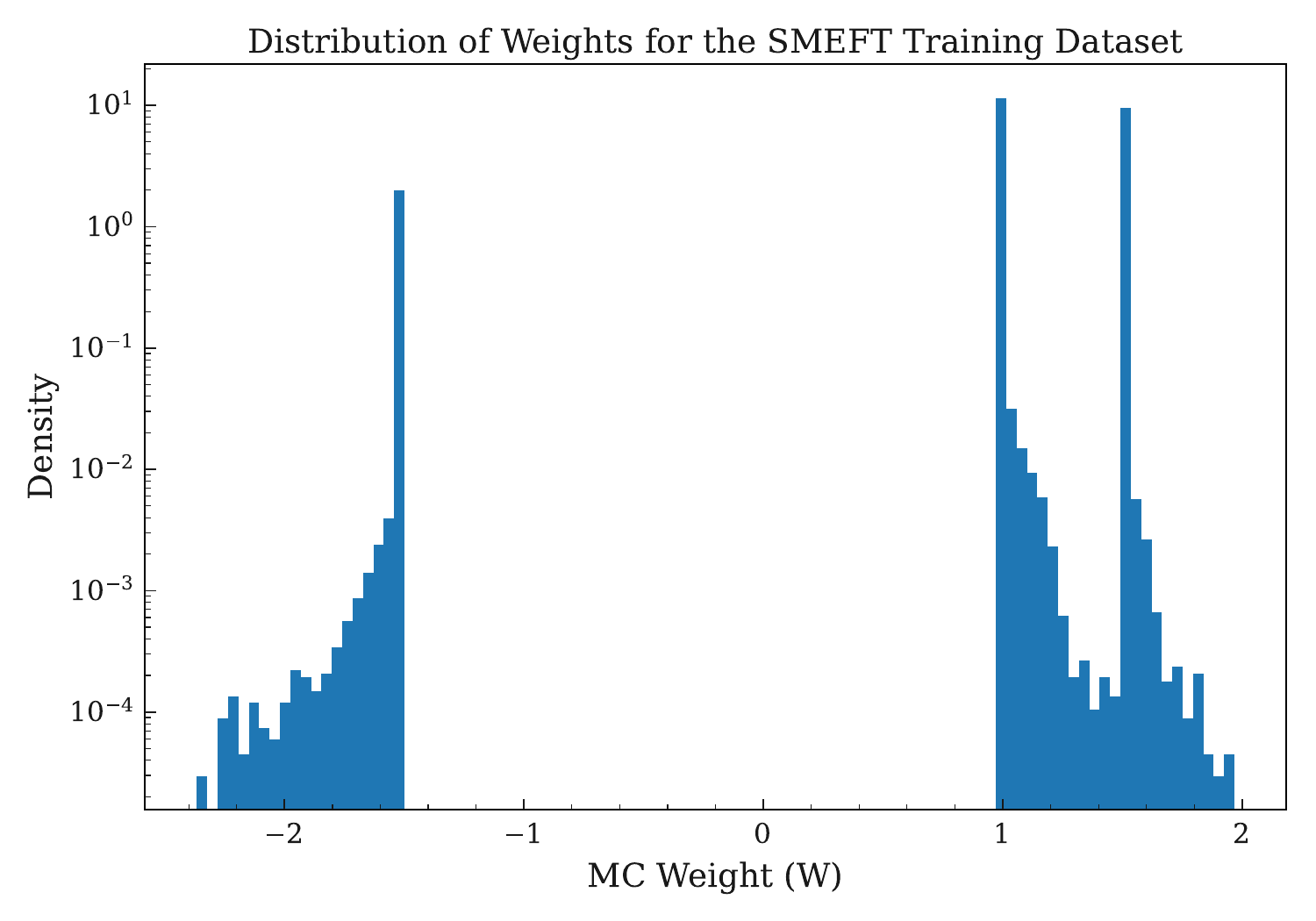}
\includegraphics[scale=0.34]{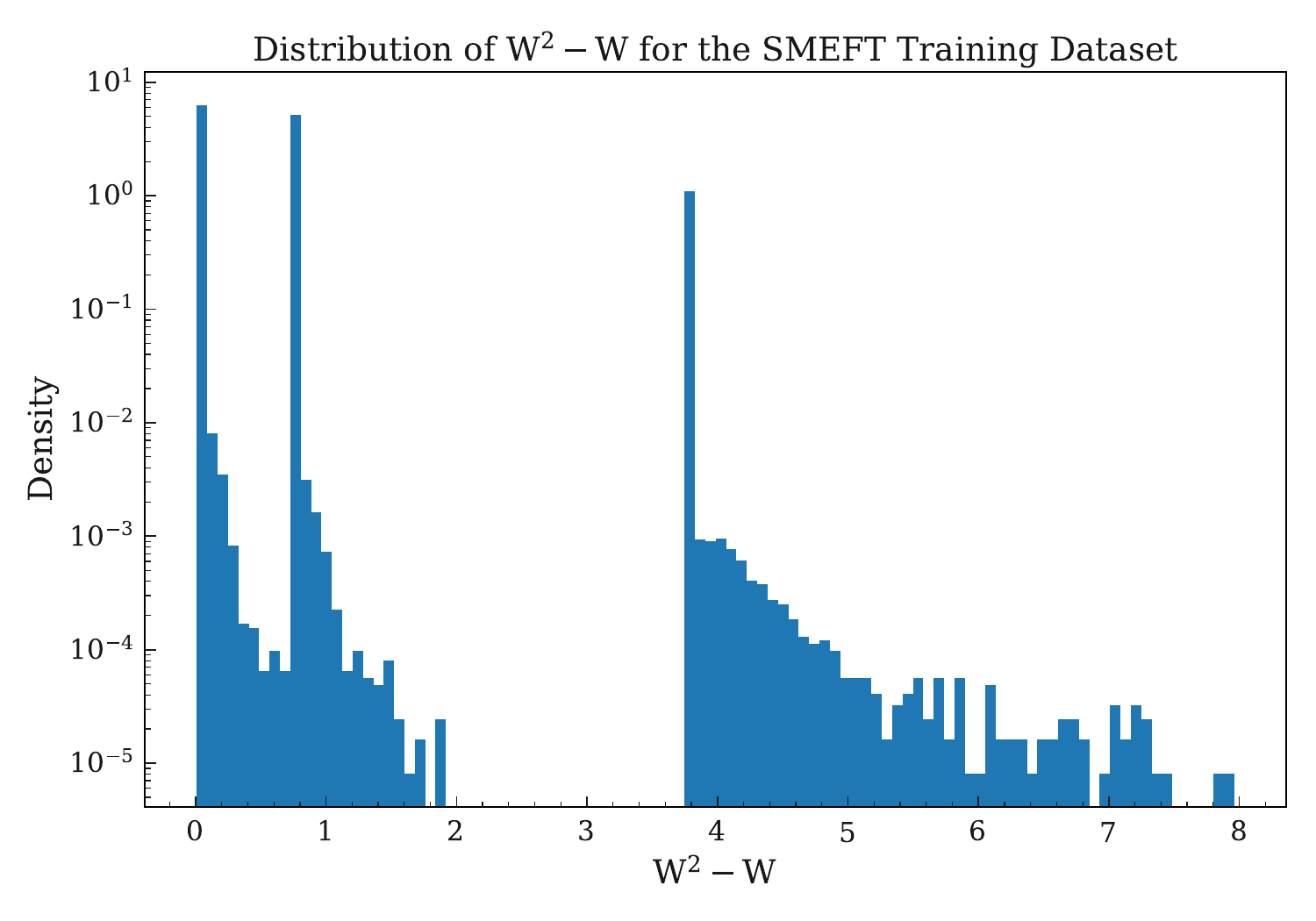}
\caption{(Left) Distribution of weights for the SMEFT training dataset. (Right) Distribution of $W^2-W$ for the SMEFT training dataset. Since the distribution of $W^2 - W$ is strictly non-negative this signifies strictly increased variance during the training process according to Equation \ref{eq:variance_gradient}.}\label{fig:smeft_weights}
\end{figure}

\begin{figure}[H]
\centering
\includegraphics[scale=0.28]{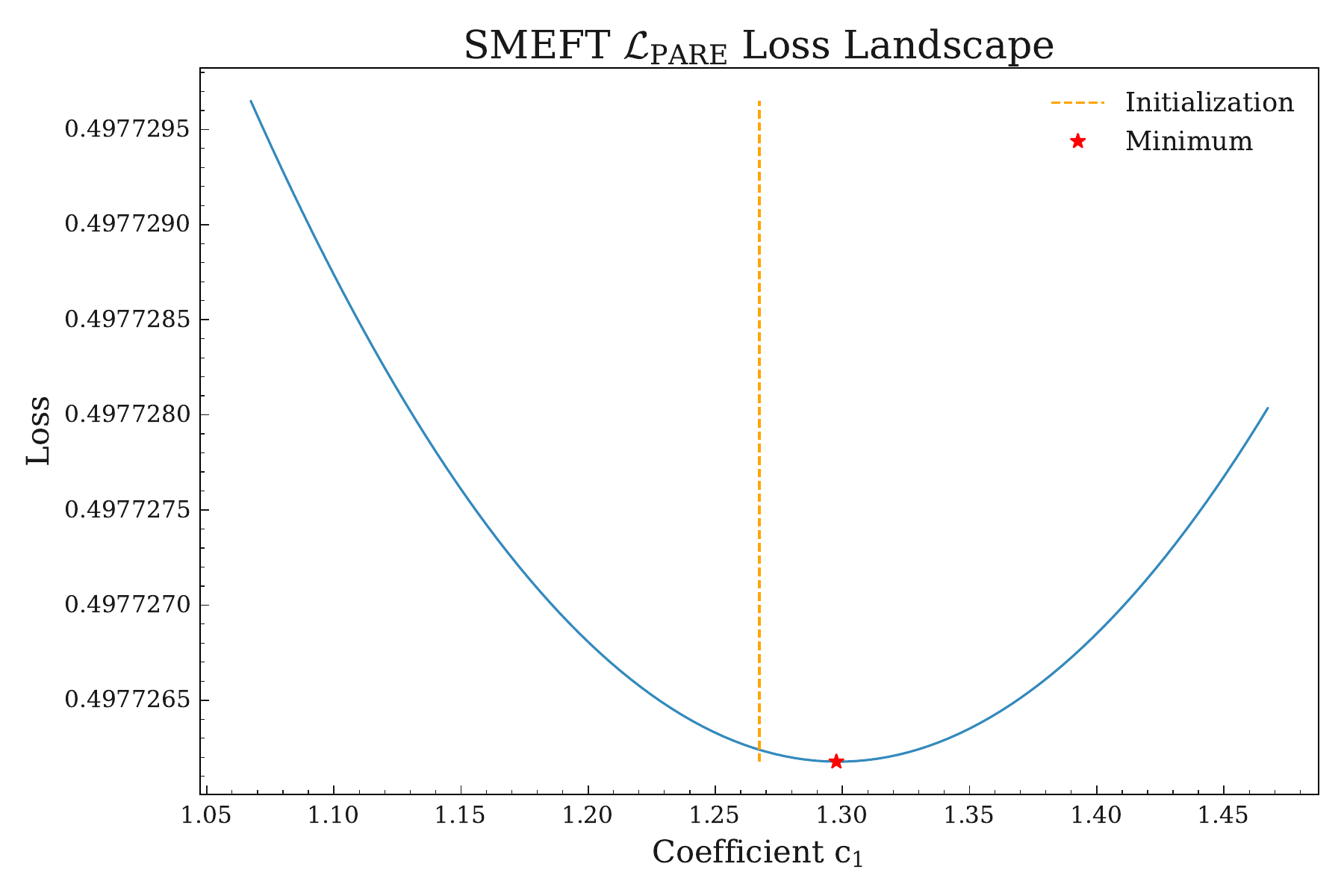}
\includegraphics[scale=0.28]{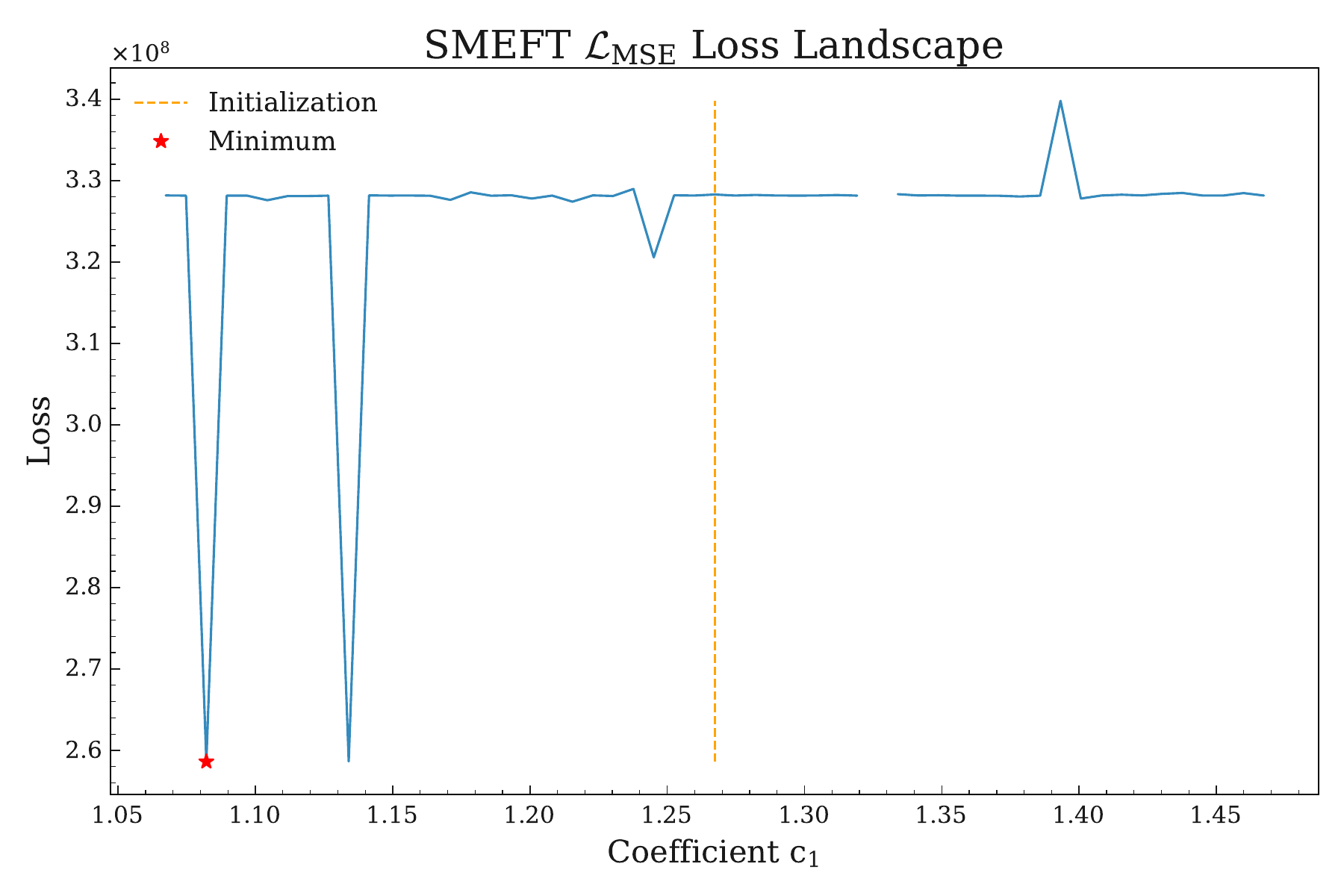}
\caption{The loss landscapes for optimizing the coefficient $c_1$ of the signed mixture model with the sub-ratio models fixed for the SMEFT application. The top figure shows the landscape using the new pole-adjusted ratio estimation loss function and the bottom figure shows landscape using the mean-squared error loss function.}
\end{figure}

\begin{figure}[H]
\centering
\includegraphics[scale=0.23]{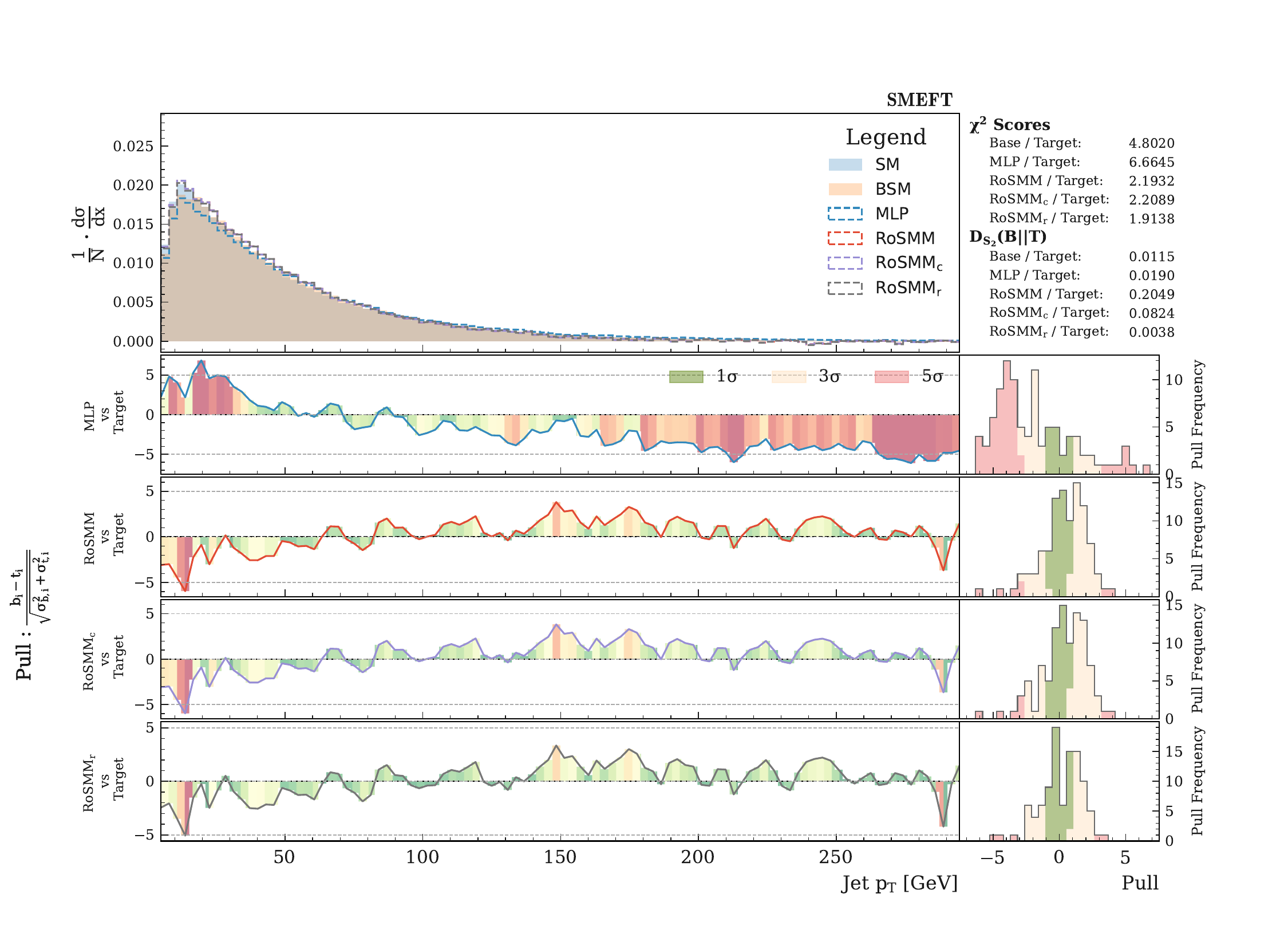}
\includegraphics[scale=0.23]{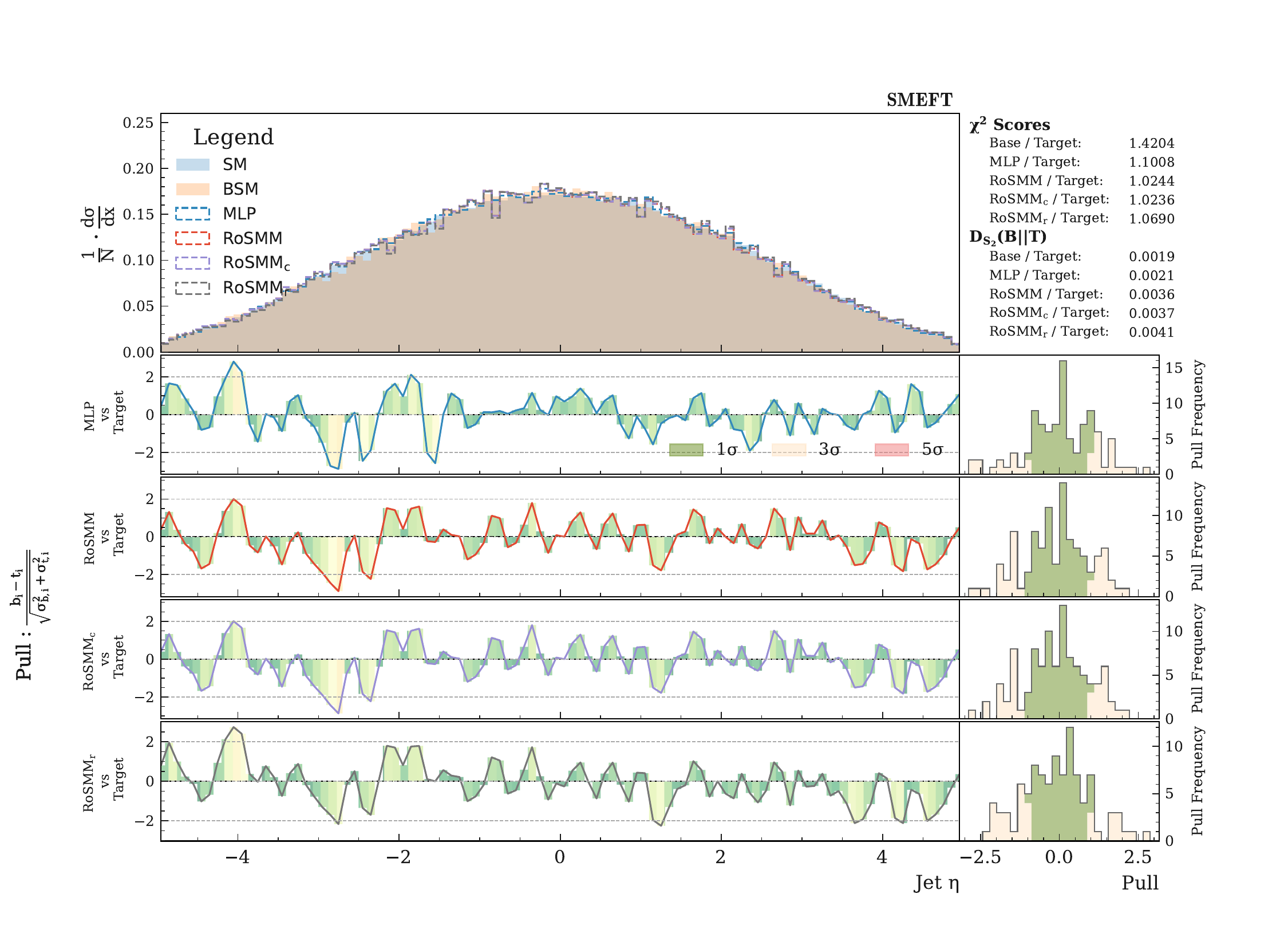}
\includegraphics[scale=0.23]{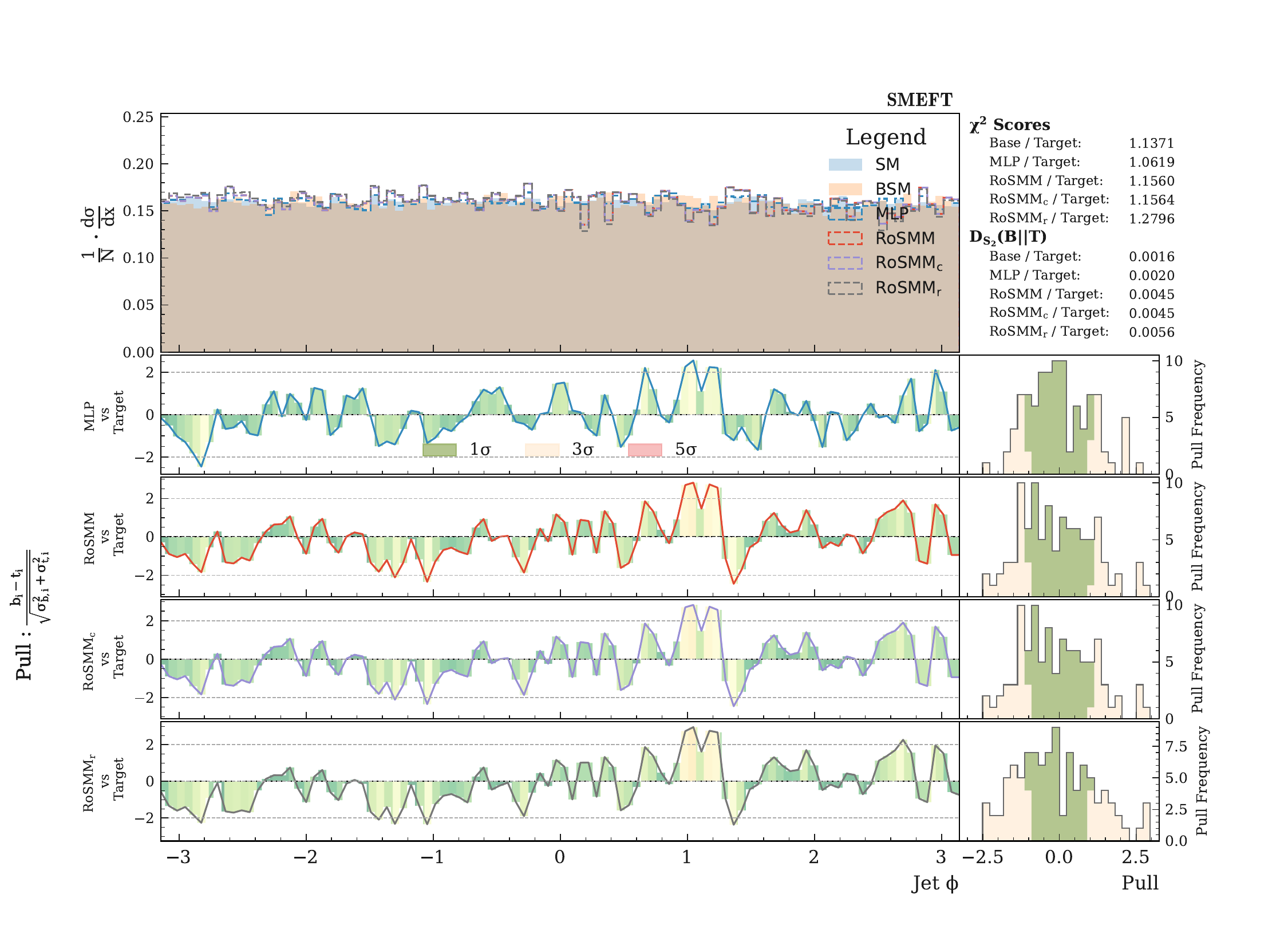}
\includegraphics[scale=0.23]{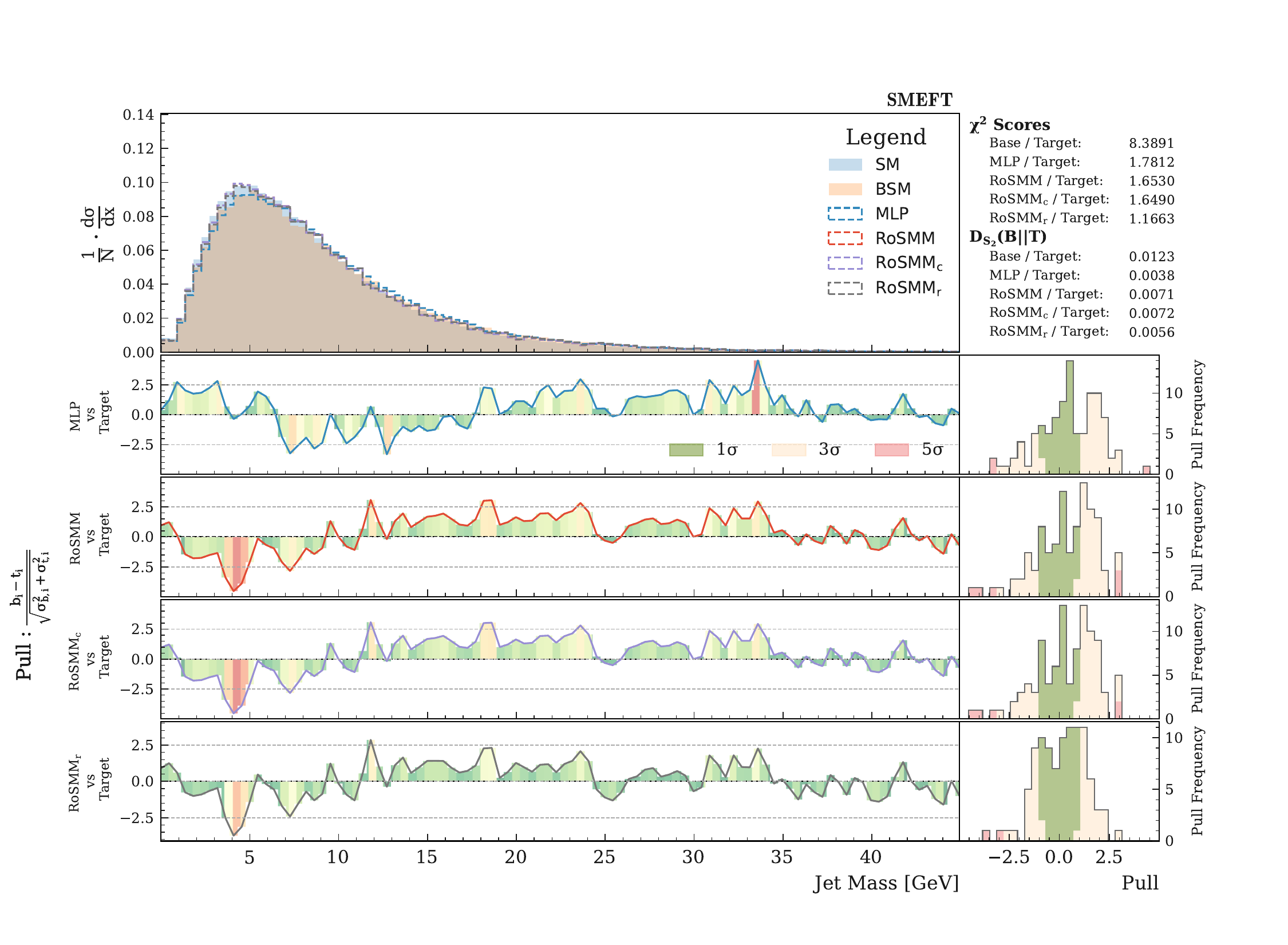}
\caption{Reweighting closure plots for the Jet features $(p_T,\eta,\phi,m)$ where the reference (Standard Model) distribution is mapped to the target (SMEFT) distribution using the different likelihood ratio estimation models.}
\end{figure}

\begin{figure}[H]
\centering
\includegraphics[scale=0.23]{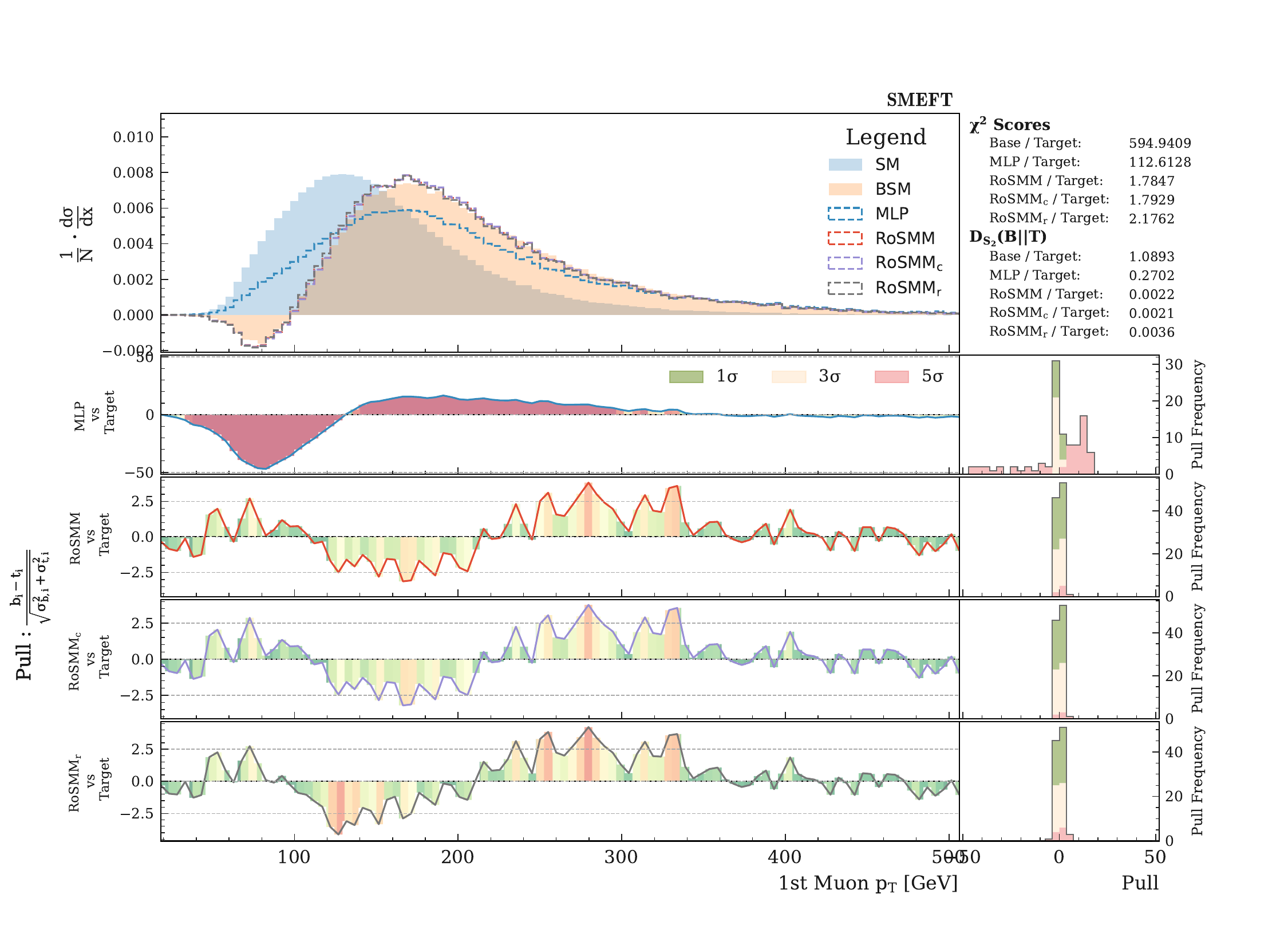}
\includegraphics[scale=0.23]{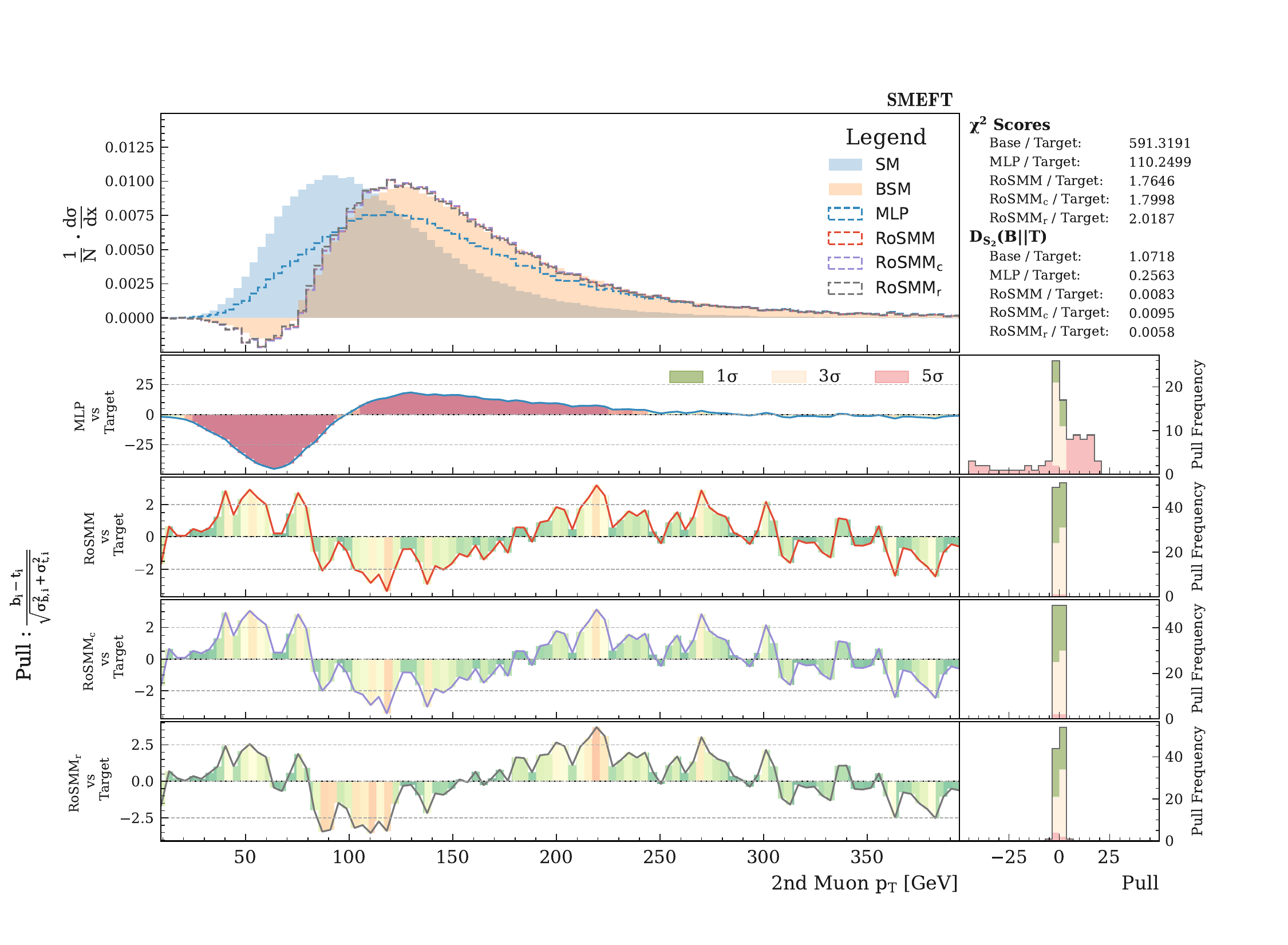}
\includegraphics[scale=0.23]{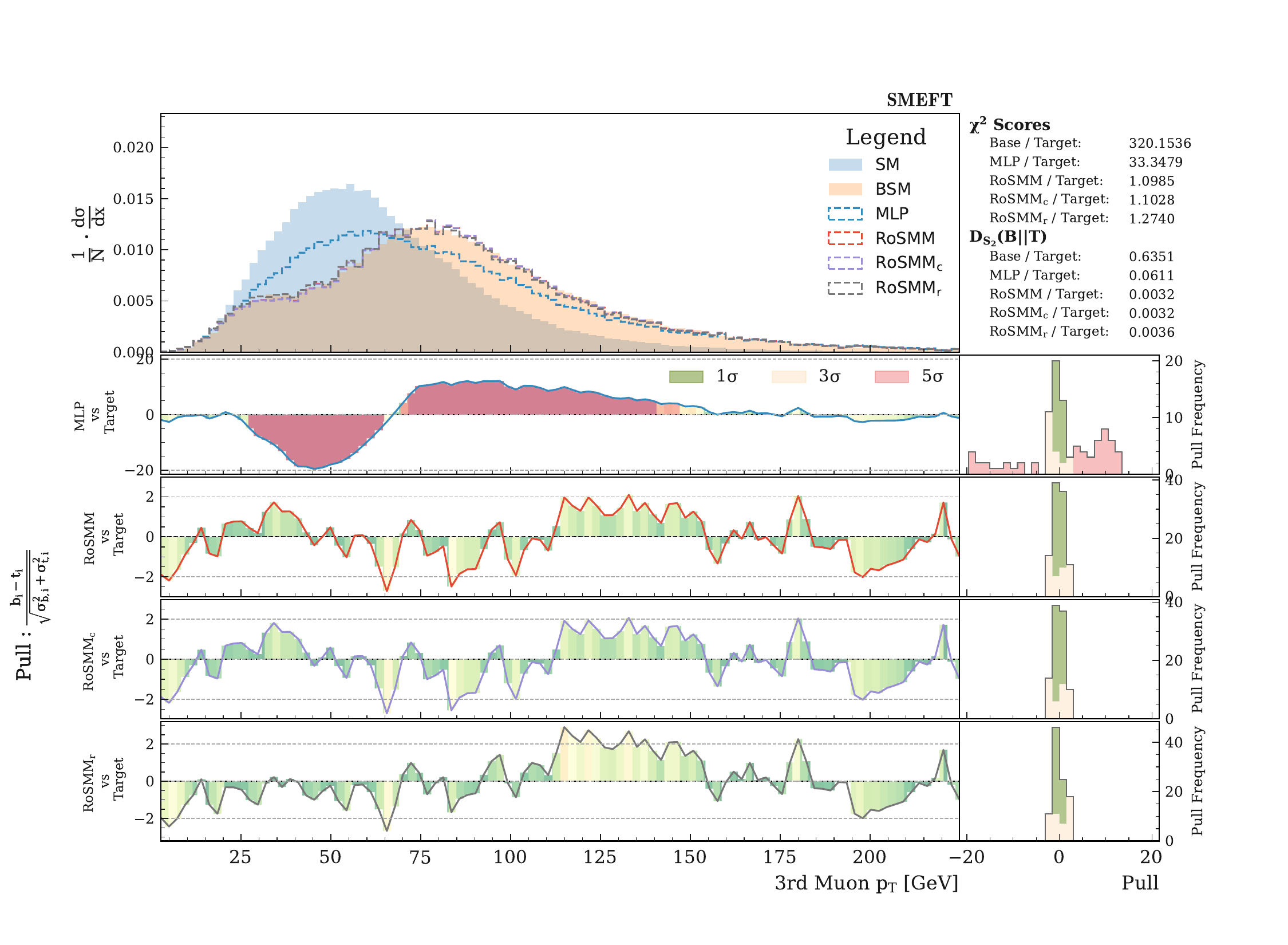}
\includegraphics[scale=0.23]{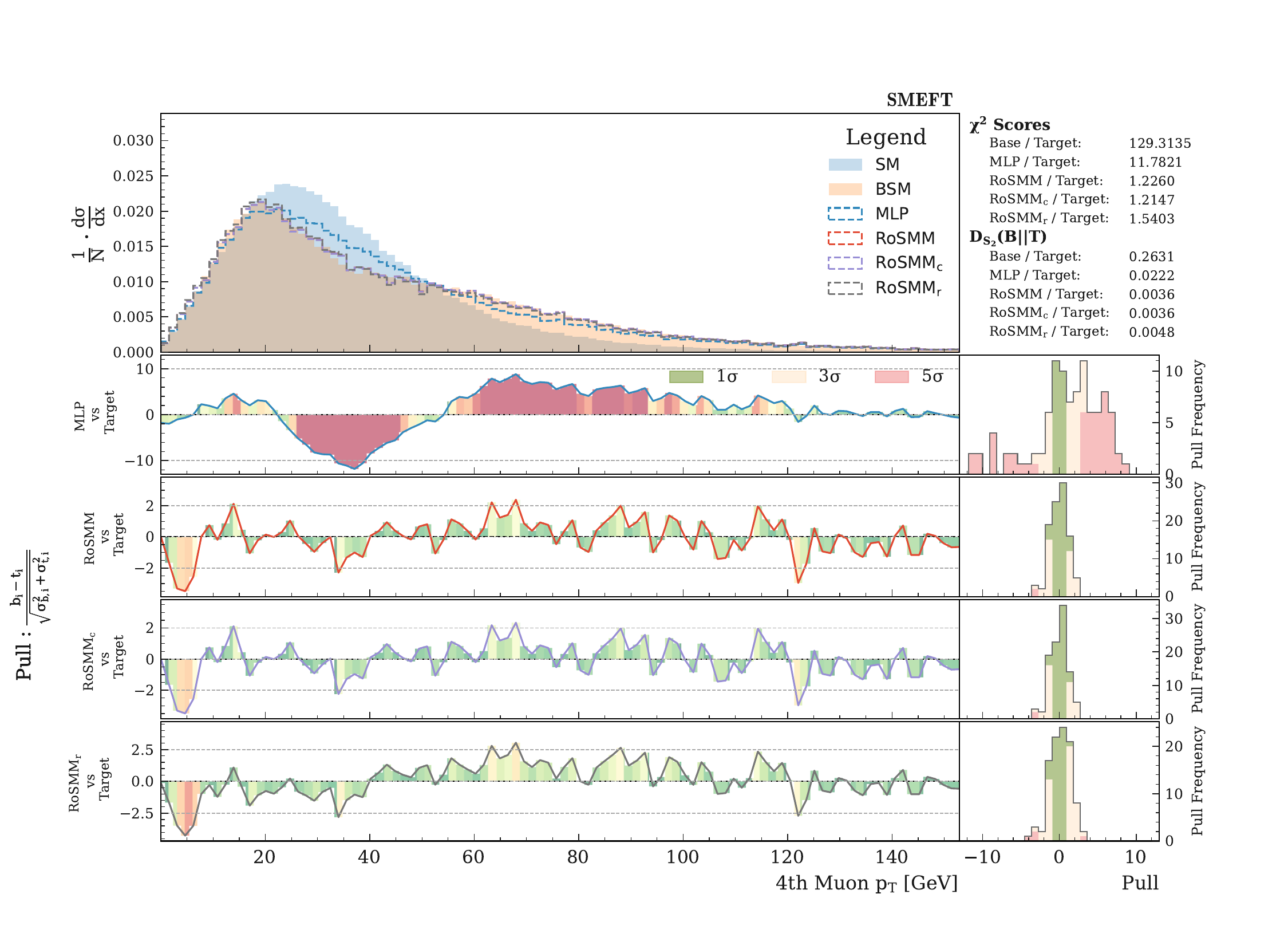}
\caption{Reweighting closure plots for the four muon $p_T$ features where the reference (Standard Model) distribution is mapped to the target (SMEFT) distribution using the different likelihood ratio estimation models.}
\end{figure}

\begin{figure}[H]
\centering
\includegraphics[scale=0.23]{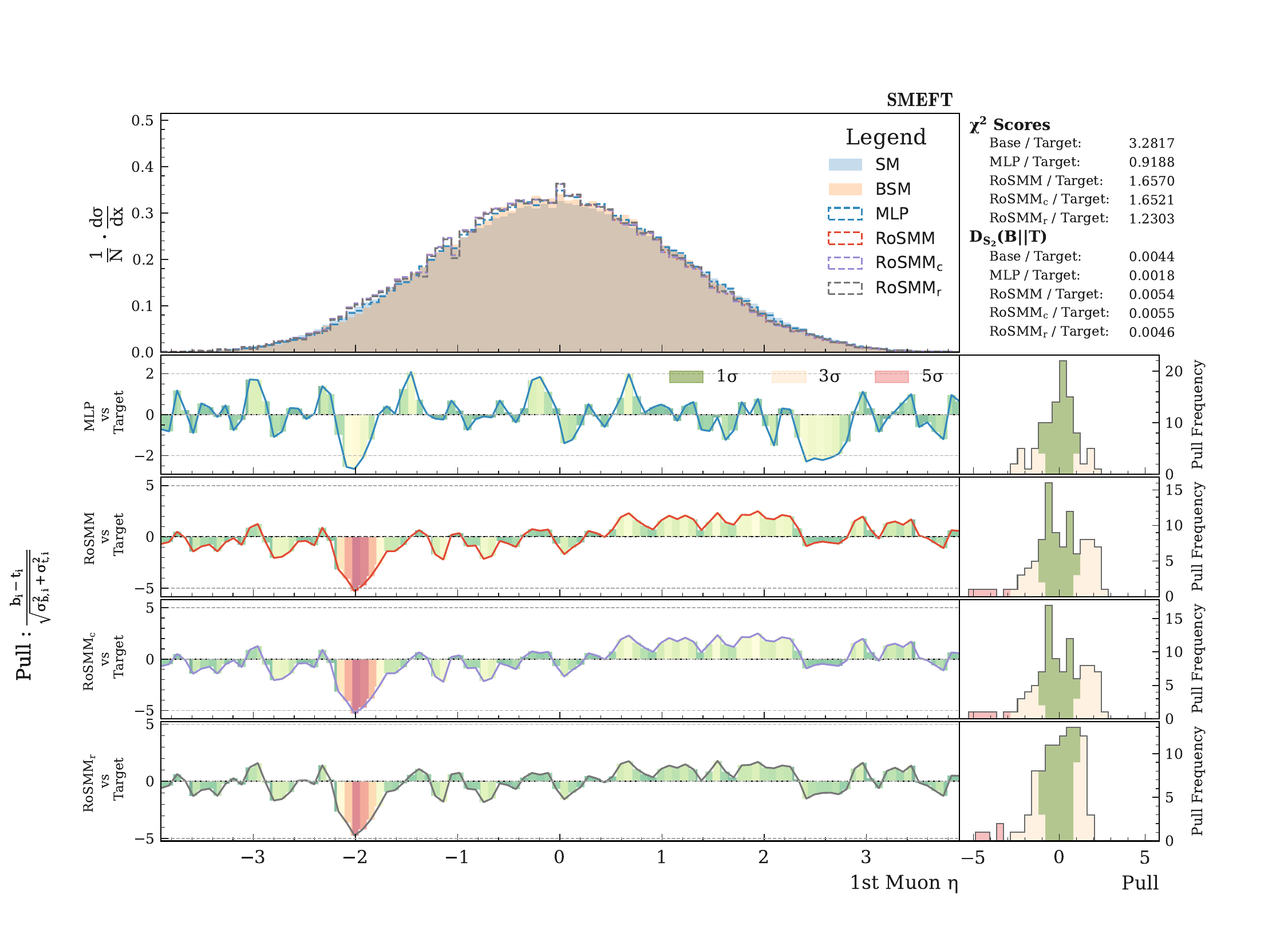}
\includegraphics[scale=0.23]{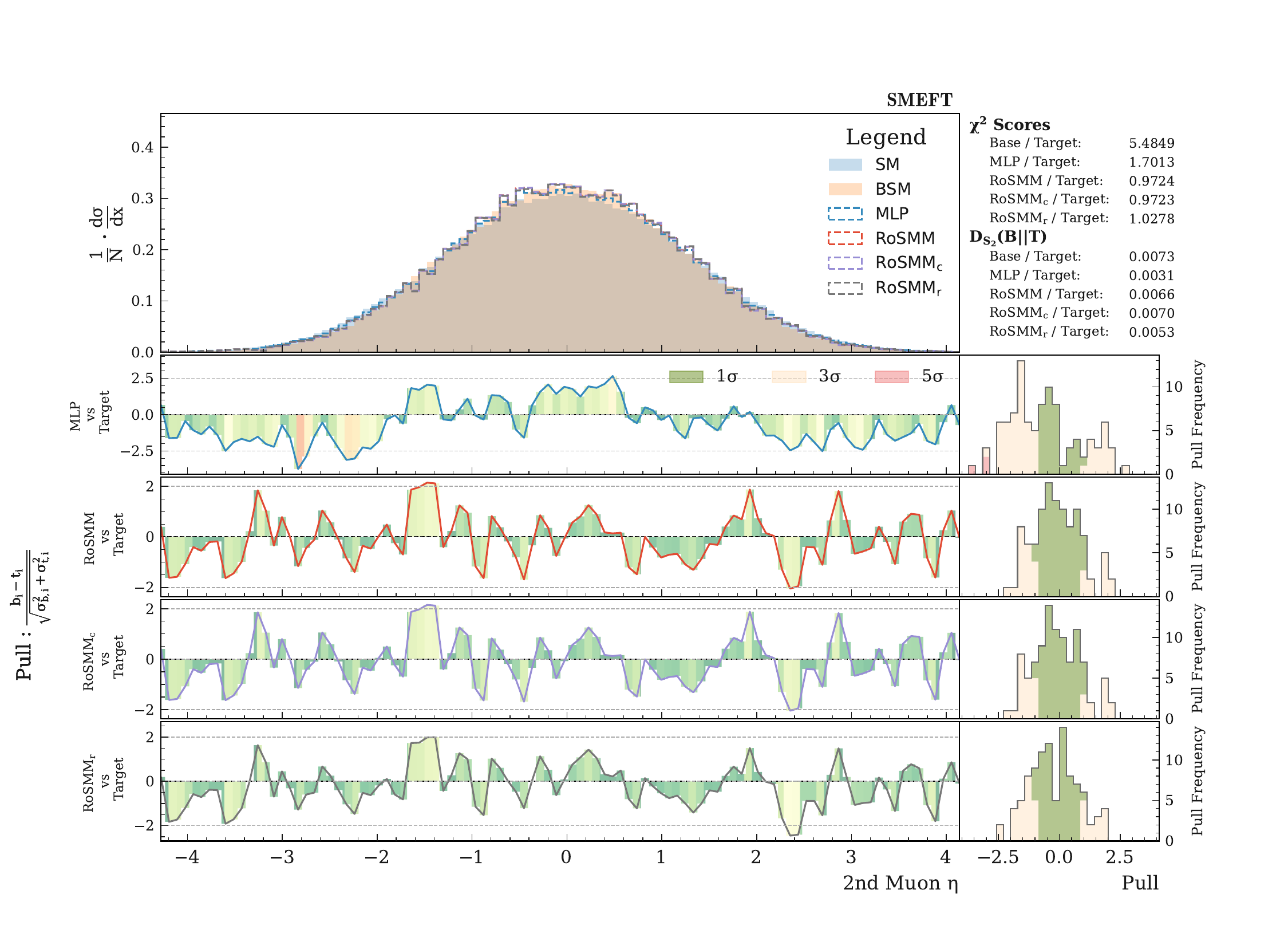}
\includegraphics[scale=0.23]{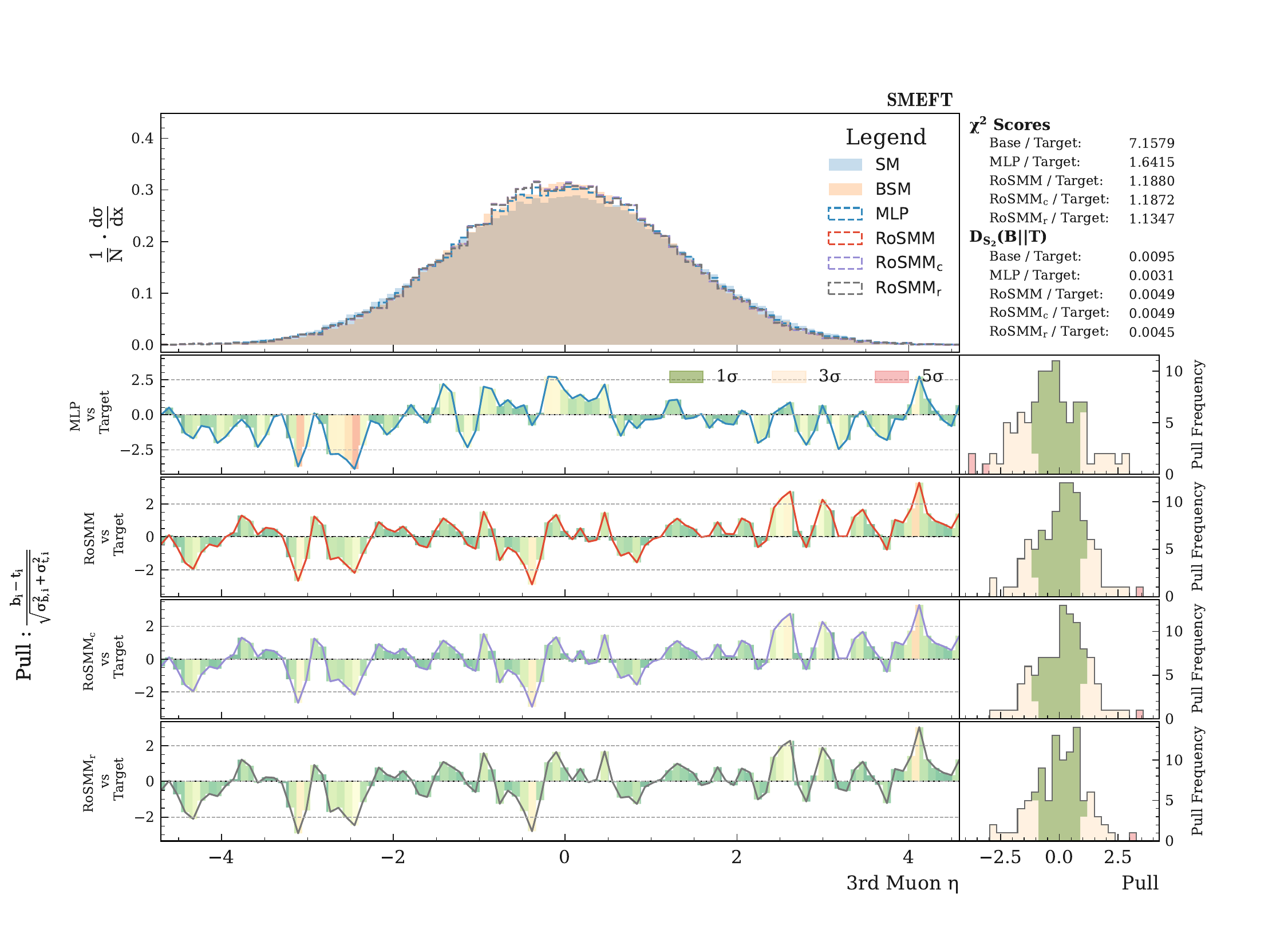}
\includegraphics[scale=0.23]{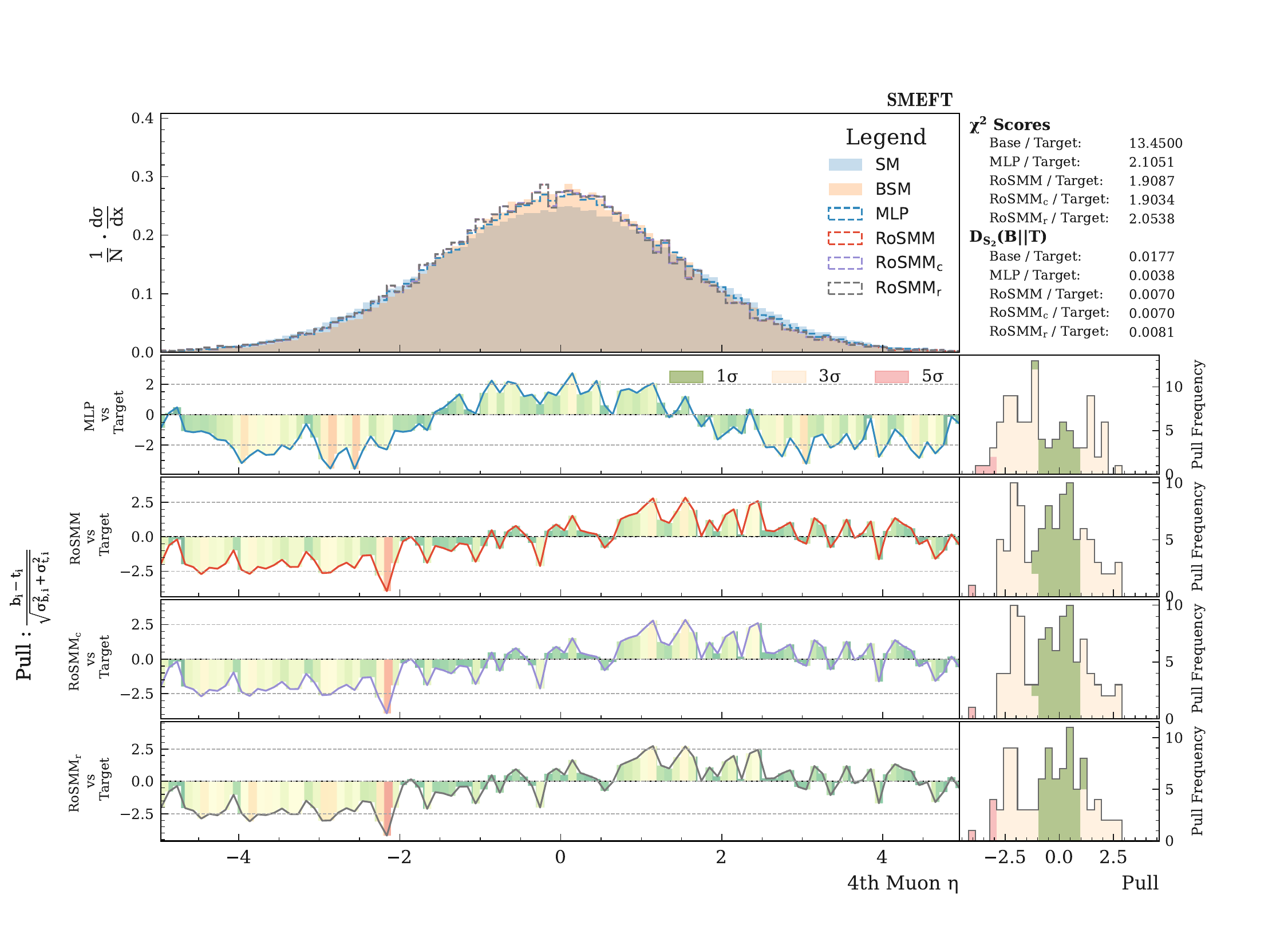}
\caption{Reweighting closure plots for the four muon $\eta$ features where the reference (Standard Model) distribution is mapped to the target (SMEFT) distribution using the different likelihood ratio estimation models.}
\end{figure}

\begin{figure}[H]
\centering
\includegraphics[scale=0.23]{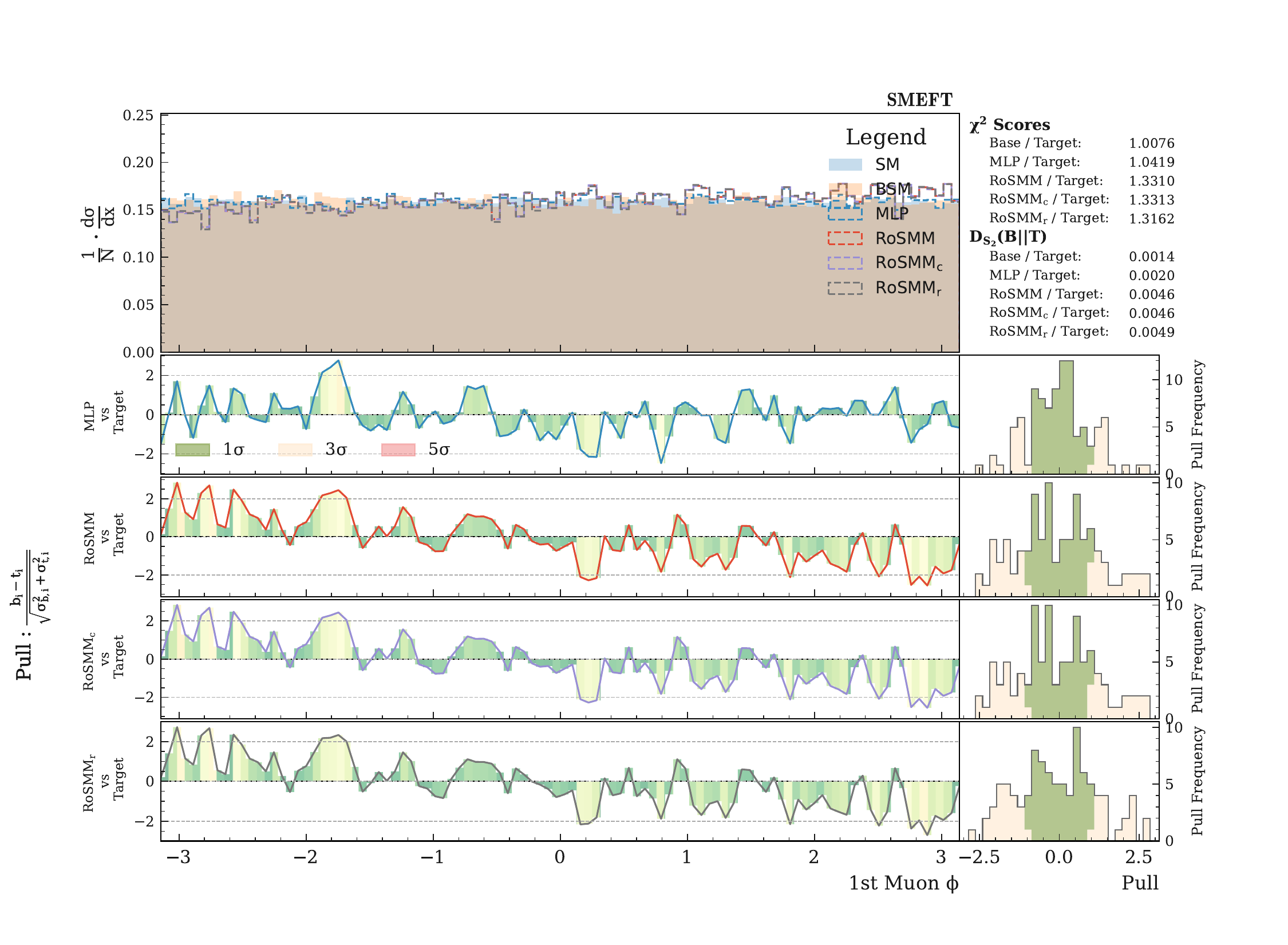}
\includegraphics[scale=0.23]{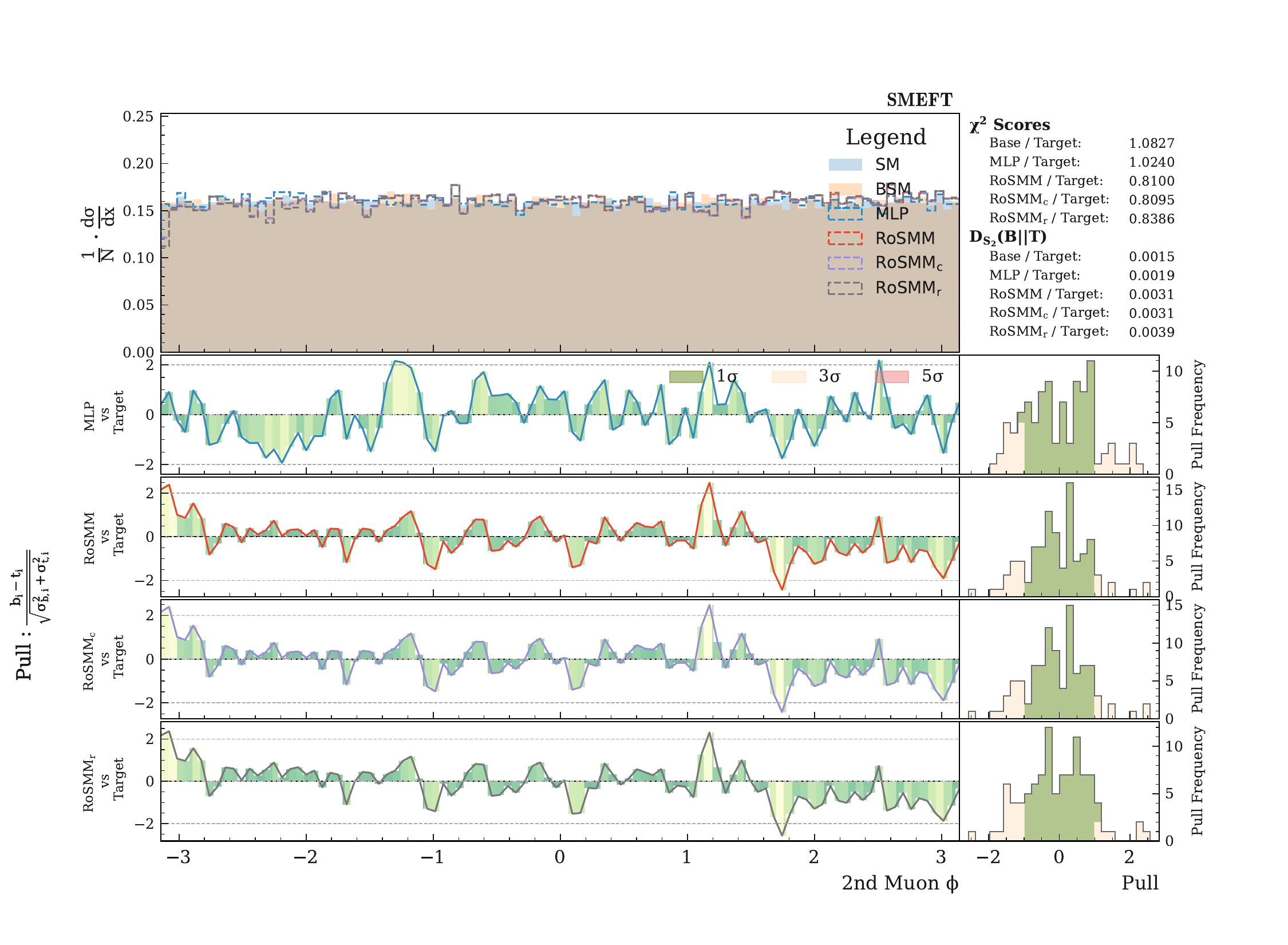}
\includegraphics[scale=0.23]{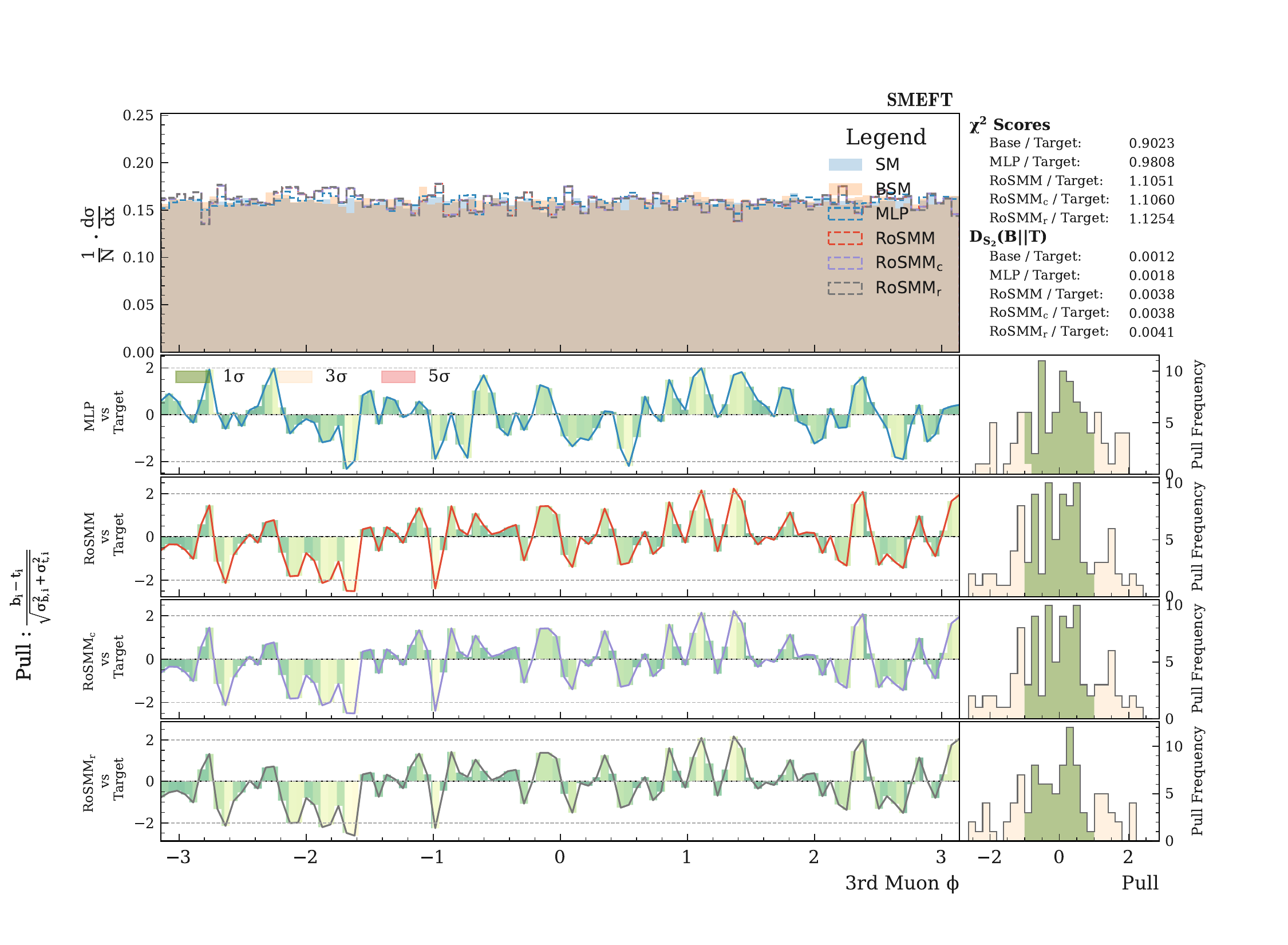}
\includegraphics[scale=0.23]{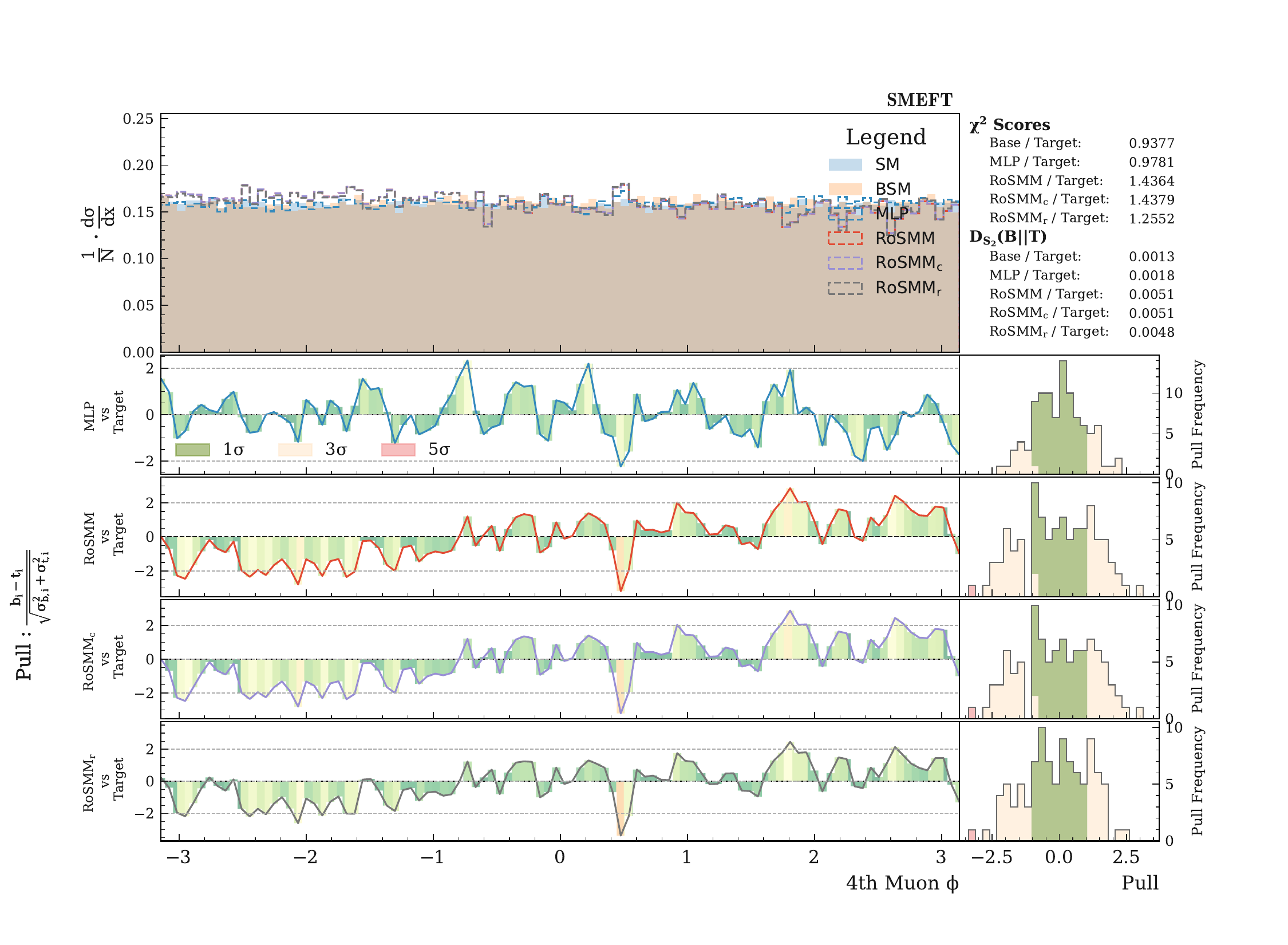}
\caption{Reweighting closure plots for the four muon $\phi$ features where the reference (Standard Model) distribution is mapped to the target (SMEFT) distribution using the different likelihood ratio estimation models.}
\end{figure}

\begin{figure}[H]
\centering
\includegraphics[scale=0.23]{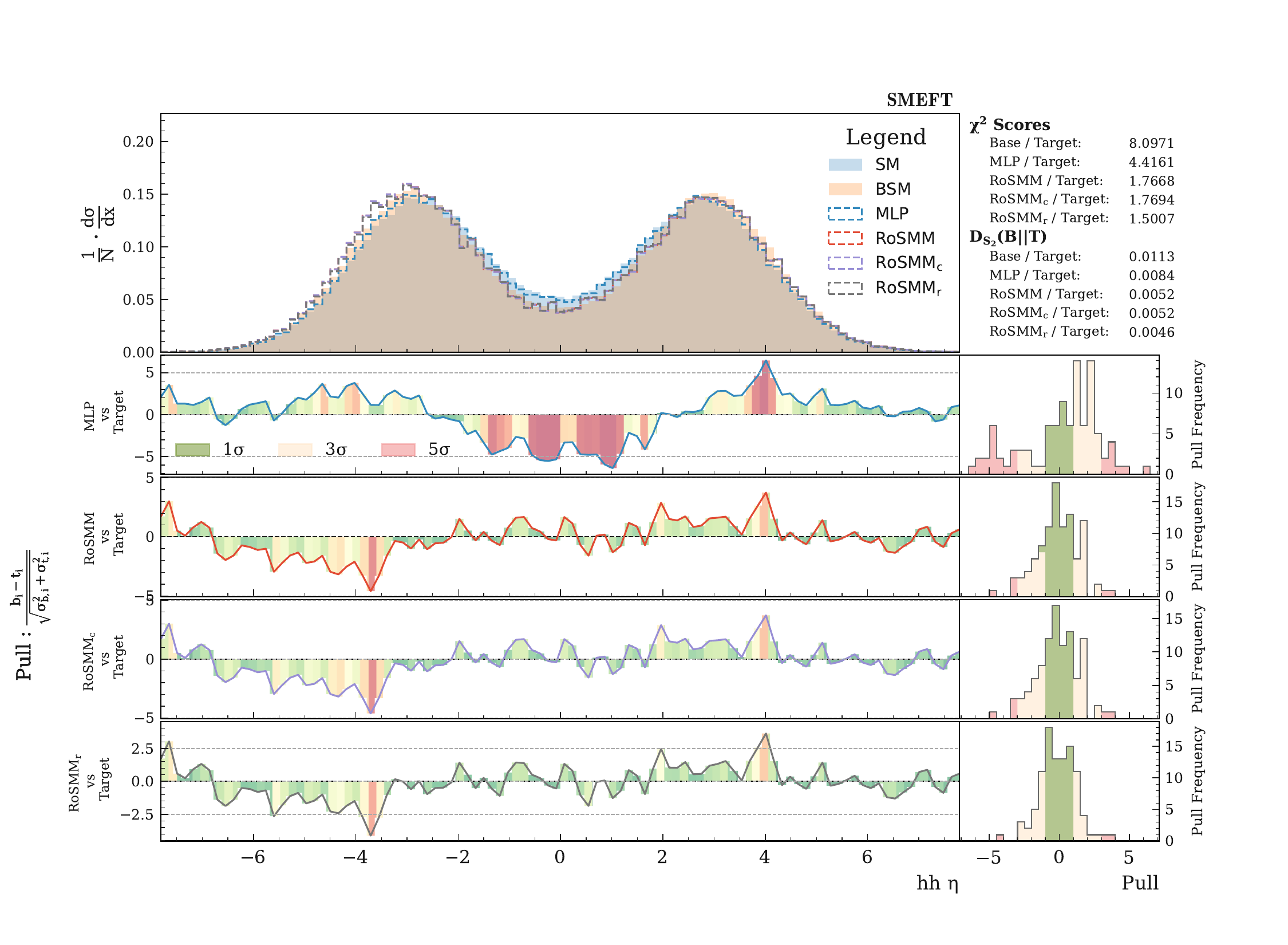}
\includegraphics[scale=0.23]{images/SMEFT-v3/SMEFT_hh_eta_closure.pdf}
\includegraphics[scale=0.23]{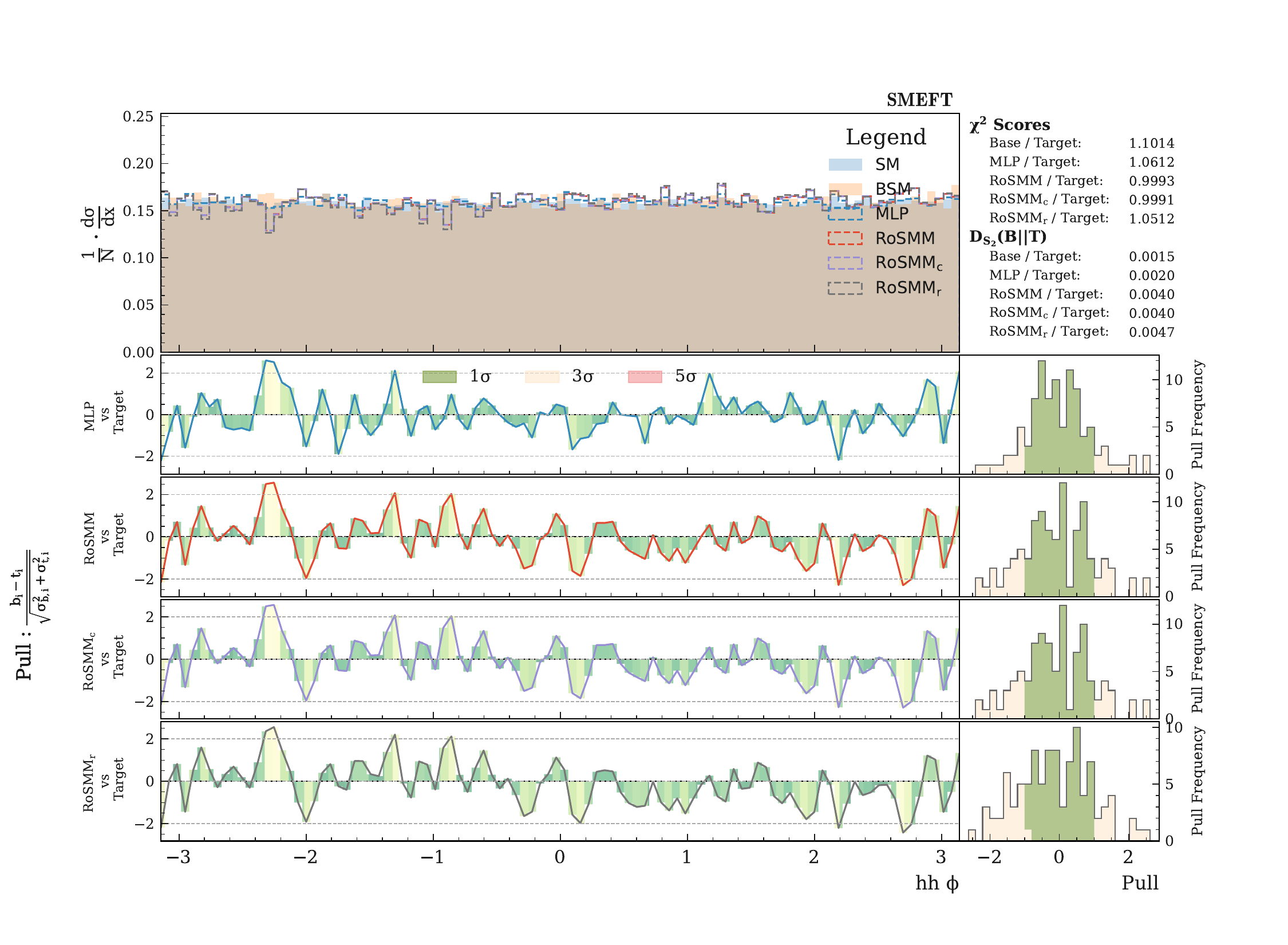}
\caption{Reweighting closure plots for the di-Higgs features $(p_T,\eta,\phi)$ where the reference (Standard Model) distribution is mapped to the target (SMEFT) distribution using the different likelihood ratio estimation models.}
\end{figure}

\end{document}